\pdfoutput=1
\documentclass[conference]{IEEEtran}

\usepackage{graphicx}
\usepackage[disable]{todonotes}
\usepackage{listings}

\providecommand{\annonymversion}[2]{#2} % 1=yes, 2=no
\providecommand{\shortversion}[2]{#2} % 1=yes, 2=no

\begin{document}
	
	%\title{Detecting Data Quality Problems by Clustering} % working title
%	\title{Detecting Quality Problems in Data Models by Clustering} % working title
	\title{Detecting Quality Problems in Data Models by Clustering Heterogeneous Data Values}

	\annonymversion
		{	\author{\IEEEauthorblockN{Anonymous Author(s)}
			\IEEEauthorblockA{\\}}}
		{	\author{\IEEEauthorblockN{Viola Wenz}
				%\orcid{https://orcid.org/0000-0002-2176-9463}
				\IEEEauthorblockA{Philipps-Universität Marburg\\		
					viola.wenz@uni-marburg.de}
				\and
				\IEEEauthorblockN{Arno Kesper}
				%\orcid{https://orcid.org/0000-0002-5042-1087}
				\IEEEauthorblockA{Philipps-Universität Marburg\\		
					arno.kesper@uni-marburg.de}
				\and
				\IEEEauthorblockN{Gabriele Taentzer}
				%\orcid{https://orcid.org/0000-0002-3975-5238}
				\IEEEauthorblockA{Philipps-Universität Marburg\\		
					taentzer@uni-marburg.de}}}
	
	% Todos:
	\providecommand{\viola}[1]{\todo[inline, backgroundcolor=yellow]{#1}}
	\providecommand{\arno}[1]{\todo[inline, backgroundcolor=cyan]{#1}}
	\providecommand{\gabi}[1]{\todo[inline]{#1}}
	\providecommand{\question}[1]{\todo[inline, backgroundcolor=green, disable]{#1}} % open questions for further research
	
	\maketitle

\begin{abstract}
         % clustering: group similar data values, (identify heterogenous values)
	% goal: give an overview of the syntax of a large set of data values and thereby support the identification of problems
	% interpretation per cluster: Why are the values structured like this? Which type(s) of problem(s) may have caused this? Find a headline for each cluster.
	% approach: bottom-up	
         % cluster analysis is suited for finding problems in sets of data values
         % these problems may occur in various forms such as sets of data, instance models
	% find problems that manifest in data
	% problems: aquisition (human, software), data model, transformations.  For example,  data model is not well understood or does not suit well to collected data 
	% expectation for data in one field: good data = homogenous, bad data = heterogenous
	% assumption: data values typically are syntactically similar iff they are semantically similar 
	
	% ------------------------------------
	
	% new version:
%	\viola{Modified}
	Data is of high quality if it is fit for its intended use.
	The quality of data is influenced by the underlying data model and its quality.
	One major quality problem is the heterogeneity of data as quality aspects such as understandability and interoperability are impaired.
	This heterogeneity may be caused by quality problems in the data model.
	Data heterogeneity can occur in particular when the information given
	is not structured enough
	and just captured in data values, often due to missing or non-suitable structure in the underlying data model.
	We propose a bottom-up approach to detecting quality problems in data models that manifest in heterogeneous data values.
	It supports an explorative analysis of the existing data and can be configured by domain experts according to their domain knowledge.
	All values of a selected data field are clustered by syntactic similarity.
	Thereby an overview of the data values' diversity in syntax is provided.
	It shall help domain experts to understand how the data model is used in practice and to derive potential quality problems of the data model.
	We outline a proof-of-concept implementation and evaluate our approach using cultural heritage data.

	% ------------------------------------
	
	% Gabi's original version:
	
%	%As scientific progress highly depends on the quality of research data, there are strict requirements for data quality coming from the scientific community. 
%       Data are of high quality if they are fit for their intended use.  
%	The first essential step to data quality assurance is to analyse inherent quality problems. 
%        A major part of information in data may be non-structured and is just captured in data values.
%        We use cluster analysis for finding clusterings in sets of data values that may point to quality problems in data acquisition, data models or transformations.
%	We evaluate our approach on cultural heritage data and focus on problems in data models that manifest in heterogeneous data. 
%	These problems may occur due to missing or non-suitable structure in data models.

	% ------------------------------------
	
\end{abstract}

%If selected data fields show heterogeneous values, cluster analysis may be well-suited for finding clusters in such sets of data values that point to quality problems.
%Due to dynamic digitalization in certain scientific fields, especially the humanities,  data models may change rather quickly. 
%	Therefore, we present an approach to cluster analysis that is generic concerning data models and underlying data base technology.

\section{Introduction}
%\viola{Currently we have exactly 8 pages.}
%\viola{Slightly modified.}
%\gabi{Revised.}
%Why is this topic interesting and relevant?
% in the context of the digital transformation data is pervasive
% high quality data are important for using data
% many data quality problems
% often undetected
The digital transformation of our society is an ongoing challenge and nearly every action in the digital world creates data.
To effectively use the collected data, it has to be of high quality.
Data quality, often defined as ``fitness for use'' \cite{dataquality}, comprises various dimensions, such as completeness, accuracy, and timeliness. 
%Quality problems such as imprecise, heterogeneous and missing data, can be manifold and may affect one or more quality dimensions. 
%Quality problems, such as heterogeneous and imprecise data, can be manifold and may affect one or more quality dimensions. 
Quality problems can be manifold and may affect one or more quality dimensions. 
%\arno{Modified following 2 sentences: Added cause}
The first essential step to better data quality is to identify existing quality problems in data and investigate their causes.
%\gabi{Is heterogeneity identified as data quality problem?}

%What is the problem?
% observation: often single data fields as uncontrolled string values, implicit encoding of information
% assumption: structure of strings should be homogenous
% types of problems: data acquisition, data model, data transformation
% show a simple but illustrative example
%When heterogeneous data is involved, the cause for quality problems may also
%\viola{Moved below sentence up here.}
%\viola{Added general sentence below with reference. Reworded following sentences. We could not find a paper that explicitly states that heterogeneous values \emph{of the same field} are problematic.}
Heterogeneity, also referred to as representational inconsistency,  is considered a data quality problem~\cite{zaveri2016quality} in general.
We focus on heterogeneity of data values stored in the same data field.
%Heterogeneity of data values stored in the same data field is a quality issue as it is more difficult to process such unstructured data than clearly structured data.
%It is more difficult to process such unstructured data than clearly structured data.
%The cause for heterogeneous values in the same data field 
Its cause may be the underlying data model as long as it allows such a heterogeneity. 
%In that sense, 
Data fields that contain heterogeneous data values further may point to a lack of structure in the related data model. 
Since ``data quality techniques become increasingly complex as data looses structure'' \cite{BCFM09}, a technique for systematically analysing data fields in this respect is needed as an essential step towards data model improvement.
%\gabi{Modified.}
%\viola{ok}
%In the context of data management, authors typically distinguish three types of data~\cite{BCFM09}: % (implicitly or explicitly): 
%(1) {\em Structured data} (often stored in relational databases) is an aggregation of entities described by attributes of elementary types such as integers and strings.
%(2) {\em Unstructured data} is a generic sequence of symbols often expressed in natural language.
%Typical examples are annotations or explaining texts.
%(3) {\em Semi-structured data} (often stored in XML databases) is data with structure that allows some flexibility. 
%%Data may be structured differently or occur more than once, even within a single data set.
%While the quality of structured data is often well analysed with metrics or patterns\todo{cite}, ``data quality techniques become increasingly complex as data loses structure.'' \cite{BCFM09} 
%\todo[inline]{Existing work that analyses the quality of semi-structured or unstructured data?}
%% hier mehr yu data model quality, wie wirkt sich Heterogenit't auf qua.itat aus>

%\arno{Cultural heritage is only one example for a domain with similar problems! Cultural heritage community continuously tries to improve data and data models.}
%Considering cultural heritage data, for example, artist names, titles of cultural objects and dates can be very heterogeneous in form and content even though domain experts are constantly trying to improve the quality of their data and data models.
Considering cultural heritage data, for example, domain experts constantly try to improve the quality of their data and data models.
Nevertheless, data values, such as artist names, titles and creation dates of cultural objects, can be very heterogeneous in form and content.
%\viola{Reworded to add positive statement about domain above.}
In this domain, data is typically collected manually and data models often allow wide ranges of data.
%\viola{Added sentence below.}
Data transformations introduce further heterogeneity.
Let's consider the width and height information of cultural objects, for example. 
Stored as semi-structured data, there may occur entries such as \lstinline|101.2 cm|; \lstinline|2 m|; \lstinline|3.5 x 4.5 cm|; \lstinline|100 m?| and even \lstinline|-|.
They show that the content of this data field is underspecified. 
It is used not only to store one value but may contain also information about the measurement unit as well as several values. 
Even meta information such as uncertainty (expressed with \lstinline|?|) and lack of knowledge (denoted by \lstinline|-|) are stored in the same data field. 
This heterogeneity may indicate quality problems of the underlying data model, in particular a lack of structure to represent such additional information.
%\viola{Added sentence above to explain relation to data model problems.}
%such as missing structure and lack of syntax constraints.
%\viola{Added reference below.}
Similar problems may occur also in other domains where data is often semi-structured and entered manually, such as the domain of biodiversity~\cite{biodiversity}.
%Similar phenomenons may occur in other semi-structured data that is entered manually, e.g. biodiversity data.
%\viola{Added biodiversity data above. GT: Slightly rephrased.}

%This 
Heterogeneity of data stored in the same field is a quality issue as it is more difficult to process such unstructured data than clearly structured data. 
The causes for heterogeneous data in single data fields can be manifold: 
The acquisition of data is not as accurate as it should be, the data management software and the underlying data model are not adequate for the kind of data acquired or the transformation that produced that data is faulty. 
We focus on the quality of the underlying data model here as it is the core of data management and therefore, plays a critical role. 
Acquisition software and actual data acquisitions as well as data transformations all depend on data models. 

%State-of-the-art of analysing data model quality
%Several quality aspects, metrics for measuring them, not enough validation of those metrics \cite{MS94,MS03}
% Bad smells for meta-models \cite{LGL14,LGL14b} and guidelines for DSLs~\cite{KKPRSV14}
%There is no technique to explore unknown problems in data models, problems that manifest by problematic data quality
%It seems that clustering has not been used to analyse data model quality
% different to schema discovery since unstructured data is considered here, clustering makes (parts of) inherent structure visible
%\viola{Maybe remove the following paragraph because of criticism by reviewer? The paper is briefly discussed in related work anyway.}
The importance of data model quality was discovered early. 
A conceptual basis for data model quality management was laid by Moody and Shanks in \cite{MS94}. 
They developed a framework with quality factors, such as completeness and understandability, 
%like completeness and understandability,
%of varying importance, 
quality metrics, and improvement strategies; they evaluated it in~\cite{MS03}. 
%The quality factors they identified for data models are completeness, simplicity, flexibility, integration, understandability, and implementability.
%They identified four metrics that are useful to measure data model quality: (1) number of entities and relationships (to measure simplicity), (2) development cost estimate( for implementability), (3) reuse percentage (for integration), and (4) number of defects by quality factor (which concerns all quality factors). 
One of their findings was that metrics are of limited use for analysing data model quality as research participants rated qualitative descriptions of quality problems more useful than quantitative ones. 
%Instead,  the authors of \cite{MS03} 
%Instead, they revealed that the involvement of all stakeholders in the data modelling process was more important than any other issue in achieving quality improvement. 
%strongly influence data model quality. 
%\viola{I think the authors made that observation before they applied their framework.}
%\gabi{Right. I reformulated the sentence.}
%These findings are clear indicators that %the effectiveness of metrics as analysis techniques for data model quality is limited. 
%some qualitative analysis of data model quality is needed to support the quality improvement process.
This has motivated us to develop a qualitative analysis of data model quality. %in the following. 
% auch beim clustering ist es n;tig, dass die stakeholders die claustering ergebnisse anschauen und auswerten...
%\gabi{Modified.}
%\viola{ok}

% What is the solution idea? Is this idea new?
% idea: detect heterogenous values by clustering; identify problems based on clusters of syntactically similar values...
% state-of-the-art of clustering approaches for detecting data quality problems
% advantages of bottom-up approach (e.g. compared to pattern-based approach)
%In this paper, we present a clustering approach to cluster heterogeneous values in selected data fields. 
%In this paper, we present a new approach to the analysis of data model quality.
Based on the observation that heterogeneous data is difficult to process, we present a  {\em bottom-up approach that clusters values of selected data fields such that
	domain experts can explore unknown quality problems and requirements of data models.}
%	quality problems of the underlying data model can be identified. }
%It can be configured according to domain knowledge.
%For achieving suitable clusterings, domain experts can 
%explore various possible clusterings by configuring the approach 
%configure the approach 
%according to their domain knowledge. 
%The clusterings provide an overview of heterogeneous values and thus support domain experts in identify quality issues in data models.
%Domain experts may use this approach to identify quality issues in data models based on their domain knowledge. 
%As the clustering approach starts from existing data, this bottom-up approach may be used for exploring unknown %kinds of 
%By starting from existing data, this bottom-up approach allows exploring unknown %kinds of \\
%By starting from existing data, this bottom-up approach provides an overview of heterogeneous data values and thereby allows domain experts to explore unknown %kinds of 
%quality problems and requirements of data models. 
%\viola{Added ``and requirements'' above. Reworded sentence below. OK?}
%Examples for such quality problems are 
For example, uncertain knowledge that is implicitly expressed and non-expected information given may indicate problems of the data model.
%Examples for characteristics of data values that may indicate problems of the data model are the following ones: 
%%Examples are 
%%	lack of knowledge and doubts that are implicitly expressed, 
%	missing and uncertain knowledge that is implicitly expressed, 
%	non-expected information such as several entries combined in one entry, 
%	non-expected literals of an enumeration, 
%	and information that is just given for disambiguating entries.
%	\todo{what more?}	
%For achieving suitable clusterings, domain experts can 
%%explore various possible clusterings by configuring the approach 
%configure the approach 
%according to their domain knowledge. % (potentially with the help of data engineers).
%Resulting clusterings are visualized and validated to find out whether they are useful for identifying quality problems in the underlying data model.
%Resulting clusterings are visualized and evaluated to find out whether they are useful for identifying quality problems in the underlying data model.
%Otherwise, further iterations of clustering with modified configuration parameters may be needed. 
The approach can be configured according to domain knowledge.
%Domain experts evaluate the computed clustering results.
%The clustering results are evaluated by domain experts.
%\viola{Remove explanation of iterations here?}
To achieve meaningful clusterings, multiple iterations with modified configurations may be needed. 
%To achieve meaningful clusterings, multiple iterations of clustering with modified configuration parameters may be needed. 
Finally,
% when being satisfied with a clustering, 
domain experts interpret the clustering concerning quality problems in the data model.
%\gabi{Modified}
%\viola{ok}

%Overview of our approach:
% model-driven aspects ...
% independence from database technology
% multiple logical/physical data models per conceptual data model
% relevancy for other types of models
Motivated by data quality analysis, we have developed this clustering approach for identifying quality problems in data models.
%	for which representative instances exist. 
As it just takes a list of data values as input, it can also be used to identify quality problems of other kinds of models such as conceptual models~\cite{Moo05} and meta-models~\cite{BV10,LGL14}. %as well. 
While quality properties of meta-models are typically concerned with the form of existing structures, our clustering approach can find out missing structure in meta-models.
Models-at-runtime (cf. \cite{SZ16}), for example,  that store and process data from system logs may show more heterogeneity than intended and may also profit from our clustering approach.
%\gabi{Modified.}
%\viola{ok}
%Heterogeneous logs may reveal quality issues of the corresponding metamodel.
%\todo[inline]{Do we find a set of real models to a given meta-model? Maybe in Remodd?}
% ank[ndigung unseres Hauptergebnisses

% Although quality problems may have different causes concerning with data acquisition, data model, data transformation, we focus on data model problems here as they are often the cause
%in digital humanities, for example, data models change quite often due to new insights
%approach may also be useful for clustering unstructured data that occur in other contexts such as software models. It would indicate meta-model problems then

%What are our achievement?
%conceptual approach based on well-known clustering algorithms, generic w.r.t. kind of data 
%implementation in Python
%evaluation in the context of cultural heritage data
We start our presentation with motivating examples in Sec.~\ref{sec:example}, 
	introduce the concepts of our clustering approach in Sec.~\ref{sec:approach}, 
	give an overview of available tool support in Sec.~\ref{sec:tool}, 
	and present an initial evaluation performed in Sec.~\ref{sec:evaluation}. 
Finally, we discuss related work in Sec.~\ref{sec:related} 
	and conclude in Sec.~\ref{sec:conclusion}.

	%Here maybe a state-of-the-art section on data model quality

\section{Motivational Examples}
\label{sec:example}
%\viola{See notes in tex file}
% TODO: find english data set with valid data field

% 1. data set and origin [organisation, scope, (origin project?)]
% 2. data field(s)  [documentation and resulting expectation]
% 3. existing problems [that we are about to find]
% 4. detecting approach
%GT: sketch several concrete examples that point to different kinds of problems (without details)
%\todo[inline]{I would call this section ``Motivational examples''.}

%\viola{Restructured first two paragraphs}
% database and formats
To motivate our approach with examples, we consider two local databases on cultural heritage objects, such as paintings and buildings.
%capturing meta data on paintings and buildings. 
%They are not publicly available.
\annonymversion
	{They use the data models MIDAS~\cite{bove_marburger_2001_anonym} and LIDO v1.0~\cite{coburn_lido_nodate}.}
	{They use the data models MIDAS~\cite{bove_marburger_2001} and LIDO v1.0~\cite{coburn_lido_nodate}.}
%\arno{Midas does not have a schema}
%As a motivational example, we use a database on cultural heritage objects, such as paintings or buildings, in the LIDO 1.0 format~\cite{coburn_lido_nodate}.
% TODO: db size?
%For demonstration purposes we introduce a running example taken from the LIDO 1.0 schema~\cite{coburn_lido_nodate}.
%\arno{Should we remove 'conceptional' here, as it is different from conceptional models as meant in introduction and \cite{Moo05}}
%\gabi{yes}
MIDAS is described in a manual but not realised as an XML schema.
%\annonymversion
%	{MIDAS is described in a manual~\cite{bove_marburger_2001_anonym} but not realised as an XML schema.}
%	{MIDAS is described in a manual~\cite{bove_marburger_2001} but not realised as an XML schema.}
%The \lstinline|measurementUnit| element will serve as our running example.
The MIDAS data we consider was created by domain experts manually.
LIDO is a CIDOC-CRM~\cite{CIDOCCRM} application and XML schema for harvesting and exchanging metadata of collectibles.
The LIDO data was created via a data transformation applied to data in MIDAS. %a different format, called MIDAS~\cite{bove_marburger_2001}.
%\gabi{This sentence is rather redundant as manual acqusition is standard.}
% TODO: was it created by multiple institutions?
%\gabi{``Data format'', ``data schema'' and ``data model'' are all synonym?}
%\arno{One can argue about that. In my opinion there are differences:
%	Data format describes how data is stored (physical data model, technology).
%	Data model describes how data is structured, which can be technology independent (conceptual data model).
%	Data schema (often database schema) is a technology specific definition of how the data model is realized (logical data model) in a specific data format.
%}

For our running example, we focus on measurement information about cultural heritage objects. 
%and their representation in the data.
% measurements
In LIDO, measurements of objects are expressed with the element \lstinline|measurementsSet|.
It contains the following three elements:
\lstinline|measurementUnit|,
\lstinline|measurementValue| and
\lstinline|measurementType|.
% TODO: visualise model excerpt
%Measurements of resources associated with the objects described by the data are supported via the same three elements, but are not considered here.
The element \lstinline|measurementUnit| 
%, which represents ``the unit of the measurement''~\cite{coburn_lido_nodate}, 
serves as our running example.
According to the LIDO schema, it may contain an arbitrary string value.
%In the documentation it is defined as ``the unit of the measurement'' and 
%The following examples are provided in the documentation: 
The LIDO documentation describes this element as follows.
\emph{``Definition: The unit of the measurement.
How to record: E.g. cm, mm, m, g, kg, kb, Mb or Gb. 
Repeat this element only for language variants.''}~\cite{coburn_lido_nodate}
%The following abbreviations of units for length, weight and digital information are given as examples in the documentation:
%\lstinline|cm, mm, m, g, kg, kb, Mb or Gb|.
%These are abbreviations of units for length, weight and data size measurements.
%\todo{explain shortcuts}
%It should contain the unit of a measurement of (a part of) an object.
%It is a child of the \lstinline|measurementsSet| element, which represents measurements of one aspect of an object.
%The value of the measurement and the kind of measurement taken are represented by the \lstinline|measurementValue| and \lstinline|measuremenType| elements, which are also children of \lstinline|measurementsSet|.
% database
%\todo[inline]{The following paragraph is the right one to start with.}
%As a motivational example, we use a database on cultural heritage objects, such as paintings or buildings, in the LIDO 1.0 format.
%The data was created via a data transformation applied to data in a different format, called MIDAS~\cite{bove_marburger_2001}.
%This original data was created through manual acquisition by domain experts.

We chose a database in LIDO that contains 87,042 \lstinline|measurementUnit| elements with 179 distinct string values overall.
Based on the LIDO documentation, we expect simple indications of the measurement unit only.
%\gabi{Is there a reference to that database?}
%\viola{No. Added ``local'' at beginning of section.}
% such as \lstinline|cm| or \lstinline|mm|.
However, a diversity of values was found.
Examples are \lstinline|-|; \lstinline|-10.5 cm|; \lstinline|x 55 cm|; \lstinline|cm / 120 cm| and \lstinline|? cm|.
%and \lstinline|cm cm|. % TODO: format of values ok?
(Note that in the original data a comma is used as the decimal separator as it is in German.)
%\todo[inline]{I would present English examples to be better understandable.}

%This heterogeneity may have multiple reasons, which may be rooted in the data acquisition, the data models involved or data transformations.

% further examples:
As a second example, we investigated a MIDAS database on cultural heritage objects that includes 118,032 distinct values in the field \lstinline|artist name|.
%\gabi{Mention the data model element.}
%\viola{Added some info from the MIDAS manual below.}
\annonymversion
	{The MIDAS manual~\cite{bove_marburger_2001_anonym} contains verbal descriptions of several rules on how certain information should be expressed in this field.}
	{The MIDAS manual~\cite{bove_marburger_2001} contains verbal descriptions of several rules on how certain information should be expressed in this field.}
%what information should be expressed in what way in this field.
%The field is supposed to hold 
Basically, the first, last and middle name(s) of an artist should be given.
%There are no constraints concerning syntax.
The field is used as an identifier for artists, for example, to relate them to an object.
Thus, the manual lists additional information that can be appended to ensure the uniqueness of the entry.
Furthermore, the manual explains how to encode different types of uncertainty concerning the artist or artist name.
% in the value.
%Uncertainty with respect to the name of an artist is encoded in this field as well.
%Thus, further distinguishing information such as the artist's birth year, are appended often.
%Uncertainty with respect to which artist(s) worked on an object cannot be expressed explicitly in MIDAS.
As the acquisition software based on MIDAS does not ensure these rules, the data found in this field is pretty heterogeneous.
Examples include
\lstinline|Munch, Edvard|;
\lstinline|B., I. C.|;
%\lstinline|Martin, A. (junior)|,
\lstinline|Zindel, Peter (1841)|;
%\lstinline|Zucchi, Carlo (1)|,
%\lstinline|Strafford, Thomas Wentworth (Strafford, Earl)|,
%\lstinline|Wittman, ?|,
\lstinline|Lay?, ? de|;
%\lstinline|Raffael?|,
%\lstinline|Kothe & Winter|,
%\lstinline|Corneille (1922)|,
%\lstinline|Kimpfel, (...)|,
%\lstinline|Georg, Michael & Ammann, Jost|;
%\lstinline|Huberti (Huybrechts), Adriaen|,
%\lstinline|I.F.F.B.|,
%\lstinline|Zwollo, Frans (1872) / Zwollo, Frans (1896)|,
%\lstinline|Maese, Esteban <1550>|,
\lstinline|Walt ..., R.| and
%\lstinline|Stein, Gottfried (1687) (1687)|,
%\lstinline|Bolt, Friedrich >>> Bolt, Johann Friedrich| and
\lstinline|Grass, A. / Graß, Adolf / Grohs, A.|.
%\lstinline|S.*****, C.|.

Furthermore, the database contains 52,523 distinct values in the field for dating objects.
%\gabi{Mention the data model element.}
%\viola{Added some info from the MIDAS manual below.}
%Any string value is allowed.
According to the MIDAS manual, values may contain the indication of a year and, if known, also the indication of a month and a day.
%When the date lies before Christ, this is encoded by appending \lstinline|ante|.
The manual further explains how time spans and different types of uncertainty should be encoded.
%Some types of datings represent time spans instead of points in time.
%The manual explains how both should be expressed.
%These are also expressed in this field.
%Furthermore, the manual contains rules on
%It further states how different types of uncertainty concerning the dating should be encoded.
% verbally or through special characters in the value.
%Any uncertainties concerning the dating are implicitly encoded as well.
%Depending on the information present the field may contain the a year, a month and a day.
Again, as those rules are not ensured automatically, the data of this field is pretty heterogeneous as well.
Examples (adapted to English) include
%In the field for the dating of objects the same database contains 52,523 distinct values, including the following examples (adapted to English):
%\lstinline|1896|,
\lstinline|1895/1902|;
%\lstinline|1908-1909|;
\lstinline|1686.10.24|;
\lstinline|x|;
\lstinline|ca. 1781|;
%\lstinline|ca. 1840-1850|,
%\lstinline|[1993]|,
\lstinline|after 174ante|;
\lstinline|since 1927|;
\lstinline|unknown|;
\lstinline|1900-around 1991|;
\lstinline|1847 (Original 1846)| and
\lstinline|Beginning 20th century|.
%\lstinline|?.03.17| and
%\lstinline|187x.09.21|.
%\lstinline|starting 1067/1069|.

% motivation/problem statement:
As the examples show, data values of a field may be quite heterogeneous, often due to missing structure in the data model.
The heterogeneity implies that the data and the data model may be of low quality.
For example, the understandability and comparability of this data may suffer.
Thus, \emph{quality assurance} is necessary.
Analysing data values manually, however, is a time-consuming task.
%Note that, in the context of cultural heritage data, such a number is actually pretty small.
%It may be much higher for other fields, e.g. the name of a person or the title of an artwork.
%\todo[inline]{The paper would gain from further concrete examples.}
%\viola{Added examples above. How should we format the examples values?}
%\gabi{We should discuss small sections of the data model to understand that they may not be developed enough for the entries found.}
%\viola{done}
%As the examples show, the data values are again very heterogeneous and thus data quality assurance is necessary.
%Note, that for other fields, e.g. name of a person, this number may be much higher.
% suggested solution:
%Since we may not know in advance what kinds of values to expect, pattern-based analysis (via, e.g., regular expressions) is also not feasible.
%\gabi{A pattern-based approach may make sense... I reformulated the corresponding sentence.}
%\viola{To implement the rules, regular expressions would be sufficient. The structural patterns supported by our pattern approach are not needed here. Should we still refer to our paper?}
%A pattern-based approach such as~\cite{kesper2020} can be used to implement all the rules given in the manuals, to find data with quality issues. 
Regular expressions can be used to implement all the rules given in the manuals, to find data with quality issues. 
%The examples above have shown, however, that there is some need to store additional information about cultural objects.
%Thus, a bottom-up 
But further heterogeneity may be present in the data values, often due to the need to store additional information.
%\viola{Reworded sentence above since Markus did not understand the connection to the following sentence.}
%\gabi{ok}
Thus, an explorative approach is needed that supports domain experts in gaining an overview of the %heterogeneous 
data values given and adapting the data model accordingly.

\section{Approach}
\label{sec:approach}
%\viola{This first paragraph is redundant to the one in the introduction. Shorten here or in intro?}
We present a bottom-up approach for supporting domain experts in detecting quality problems in data models.
%The workflow is intended to be usable by domain experts finally.
%Currently, the help of data engineers may be needed.
%and interpreting data quality problems and their causes in the underlying data model.
%The idea is to cluster all data values of a field of interest by syntactic similarity based on domain knowledge. % at a relatively high level of abstraction.
%\gabi{Why is ``high level of abstraction'' important?}
%\viola{I chose this wording in order to distinguish our approach from other approaches that use clustering and edit distances to detect spelling variants. They calculate the distance on the concrete values and thus cluster by syntactic similarity on a very concrete level. In contrast to that, we abstract from certain features (e.g. length of letter sequences) before calculating distances and clustering as we aim to detect groups of syntactically similar values.}
%The kind of clustering we present is based on the assumption that typically syntactically similar data values are also semantically similar.
%Based on the observation that typically semantically similar data values are handled similarly in the data,
%we propose to cluster all data values of a field of interest by syntactic similarity based on domain knowledge.
%We observed that typically semantically similar data values are handled similarly in the data.
%Therefore, 
The idea is to cluster all data values of a field of interest by syntactic similarity.
%based on domain knowledge. \arno{This does not make sense here without further explanation}
%\viola{Added sentence below.}
%Which syntactic features influence the similarity between data values to which degree depends on the data field analysed.
The degree to which certain syntactic features influence the similarity between data values depends on the field analysed.
Thus, domain experts can configure the clustering process according to their domain knowledge.
%The kind of clustering we present is based on the assumption that typically syntactically similar data values are handled similarly in the data model.
%\todo[inline]{Maybe this is too strong. Do we have any reference here? Syntactically similar values are handled similarly in the data model. }
%\viola{Do you mean ``semantically similar values are handled similarly in the data model''?}
%\viola{Removed mentioning of semantic similarity.}
%The clustering approach can be configured by domain experts according to their domain knowledge.
%By presenting clusters to domain experts we enable them to get an understanding of the data values' diversity in syntax and semantics.
%The clustering enables users to get an understanding of the data values' diversity in syntax. %and semantics.
The resulting clustering provides an overview of the data values' diversity in syntax. %and semantics.
%Thus, the goal is to give domain experts an overview of large sets of data values by presenting them clusters of syntactically and potentially semantically similar values.
By interpreting it, domain experts may identify quality problems in the data model.
%These may be caused by the data acquisition, the corresponding data model or a data transformation that produced the data.
%\viola{Removed mentioning of data acquisition and transformations.}
%\gabi{What are the kinds of data quality problem that can be identified with clustering?}
%\viola{Added sentences below. Adopted your formulation from abstract.}
In particular, the clustering can reveal the encoding of diverse information in a field through specific syntax.
This is often caused by missing or non-suitable structure in the data model.
%We will focus on data model problems in this paper.
%\todo[inline]{We will focus on data model problems in this paper.}
%These problems may have one of three causes: the data acquisition, the corresponding data model or a data transformation that produced or modified the data.

%\gabi{Please formulate all open questions throughout the paper to indicate how far this paper can be reasonably categorised as new ideas paper.}
%\viola{Done. Questions have green background.}

%\subsection{Overview of the Approach}\label{sec:workflow} % alternative titles: Vision, Envisioned Workflow
%\gabi{The overview of the workflow is too long. It can be short since each step is explained in detail later on.}
%\arno{added 3 alternative workflow diagrams}
%\gabi{The workflow in Fig. 1 is good.}
%\gabi{Title: Overview of the approach?}

% TODO: diagram rework? GT: or a pseudocode algorithm

\begin{figure*}	
	\centering
	\includegraphics[width=\linewidth]{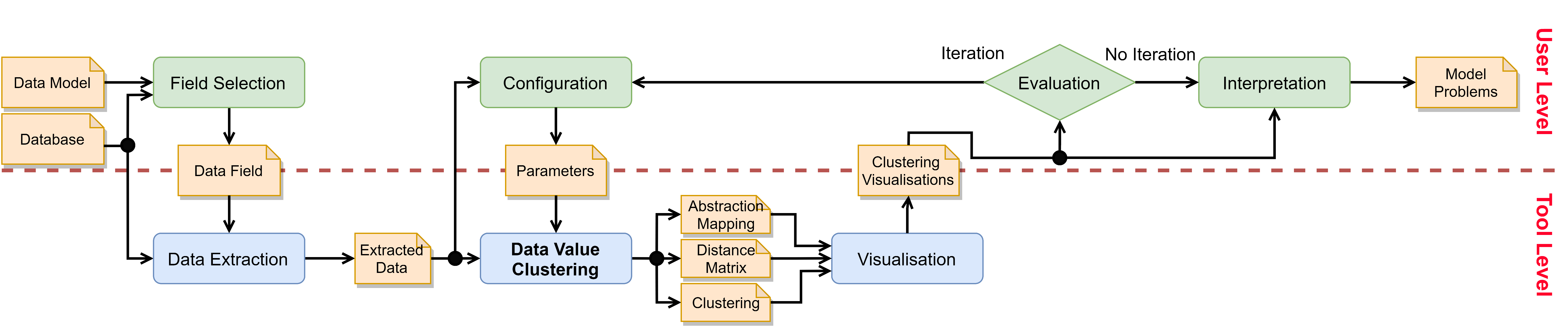}
	\caption{Workflow of the approach to detecting quality problems in data models by clustering heterogeneous data values.}
	\label{fig:workflow}
\end{figure*}

% TODO: GT: What are the input and the output of this workflow? (Configuration settings are also input...) Please explain why the workflow looks as it does. 
% TODO: in jedem paragraph das running example entsprechend erwähnen/diskutieren?
% TODO: Paragraphen vielleicht raus nehmen?
%\viola{Added brief discussion of inputs, outputs and iterativeness.}

The \emph{interactive workflow} of our approach is visualised in Fig.~\ref{fig:workflow}.
It is intended to be usable by \emph{domain experts} who %have domain knowledge and  \arno{This is clear from tthe context.}
are interested in analysing and improving the quality of their data model.
%Currently, the help of data engineers may be needed.
% input: database, chosen field, configuration settings
% output: clustering 
The \emph{inputs} to the workflow are a database and the underlying data model.
%	a database and
%%	the selection of 
%	a field of interest of the underlying data model.
%	domain knowledge concerning the data,
%	configuration settings of the algorithm. 
%	\viola{The latter is not consistent with the diagram. Should we remove it here? Alternatively, we could replace it by ``domain knowledge'' and add Domain Knowledge to the diagram as an input to the Configuration and Interpretation activities?}
%	\arno{I would only list database and field here. Configuration settings are worked out within the workflow and are subject of change. Domain knowledge is too abstract.}
%	an evaluation and interpretation of the produced clustering concerning data quality based on domain knowledge.
%	an assessment of the produced clustering concerning data quality.
%	\gabi{Why is evaluation and interpretation input?}
%	\viola{You are right. It is not an input but an action performed within the workflow. Removed.}
%	\todo[inline]{What does an interpretation specify?}
%	\viola{By interpretation we mean the mapping between clusters and quality problems.}
The \emph{output} of the workflow is 
%	a clustering of the data values of the chosen field and,
%	depending on its interpretation by the domain experts, 
	a list of quality problems of the data model.
%	depending on the interpretation of the calculated clustering by the domain experts.
%
%	data, the data model or related data transformations.
%	\todo[inline]{do you really mean improvements here? Not problems? I doubt that the improvements can be given fully automatically.}
%	\viola{Problems can also not be given fully automatically. We need to discuss whether we view the interpretation of clusters by domain experts and the selection and application of quality improvements as part of our approach.}
%\arno{Add "Quality Problems" and "Improvements" as output to diagram?}
The workflow is \emph{iterative} as it may be necessary to adjust the configuration and re-execute the clustering algorithm several times to achieve a useful clustering.
%\todo[inline]{Who decides whether a clustering is useful? If this is done by the user, the output can only be a clustering.}
%\viola{Yes, this is ultimately done by the user. We need to discuss the inputs and outputs of our approach.}
%The premise is that the domain experts have already chosen one field of the data model that they intend to analyse.
%In some cases one might not want to analyse all values but just those within a specific context.
%The individual steps of the workflow are presented in more detail in the following paragraphs.
The steps of the workflow are presented in more detail below using the running example.

\subsubsection{Field Selection}
%\viola{Should we use the term ``user'' or ``domain expert'' throughout the whole workflow?}
%\gabi{I would use the term ``user'' as it is open whether the user is a domain expert or a data engineer.  One or two sentences would be good that clarify that the workflow is intended to be usable by domain experts finally.  Currently, the help of data engineers may be needed.}
%\arno{Changed. Added sentences in first paragraph.}
Given a database and the underlying data model, domain experts select a data field to be analysed.
For the running example, we used a database on cultural heritage data based on LIDO.
We selected the field \lstinline|measurementUnit|.

\subsubsection{Data Extraction}
% extract all data values from a single data field from a database
%The inputs of the data extraction step are a database and the selection of one field of the underlying data model.
%\arno{
%	Replacement suggestion:
%	The workflow starts by extracting all desired data values from a given database. 
%	Typically this are all data values from a specific data field.
%	However in some cases, the user might want to analyse values from another selection with specific conditions.
%}
%Given a database and a data field of interest, the workflow starts by extracting all corresponding data values from the database.
Next, all data values of the selected field are extracted from the database automatically.
%In some cases, the users might not want to analyse all the values but just those within a specified context. %range?
%For the running example, we extracted all the values of the field \lstinline|measurementUnit| from the database.
%of a cultural heritage database.
%The extraction typically requires a simple database query to be executed.
% TODO: How do we envision support for domain experts to perform this task?
The result is a list of data values; 
note that the remaining part of the workflow is independent of the database technology used.

\subsubsection{Configuration}\label{par:config}
% completely configurate clustering algorithm 
% includes configuration of compression, distance function, algorithm parameters
%The next step is to configure the data value clustering.
%The data value clustering is configurable.
Based on the observation that ``incorporating domain expert input often improves clustering
performance''~\cite{CCC+17}, the data value clustering process is configured by domain experts.
%\gabi{Syntax des Zitats wirkt ein bischen falsch.}
%\viola{Checked. It is correct.}
%In the next step, the users configure the data value clustering process.
%For the configuration of the data value clustering, 
%This means that their 
For this purpose, their domain knowledge has to be mapped to parameter values.
%in a suitable way.
%The configuration step is based on the observation that ``incorporating domain expert input often improves clustering
%performance''~\cite{CCC+17}.
%\viola{Added reference above.}
%This means that the domain knowledge given by domain experts is mapped to parameter values.
%This means, the parameter values are filled based on the experts' domain knowledge.
%This means, the domain experts specify parameters based on their domain knowledge.
%about the data, such as details about the acquisition process or the data model.
%Hence, knowledge and experience concerning the data must be mapped to parameter values.
%This knowledge may, for example, include details about the data model or the data acquisition process.
%This knowledge may include, for example, details about the data creation and the data model.
Considering the running example on measurement units, the domain knowledge includes, for example, the fact that small sequences of letters are expected.
%Further details on the configuration are provided in the following paragraph.
The configuration facilities are further explained in the following paragraph.
% while digits and special characters are not expected.
%\todo[inline]{Make this concrete at the running example.}
%\viola{Replaced abstract example by concrete example.}
%This mapping could be supported by a questionary.
%Through the configuration the experts basically specify which syntactical features of the data values are unexpected and could have a significant impact on the values' meaning, thus should impact the clustering.
%are of interest for the selected field and thus should impact the clustering.
%\gabi{Here an overview of configuration facilities for clustering.}
%\viola{Added sentence above. As the three steps of the algorithm are explained not until Sec.~\ref{sec:algorithm}, we cannot really give an overview of configuration facilities here.}
%The configuration facilities of the algorithm are discussed in Section~\ref{sec:algorithm} in more detail.
%The configuration facilities of the algorithm and our configuration for the running example
%, which we developed based on our domain knowledge, 
%are discussed in Section~\ref{sec:algorithm} in more detail.
%For the running example, we configured the algorithm based on our domain knowledge.
%Per step, we also explain our configuration for the running example on measurement units
%%, which we specified 
%based on our domain knowledge.
%Details on the 
%The development process of this configuration is outlined in Sec.~\ref{sec:evaluation_setup}.
%the following subsections in more detail.
%\todo{Where?}

\subsubsection{Data Value Clustering}\label{par:clustering}
%\viola{Shortened this paragraph.}
%\viola{Ideas for alternative name?}
%\gabi{Here a short description of this step first?}
%\viola{Added below.}
%The data value clustering comprises the following three steps: an abstraction from the data values, a calculation of pairwise distances between abstracted values and 
%Based on the configuration, the data values are clustered.
%The clustering of data values is performed in three steps:
%%consisting of three steps:
%	(1) the data values are abstracted, 
%	(2) their pairwise distances are calculated and 
%	(3) they are clustered by syntactic similarity.
%\gabi{Why are data values abstracted first?}
%\viola{Modified.}
%
%Based on the configuration, the data values are clustered.
%The inputs for clustering are the extracted data and the configured parameter values.
%Before clustering, we abstract from the original data values to remove irrelevant syntactical details.
%%Before clustering we group the values by removing irrelevant syntactical details.
%This results in a first grouping of syntactically similar values.
%After calculating distances between these groups, the clustering is performed.
%The output of this step is a clustering of values that serve as representatives of these groups.
%The output of this step is a clustering of representative values of these groups.
%%a set of representative values, which represent groups of syntactically similar values.
%
%Based on the configuration, the data values are clustered.
%
%The inputs are the extracted data and the configured parameter values.
%As explained in the following, 
The data value clustering is performed in three steps as outlined in the following.
\shortversion
{\annonymversion
	{Details and examples are explained in \cite{supplementary_material_anonymous}.}
	{Details and examples are explained in \cite{supplementary_material}.}}
{Details and examples are explained in Section~\ref{appendix:clustering} of the appendix.}
%
%\viola{TODO: remove all references to appendix if we cannot submit supplementary material}
%\viola{Add reference to tool support?}
%\viola{Integrated highlights from previous 3B below.}
Since our goal is to provide an overview of significant differences in the syntax of the data values, the first step is
%First, 
an \emph{abstraction}.
% from the original values.
Thereby, syntactic features that, according to the configuration by domain experts, are irrelevant for clustering are removed.
For example, we can abstract from concrete characters of a specific group of similar characters, such as letters, or from the length of certain character sequences.
%, such as digit sequences.
For the measurement units, for example, we abstract from the length of digit sequences.
The configuration depends on expectations about the syntax of the data values
% of interest.
%It further depends on 
and the kinds of syntax variations of the values that cause significant variations in their meaning.
%The abstraction is a first grouping of similar values as each abstracted value represents a set of original values.
%Thus, this is a first grouping of similar values.
Ultimately, the original data values are mapped to a smaller set of shorter values via a set of abstraction rules determined by the configuration.
The result is the abstraction mapping.
The abstraction step is a first grouping of similar values since each abstracted value represents a set of original values.
%The configuration determines which abstraction rules are applied
%Ultimately, the configuration determines which abstraction rules are applied.
%Ultimately, the abstract produces an abstraction mapping between original and abstracted values.
In the running example, 179 values given originally are abstracted to 22 values.
%this abstraction mapped the 179 distinct values given originally to 22 abstracted values.
%Table~\ref{tab:mapping} shows an excerpt of the resulting abstraction mapping.
The coloured boxes in Fig.~\ref{fig:clustering} %~\ref{tab:mapping} 
%The columns of Table~\ref{tab:mapping} 
represent groups of original values that were mapped to the same abstracted value.

If the abstraction does not produce a manageable amount of groups,
%If the abstraction produces a manageable amount of values, 
%this may be sufficient to provide an overview of the values' diverse syntax.
%This is especially the case when the abstraction produces a manageable amount of values.
%Otherwise, 
our approach suggests clustering the abstracted values.
%to produce a manageable number of clusters containing similar abstracted values.
%\gabi{This should come earlier. Actually it is a nice motivation for the clustering (which needs distance definition before).}
As a prerequisite, pairwise \emph{distances} (i.e., dissimilarities) between abstracted data values have to be computed.
%again depending on domain knowledge.
%, to quantify the amount of dissimilarity.
%For example, special characters are often used to encode special meaning and thus should cause high dissimilarity.
%String dissimilarity is typically measured via {\em edit distances} allowing different kinds of string operations~\cite{Navarro01}.
Having applied the approach to cultural heritage data, we achieved reasonable results with
the basic {\em edit distance} that allows insertions and deletions of string characters only, 
and the Levenshtein distance~\cite{levenshtein1966}, which additionally allows substitutions.
%They may be accomplished with further distances when needed.
The dissimilarity between two values %that differ by insertion, deletion, substitution or transposition of certain characters 
depends on the data field analysed.
%In a field representing the name of a person, for example, the insertion of an additional letter does not really cause dissimilarity whereas for a field representing the height of an object it does.
Therefore, we use configurable weights %(i.e. costs) 
for each edit operation (namely insertion, deletion and substitution) applied to each possible character.
Domain experts configure the weights based on their domain knowledge.
%The chosen weights for the running example are presented in Table~\ref{tab:weights}.
%\begin{table}
%	\renewcommand{\arraystretch}{1.3}
%	\caption{Distance Weight Matrix for measurement units}
%	\label{tab:weights}
%	\centering
%	\begin{tabular}{l|cccc}
%		& \multicolumn{1}{l}{-}    & \multicolumn{1}{l}{Digits} & \multicolumn{1}{l}{Letters} & \multicolumn{1}{l}{Special} \\ \hline
%		-       & -                        & {\color[HTML]{656565} 2}   & {\color[HTML]{656565} 1}    & {\color[HTML]{656565} 2} \\
%		Digits  & {\color[HTML]{656565} 2} & 0                          & {\color[HTML]{656565} 3}    & {\color[HTML]{656565} 4} \\
%		Letters & {\color[HTML]{656565} 1} & {\color[HTML]{656565} 3}   & 1                           & {\color[HTML]{656565} 3} \\
%		Special    & {\color[HTML]{656565} 2} & {\color[HTML]{656565} 4}   & {\color[HTML]{656565} 3}    & {2}              
%	\end{tabular}
%\end{table}
%The first column and row contain the weights for character deletions and insertions, respectively.
%The other cells show the weights for substitutions of corresponding characters.
%Typically, vague relations between weights can be derived from domain knowledge.
%The configuration of the weights is based on the same aspects as the configuration of the abstraction.
In general, edit operations of unexpected characters and operations that may have a significant impact on the values' meaning should be weighted high as they may indicate quality problems.
%in the data model.
For the measurement units, we therefore chose the weight for inserting letters lower than those for digits and special characters.
%Therefore, the weight for inserting or deleting letters is lower than those for digits and special characters, for example.
%\gabi{Why?}
%\viola{Added reason.}
%Additionally, the more influence edit operations of certain characters have on the meaning of data values, the higher those weights should be.
%the weight of those characters should be.
%, such as ``A should be weighted much higher than B''.
%However, there remains some leeway in determining concrete values that satisfy these relations.
%Therefore, we support multiple iterations to experiment with different configurations.
%The other cells show the weights for substitutions of corresponding characters.
%Digits and special characters are unexpected, while letters are expected. 
%Therefore, the weight for letters is lower than those for digits and special characters.
%
%First, irrelevant syntactical details are removed from the original data values, resulting in abstracted data values.
%Since each abstracted value represents a set of original values, this is a first grouping of syntactically similar values.
%Second, pairwise distances between the abstracted values are calculated.
%Ultimately, the distance matrix containing pairwise distances between abstracted values is calculated.

Finally, the abstracted values are \emph{clustered} by syntactic similarity based on the calculated distance matrix.
Only clustering algorithms that can operate on string distances can be applied,
%We can only use clustering algorithms that operate on string distances.
such as hierarchical clustering~\cite{hierarchical}, k-medoids~\cite{kmedoids} and DBSCAN~\cite{dbscan}.
%The set of suitable algorithms includes hierarchical clustering~\cite{hierarchical}, k-medoids~\cite{kmedoids}, DBSCAN~\cite{dbscan}, OPTICS~\cite{optics}, affinity propagation~\cite{affinity}, and spectral clustering~\cite{spectral}.
Domain experts have to select an algorithm and configure its parameters dependent on the data field analysed.
We provide a setting that allows experimenting with a variety of clustering algorithms.
For the running example, we chose hierarchical clustering.
It is up to future research to investigate which clustering algorithms and parameter settings are most suitable %in the context of our approach 
for detecting quality problems in data models
and how this depends on the kind of data considered.

\subsubsection{Visualisation}\label{par:visualisation}
\begin{figure}	
	\centering
	\includegraphics[width=\linewidth]{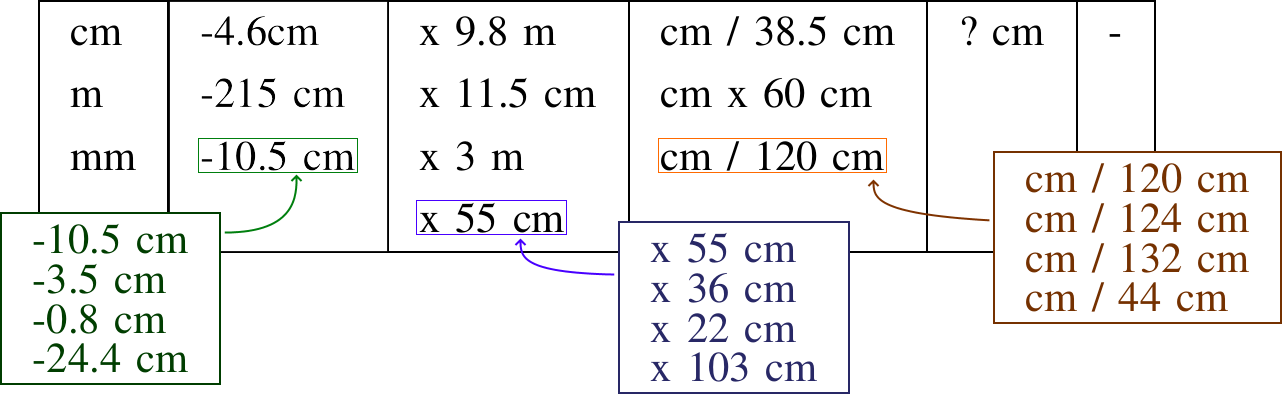}
	\caption{Excerpt from clustering of measurement unit values showing representatives. Excerpt of represented original values shown in boxes.}
	\label{fig:clustering}
\end{figure}

For an overview, the clustering of the abstracted data values is presented.
Per abstracted value, a corresponding original value is shown as a representative.
%\gabi{As abstraction is not explained here,  do not mention it here. }
%\viola{After modifications in Sec.~\ref{par:clustering}, this is valid here.}
%for all original values that were mapped to the abstracted value.
% via original values as representatives.
%For each abstracted value, a corresponding original value is shown, which stands for all original values that were mapped to the same abstracted value.
%Each representative value is an original value and it stands for all original values that were mapped to the same abstracted value.
%Table~\ref{tab:clustering} 
Fig.~\ref{fig:clustering} shows an excerpt of this representation for the running example.
% on measurement units.
Each column of that table represents a cluster.
Six out of ten clusters are shown.
%\begin{table}
%	\renewcommand{\arraystretch}{1.3}
%	\caption{Excerpt from clustering of measurement unit values showing compressed values.}
%	\label{tab:clustering}
%	\centering
%	\begin{tabular}{ |l|l|l|l|l|l|l| } 
%		\hline
%		cm & -2 cm & x 1 cm & cm / 1 cm & ? cm & - & ... \\ 
%		m &  -1 cm & x 2 cm & cm x 1 cm &  &  &  \\
%		mm &  -2cm & x 1 m & cm / 2 cm &  &  &  \\
%		&  &  x 2 m &  &  &  & \\
%		\hline
%	\end{tabular}
%\end{table}
%\begin{table}
%	\renewcommand{\arraystretch}{1.3}
%	\caption{Excerpt from clustering of abstracted measurement unit values showing original values as representatives. Some }
%	\label{tab:clustering}
%	\centering
%	\begin{tabular}{ |l|l|l|l|l|l| } 
%		\hline
%%		\rotatebox[origin=c]{90}{Cluster 1} & Cl. 2 & Cl. 3 & Cl. 4 & Cl. 5 & Cl. 6 & ... \\ \hline
%		cm & -10.5 cm & x 55 cm & cm / 120 cm & ? cm & -  \\ 
%		m &  -215 cm & x 11.5 cm & cm x 60 cm &  &    \\
%		mm &  -4.6cm & x 3 m & cm / 38.5 cm &  &    \\
%%		&  &  x 9.8 m &  &  &   \\
%		\hline
%	\end{tabular}
%\end{table}
%
%
%
%For further exploring the clustering, the mapping between original and compressed values may be shown additionally.
For further exploring the clustering, the mapping between original and representative values is shown, as by the coloured boxes in Fig.~\ref{fig:clustering}.
%. %as done in Fig.~\ref{fig:clustering}.
%The coloured boxes in Fig.~\ref{fig:clustering} %Table~\ref{tab:mapping} 
%show excerpts from the abstraction mapping. % example mapping.

\begin{figure}	
	\centering
	\includegraphics[width=\linewidth]{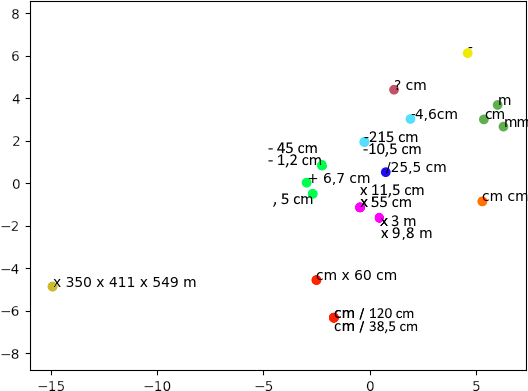}
	\caption{Scatter plot of clustered measurement unit values resulting from multidimensional scaling. Note that a comma is used as the decimal separator.}
	\label{fig:mds}
\end{figure}

%\gabi{This table just show a few original values. How are they selected? Or are they just an excerpt?}
%\viola{It is an excerpt. Modified sentence.}
%In Table~\ref{tab:mapping} the first line contains representative values from Table~\ref{tab:clustering}.
%Some of the original values related to the representative value are listed in the corresponding column below.
%\gabi{Also in that table the dots are confusing.}
%\begin{table}
%	\renewcommand{\arraystretch}{1.3}
%	\caption{Excerpt from mapping between representative and original measurement unit values.}
%	\label{tab:mapping}
%	\centering
%	\begin{tabular}{ |l|l|l|l| } 
%		\hline
%		-2 cm & x 1 cm & cm / 1 cm & ... \\
%		\hline
%		-10.5 cm & x 55 cm & cm / 120 cm & ... \\
%		-3.5 cm & x 36 cm & cm / 124 cm & ... \\
%		-0.8 cm & x 22 cm & cm / 132 cm & ... \\
%		-24.4 cm & x 103 cm & cm / 44 cm & ... \\
%		... & ... & ... & ... \\ 
%		\hline
%	\end{tabular}
%\end{table}
%\begin{table}
%	\renewcommand{\arraystretch}{1.3}
%	\caption{Excerpt from mapping between representative values (first line) and original measurement unit values (below)}
%	\label{tab:mapping}
%	\centering
%	\begin{tabular}{ |l|l|l| } 
%		\hline
%		-10.5 cm & x 55 cm & cm / 120 cm  \\
%		\hline
%		-3.5 cm & x 36 cm & cm / 124 cm  \\
%		-0.8 cm & x 22 cm & cm / 132 cm  \\
%		-24.4 cm & x 103 cm & cm / 44 cm  \\
%%		... & ... & ...  \\ 
%		\hline
%	\end{tabular}
%\end{table}
%
Moreover, we apply multidimensional scaling~\cite{kruskal1964} to 
present the data values in a two-dimensional Cartesian space based on their distances. %space using the distance matrix.
%translate all the pairwise distances between the data values to points in a two-dimensional Cartesian space.
%that the distances between values can be visualised.
A corresponding scatter plot of the running example is shown in Fig.~\ref{fig:mds}.
Each dot is labelled with a representative value.
%Each dot stands for an abstracted value. \gabi{Each dot is labelled with a representative value? Shorten the next sentence.}
%The labels correspond to representative values, most of which are also shown in Table~\ref{tab:clustering}.
%a representative value.
%\viola{Use representatives to annotate points in MDS scatter plot.}
%The compressed values are used for the annotation of points.
Each cluster is represented by a different colour.
This visualisation allows domain experts to get a quick impression of key properties of a clustering:
a clustering is {\em compact} if there is a high similarity within clusters %\cite{LiuLXGW10}.
% A clustering is 
and it is {\em separate} if there is a low similarity between clusters \cite{LiuLXGW10}. 
%Both properties are key. %desired. 
%\gabi{A reference for these two properties is missing.}
%\viola{Added}
%relations between inter and intra cluster distances.
%In general, high compactness and separation of clusters are desired
%Note that some of the points overlap completely. 
%We will consider the reason for this effect in Sec.~\ref{sec:distance}.
%The points are annotated via 
%\todo{in addition?}
%, which allows users to quickly get an overview of the clustering.
%Additionally, it should allow the exploration of the clusters in more detail if desired.
% TODO: Listen, Anzahlen
%\arno{end replace suggestion}
%For both key properties, compactness and separation of clusters, internal cluster validation scores are calculated and shown to the user.
%Both key properties, compactness and separation of clusters, 
%These properties are %further 
%measured by internal cluster validation scores such as the Dunn index~\cite{dunn1974}.
%\viola{The scores typically combine both properties. I am not sure if the above sentence expresses that.}
%Furthermore, internal cluster validation scores, which measure the compactness and separation of clusters, are calculated and shown to the user.
%The Dunn index~\cite{dunn1974},  for example,  divides the minimum inter-cluster distance by the maximum intra-cluster distance.
%\gabi{To measure what?}
%\viola{Removed sentence as it is not important. Reworded sentence above.}
%They are discussed in more detail in the following paragraph.
%\todo[inline]{Also here, a concrete example is needed.}
%\viola{Added Dunn index as example. Is that what you meant?}
%\gabi{This example is too technical here. It should occur when the details of the algorithm are explained. The overview should be well understandable for domain experts, the next subsection is for computer scientists.}
%\viola{Added explanation of scatter plot and Dunn index.}
\question{Empirical evaluation: How should clusterings be visualised to maximize usability?}

\subsubsection{Evaluation}\label{par:eval}
% give user feedback about result of clustering
% automatic metrics etc. concerning the quality of the clustering
% show questionary to find out how satisfied user is with clustering
% if user chooses iterative modification then jump to configuration step with helpful hints

To achieve a useful clustering that enables the detection of quality problems, it may be necessary to perform several iterations with modified configurations.
Thus, the quality of the clustering must be analysed to decide
%to decide
%The evaluation step basically consists of deciding
whether another iteration is necessary. %or continue with the current clustering.
%and identifying possible improvements of the configuration.
% to achieve a more useful clustering.
%there is a need to support domain experts in validating the clustering and identifying modifications of the configuration to achieve a more useful clustering.
%For this purpose, the domain experts familiarise themselves with the clustering by means of visualisations and scores.
%Compression results and calculated distances between values can be examined as well.
% TODO: wie genau?
% clustering validation techniques
%\viola{Moved discussion of internal clustering validation here.}
In general, the quality of clusterings can be evaluated by
%There are two ways to evaluate the quality of cluster analysis results: 
\emph{internal} and \emph{external} clustering validation \cite{LiuLXGW10}.
Internal validation means evaluating the quality of the clustering without the use of any external information.
It does not evaluate the quality of the clustering in the intended usage scenario.
%This, however, is required to answer the research question.
%
Therefore, external clustering validation takes external information into account \cite{LiuLXGW10}.
%In our case, this information consists of an assessment of the clustering by domain experts.
%In our case, this information consists of the users' evaluation of the clustering based on their domain knowledge.
In our case, this means that domain experts evaluate the clustering based on their domain knowledge.
%with respect to how far the clustering reveals quality problems. 
%For that purpose, the users familiarise themselves with the clustering by means of visualisations.
For each cluster, they should assess what kinds of values it includes, whether the grouping of these values makes sense, and how useful it is for the detection of quality problems.
%The key question is whether the clustering brings new insights.
%Part of the evaluation is also checking whether the expert's domain knowledge was adequately translated into parameters.
%If the clustering does not make sense, 
%If that is not the case, 
If the clustering does not bring new insights, the experts' domain knowledge may not have been adequately translated into parameters.
%\viola{Discuss iteration in detail?}
%\viola{Added above sentence. OK?}
%\gabi{How is that done?}
%\viola{Modified.}
%We consider the clustering of the measurement unit values partly visualised in Table~\ref{tab:clustering} as useful since it gives a good overview of the groups of values that differ in syntax and semantics and may have different causes.
%\viola{Move below mentioning of questionnaire to future work?}
%We envision a questionnaire containing concrete questions concerning the expert's satisfaction with the clustering.
%The idea is to use the answers to provide improvement tips during the configuration phase of the next iteration.
%The idea is to provide improvement tips during the configuration phase of the next iteration. 
%The idea is to generate tips for improvements of the configuration based on the answers.
The result of the evaluation is the decision whether another iteration is necessary.
%\viola{Added sentence below.}
Note that an additional iteration does not always improve the clustering.
%\viola{Added sentence below.}
If most clusters are considered useful and just a few seem internally heterogeneous, it may be profitable to cluster the values contained in the problematic clusters separately with a modified configuration.
%\viola{Tried to explain how homogeneity manifests below. OK?}
If ultimately the domain knowledge is appropriately translated into parameter values but the clustering still does not reveal significant differences in the data values, they may be homogeneous.
%If the clustering does not reveal significant differences in the data values after correcting the translation of domain knowledge into parameter values, this indicates that the data is homogeneous.
%If the clustering still seems arbitrary and useless after correcting the translation of domain knowledge into parameter values, this indicates that the data is homogeneous.
%If after multiple iterations of correcting the translation of domain knowledge into parameter values, the clustering still seems arbitrary, i.e. the grouping seems to be based on minor differences in the data values, this may indicate that the data is homogeneous.
%If clusterings of multiple iterations seem arbitrary or reveal only minor differences in the data values, this may indicate that the data is homogeneous.
\question{How can we further support domain experts in evaluating the clustering?}
\question{How can we support domain experts in making adjustments to the configuration in the next iteration?}

%\gabi{Indicate explicitly that the evaluation of the example starts here.}
%Concerning the internal validation of the clustering of the \emph{running example}, Fig.~\ref{fig:mds} does not raise major concerns regarding compactness and separation. % of the clustering.
%Fig.~\ref{fig:mds} can be used for the internal validation of the clustering in our \emph{running example}.
%This visualisation does not raise major concerns regarding compactness and separation. % of the clustering.
%However, it shows that the compactness of cluster four (which is coloured turquoise) is relatively low.
%\gabi{After the iteration indicated above? The example at the beginning of the evaluation paragraph should come here.}
%\viola{Restructured discussion of example. See below.}
%we conclude that the compactness and separation of the clustering of the measurement unit values is ok.

%\arno{This paragraph is really excessive. Can we remove the classification of clusters?}
%\viola{Shortened.}
In the following, we sketch an external evaluation of the clustering in our \emph{running example} (Fig.~\ref{fig:clustering}) based on several interviews with domain experts. 
%We consider the visualisation in Fig.~\ref{fig:clustering}. % Table~\ref{tab:clustering}.
%, which is partly visualised in Table~\ref{tab:clustering}.
The first cluster contains the expected values.
%The experts agreed that their grouping into a single cluster makes sense.
%The second cluster contains values starting with a minus, followed by a measurement value and a unit.
The minus in the values of the second cluster %, according to the experts, 
indicates measurement values in form of intervals in the original MIDAS data. 
%in the data originally stored in MIDAS, 
%in MIDAS measurements are sometimes expressed in form of an interval with the unit appended.
%The values of the second cluster imply that in the original MIDAS data the measurement value was given in form of an interval with the unit appended.
%During the transformation to LIDO the 
%This indicates the lack of options to express measurement values in form of intervals and an incorrect splitting of the value in the data transformation.
%The values in the third cluster start with an ``x'' followed by a measurement value and a unit.
The ``x'' in the values of the third cluster indicates measurements of multiple dimensions.
%They could indicate measurements of multiple dimensions.
The values in the fourth cluster 
%includes values starting with a measurement unit, followed by a separator (a slash or ``x''), followed again by a measurement value and a unit.
%The values in the fourth cluster probably 
%They 
%probably represent
imply
%the second part of an indication of
measurements of multiple dimensions or alternative measurements of the same dimension that are taken from different sources.
%\gabi{What kind of adoption?}
%\viola{Reworded. We meant ``übernommen''.}
%Hence, the last three clusters group unexpected values with similar structures, meaning and causes and thus, are useful for quality assurance.
%Cluster five and six contain a single value each that significantly differs from the other values and encode uncertainty, more specifically uncertain or missing information concerning the measurement unit.
%According to the experts, 
%The last two clusters of Table~\ref{tab:clustering} 
Cluster five and six represent values that encode different types of uncertainty namely doubtful or missing information. %concerning the measurement unit.
%Presenting these outliers in separate clusters makes sense.
%, thus can be considered an outlier.
%helps tracking down the cause for these unexpected values.
%The question mark in cluster five indicates uncertainty.
%In summary, this clustering helps in investigating possible causes in the data acquisition process, related data models and data transformations.
%
The domain experts considered the clustering useful in general since it gives a systematic overview of the groups of values that differ in syntax, semantics and cause. 
But as indicated by Fig.~\ref{fig:mds},
cluster four (which is coloured red) is not very compact and the included characters ``x'' and slash may have significantly different meanings.
%imply a significant difference in meaning.
%some of the included values contain a slash, whereas others contain an ``x'', which may imply a significant difference in meaning.
Hence, we might consider a clustering more useful where cluster four is split up correspondingly.
%If we assume that these characters represent significantly different meanings,
%we might consider the clustering more useful if this cluster is split up.
To achieve this, we would need another iteration with a modified configuration.
%Subsequently, the clustering seems useful enough.
%Thus we move on to the interpretation step.
%Note that the additional iteration in this simple example is not profitable since the small modification of the clustering could easily be done manually.
Admittedly, additional iterations are more profitable if the induced change in the clustering is of greater magnitude.
%such minor optimisations may not be worth another iteration.
%a larger quantities of values are affected, i.e. 
%We would have to modify the configuration such that minus is treated more differently from an x and a whitespace.
%\todo[inline]{Ideally, the need for further iterations can also be illustrated at the concrete example.}

%\gabi{not a clear story line.  First validate a computed cluster, internally and/or externally, then find out that the clustering could be better, then do another iteration and conclude that it is good enough.}
%\viola{Restructured discussion of example.}
%\viola{Added argumentation that example is very simple but additional iterations may still be profitable in other cases.}
%\viola{I am afraid that the example is too simple. Doing another iteration just to extract a single value from a cluster is not profitable. We could easily do that manually. Nevertheless, I think this simple example might still illustrate the need for further iterations.}

\subsubsection{Interpretation}\label{par:interpretation}
% if user is satisfied with clustering then give suggestions for improvements (refactorings, ...)
% not worked out yet!
%When the domain experts are satisfied with the clustering, 
Ultimately, the domain experts interpret the clustering concerning data model quality.
%they interpret it concerning data model quality.
%\gabi{...concerning data model quality?}
%\viola{Yes, modified.}
%before deciding which quality improvement steps to take.
They should evaluate what quality problems of the data values the clustering reveals and what may be their causes in the data model.
\question{How can we further support domain experts in interpreting the clustering? Provide a list of quality problems and improvement suggestions based on expert interviews or machine learning ...}

In the following, we outline the interpretation of our \emph{example clustering} %of the measurement unit values,
(Fig.~\ref{fig:clustering})
by domain experts. 
%Remember, it is partly visualised in Fig.~\ref{fig:clustering}. % Table~\ref{tab:clustering}.
%The domain experts expected to find pure indications of measurement units only.
They recognised the problem that, besides the expected units, the data values also include measurement values and special characters.
% encoding special meanings.
%First of all, 
%They attributed this to the allowing of arbitrary string values.
They inferred that values of an enumeration only should be allowed.
%Thus, the clustering shows that the allowing of arbitrary string values as entries causes problems.
%it is problematic that the LIDO 1.0 schema allows arbitrary string values as measurement units.
%the syntax of measurement unit values should be more strictly constrained (e.g. to prevent digits).
%Alternatively,
%Values of an enumeration only should be allowed.
%One expert suggested alternatively adding syntax constraints depending on values of related fields such as \lstinline|measurementType|.
%One expert additionally suggested to reconsider the field \lstinline|measurementValue| with respect to syntax constraints.
Based on the second cluster, the experts identified the need to reconsider the representation of intervals as measurement values in LIDO.
%The minus sign in the values of 
%The second cluster 
%%implies that in the original MIDAS data the measurement value was given in form of an interval with the unit appended.
%%During the transformation to LIDO the 
%%This 
%indicates a lack of options to express measurement values in form of intervals in LIDO.
%Data model quality could be improved by supporting this explicitly.
%It further hints at an incorrect splitting of the value by the data transformation, as do clusters three and four.
%The ``x'' in the values of 
Clusters three and four imply that support for measurements of multiple dimensions and alternative measurements of the same dimension should be investigated.
%\viola{Added mentioning of the data transformation below.}
Also, the documentation and the data transformation must be checked correspondingly.
%Clusters three and four indicate the measurement of multiple dimensions and alternative measurements of the same dimension.
%The third cluster indicates the measurement of multiple dimensions.
%The fourth cluster indicates alternative measurements of the same dimensions.
%The values in the fourth cluster probably represent the second part of an indication of multiple measurements of the same or different dimensions each followed by the measurement unit.
%Both issues point to the problem that the LIDO documentation does not clearly state how to represent multiple (alternative) measurements and thus should be reworked.
Clusters five and six indicate the need to support uncertain information explicitly.
%\viola{Added summarising sentence below.}
In sum, the clustering helped the experts to identify several quality problems in LIDO. 

\section{Tool Support}
\label{sec:tool}
%\viola{Move this section behind evaluation section?}
%\gabi{I would leave it here as there has to be a tool to evaluate.}
% technology used: python packages
% API?
% our visions concerning the GUI for domain experts?
\begin{figure}	
	\centering
	\includegraphics[width=\linewidth]{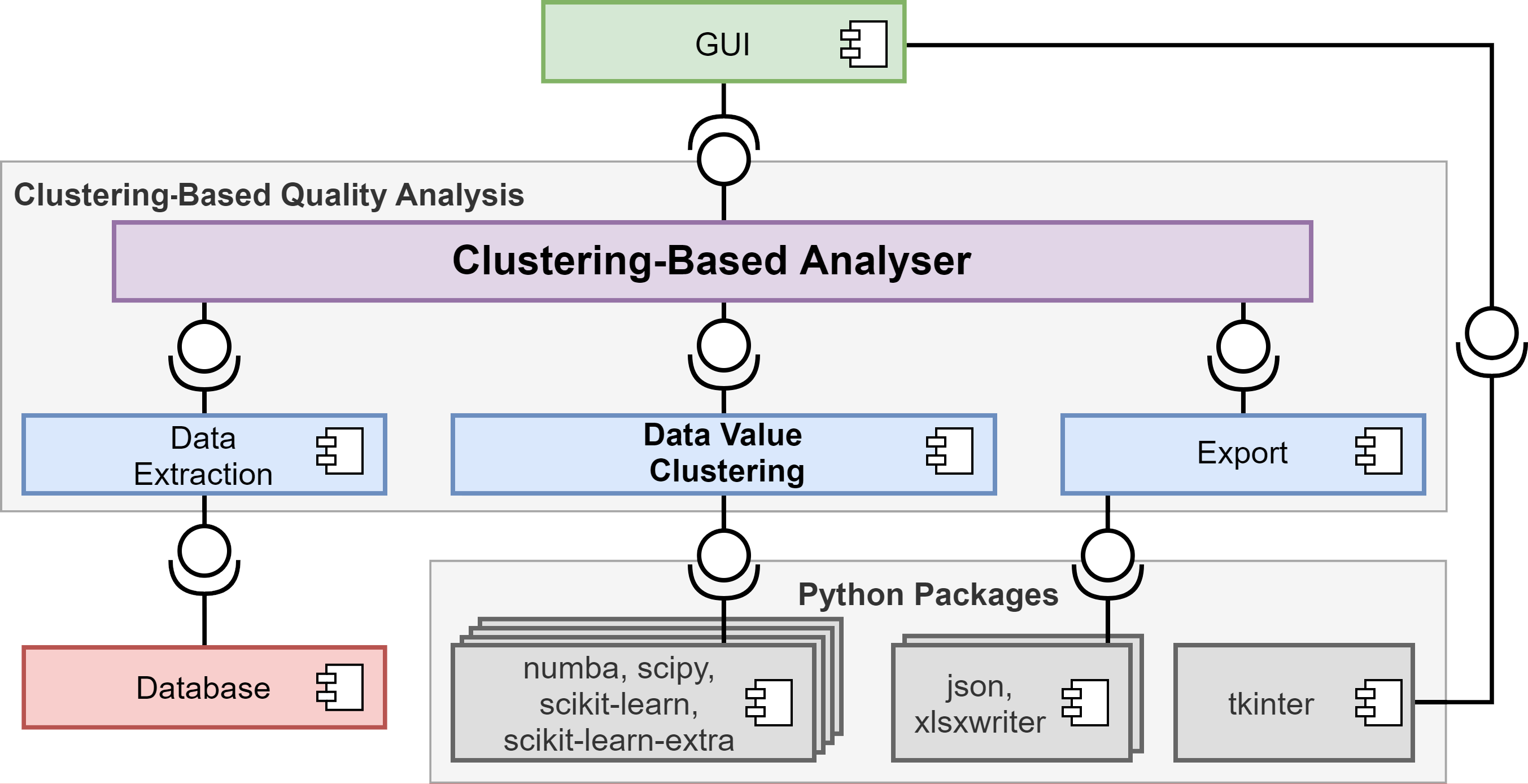}
	\caption{Component diagram of the implementation.}
	\label{fig:components}
\end{figure}
%\gabi{Clustering-Based Analyser}
%
In the following, we report on a proof-of-concept implementation of our approach~\cite{implementation}. % which is still under development.
%\arno{add link to Git repository - here?}
%\viola{done}
%\gabi{What is the overall architecture of the tooling? }
%\arno{Added component diagram. Restructure section accordingly? (currently Backend $<>$ Frontend)}
Fig.~\ref{fig:components} shows the tool architecture. % explained in the following.
%\gabi{Where explained?}
The \lstinline|Clustering-Based Analyser| realises the workflow presented in Sec.~\ref{sec:approach}.
%For this, it operates all components including \lstinline|Data Value Clustering| and GUI by controlling the data transfer between them.
%It operates on several components.
%In particular, 
It controls the data transfer between the %\lstinline|Data Value Clustering| components and the \lstinline|GUI| 
other components.
%These are discussed in the following subsections in detail.
%It further controls the \lstinline|Data Extraction| and the \lstinline|Export| components.
The data values are queried from a database in the component \lstinline|Data Extraction|.
%according to the user's choice.
The component \lstinline|Export| allows saving all parameters and the clustering in JSON format and creates a representation of the clustering as an Excel file.
%\gabi{which results? the clustering? what more? Please make that explicit here.}
%\viola{Modified.}
The components \lstinline|Data Value Clustering| and \lstinline|GUI| are discussed in the following subsections. % in detail.
%In the following subsections we will discuss the \lstinline|Data Value Clustering| component and the graphical user interface (GUI) in detail.
%In the following subsections we will discuss the \lstinline|Data Value Clustering| component and the graphical user interface (GUI) in detail.
%The \lstinline|Clustering-based Analyser| realises the workflow presented in Sec.~\ref{sec:workflow}, operates both of these components and controls the data flow between their subcomponents.
%%Clustering results can be exported to JSON or Excel via the \lstinline|Export| component.
%\gabi{I do not understand the second half of the sentence above.}
%\arno{Should we mention Data Extraction and (JSON/Excel) Export? How?}
%\gabi{The whole figure has to be explained or the figure is simplified.}
%\arno{done}
%\arno{Added alternative diagram}
%\gabi{Good. Seems to be enough.}

\subsection{Data Value Clustering}
% basics
%\gabi{Tell the reader first what will be inside the backend.}
%\viola{Reworded}
%The actual clustering of the data values is performed by three 
The component \lstinline|Data Value Clustering| 
%is the core of our implementation.
%It 
is a Python implementation of the algorithm presented in Section~\ref{par:clustering} and allows performing data value clustering via an API.
%We implemented the algorithm presented in Section~\ref{sec:algorithm} with Python.
%In the following, we give a brief overview of its 
%It includes three subcomponents.
% functional

% compression
% - 18 compression rules + duplicate removal, can be combined to define compressions, dependencies, questionary, ...
% - 11 predefined compressions
% - function for application of compression
%\subsubsection{Compression}\label{sec:compression_impl}
For supporting the abstraction step, we implemented a set of abstraction rules %(see Sec.~\ref{sec:compression}) 
and a mechanism for applying those rules to data values.
It can be extended by further rules in the future.
For the calculation of the basic edit distance and the Levenshtein distance, we implemented a parallelisable version of the Wagner–Fischer algorithm~\cite{wagnerF74}. %that computes all cells on an anti-diagonal in parallel\todo{cite}.
%\gabi{I do not understand the last part of this sentence. Do we need this information here?}
%\viola{Shortened.}
%the Wagner–Fischer algorithm\todo{cite}, which is a dynamic programming algorithm based on calculating the pairwise distances of all prefixes of the two string values.
%For allowing parallel calculations we use the diagonal implementation suggested by ... .
%If all weights for substitutions of two characters are set to the sum of the weights for deleting and inserting the characters, the weighted Levenshtein distance, that in general supports the deletion, insertion and substitution of characters, corresponds to the edit distance that only allows insertion and deletion operations.
%This implementation also supports calculating the edit distance that only allows insertion and deletion operations.
%The edit distance that only allows insertion and deletion operations can also be calculated with this implementation if the weights are configured in a certain way.
%\gabi{Here is some redundancy. This distances have been defined above already.}
%The Damerau–Levenshtein distance, which additionally allows transpositions, is not yet supported but may be added in the future.
For improving the performance of distance calculations for large sets of potentially long data values, we use Numba~\cite{numba}, a compiler that translates a subset of Python code into faster machine code.
%\gabi{Why is performance a problem here? How is the basic edit distance implemented? }
%\viola{Performance is important as we have large sets of sometimes long data values. Distance calculation took up to an hour in our experiments. Added mentioning of basic edit distance above.}
%\gabi{Include the motivation into the text.}
%\arno{I do not think that we need to justify performance improvements for an interactive approach. We could add ``for large sets of potentially long data values''. Instead we also can remove ``of distance calculations'' or even the whole sentence. }
% TODO: diagonale method

%\subsubsection{Clustering}
% clustering
% - reference Python packages
% - pass distance matrix
%Our implementation of clustering algorithms 
%\viola{Shortened the following.}
We use implementations of clustering algorithms offered in the packages scikit-learn~\cite{scikit-learn}, scikit-learn-extra~\cite{scikit-learn-extra} and  scipy~\cite{scipy}.
%Our implementation uses several packages for machine learning in Python to implement the clustering:
%%\arno{does this way of clustering really count as mashine learning? These only try to find the best clustering for the values depending on their own optimization function.}
%%\arno{We give a choice between ... ?}
%It uses the clustering algorithms Affinity Propagation, DBSCAN, OPTICS and Spectral Clustering provided by the package scikit-learn~\cite{scikit-learn}.
%Furthermore, it uses the implementation of k-medoids offered in the package scikit-learn-extra~\cite{scikit-learn-extra} and the implementation of hierarchical clustering offered in the package scipy~\cite{scipy}.
% TODO: explain why we do not use the implementation of hierarchical clustering from scikit-learn?
% reason: does not allow to show dendrogram before specifying #cluster or distance threshold
%\gabi{Usage out of the box or any adaptations?}
%\arno{no adaptations.}

% GUI
% - tkinter
%For the implementation of the user interface we use the tkinter~\cite{tkinter} package. % TODO: move down?
%\gabi{This is frontend?}

\subsection{GUI}\label{sec:gui}
%\viola{Rename to  ``Front End'' or ``Vision of the Front End'' or ``Vision of the GUI''?} 
%\gabi{Show views here or anywhere else (separate report).}
%\arno{Most views are basic views (using text, check boxes, dropdown menus and slider; rework and improvements required). 
%	Only the 'Blob View' is graphically appealing. 
%	I added a screenshot below, however we maybe should move it to the appendix.
%}
%
%To enable domain experts to use our tool, 
%To facilitate the usage of our tool, especially for domain experts, we provide a graphical user interface.
To facilitate the usage of our tool, we provide a graphical user interface.
%Its key task is to enable domain expert to configure the tool such that all relevant domain knowledge is taken into account.
%To enable domain expert to configure the tool such that all relevant domain knowledge is taken into account, we provide a graphical user interface.
It is still under development and not yet empirically evaluated.
We give a short overview of the GUI developed so far.
\shortversion
{\annonymversion
	{Details are provided in \cite{supplementary_material_anonymous}.}
	{Details are provided in \cite{supplementary_material}.}}
{Details are provided in Section~\ref{appendix:tool} of the appendix.}
%\versionchoice
%{Screenshots of the views are provided in \cite{supplementary_material_anonymous}.}
%{Screenshots of the views are provided in \cite{supplementary_material}.}
%{Screenshots of the views are provided in Section~\ref{appendix:tool} of the appendix.}
%{Screenshots of the views are provided in Section~\ref{appendix:tool} of the appendix.}
The implementation of our GUI is based on the package tkinter~\cite{tkinter}.

%The key task of the GUI is to 
The GUI should enable domain experts to configure the tool such that all relevant \emph{domain knowledge} is taken into account.
%\gabi{The key task of the GUI is to enable the domain expert to configure the tool such that all relevant domain knowledge is taken into account. Please clarify this first.}
%\viola{Added sentence above.}
%The GUI allows selecting predefined configurations or launching a guided configuration via the views outlined in the following.
%First of all, the GUI provides a \lstinline|Basic Configuration View|, which allows selecting predefined configurations or launching a guided configuration via the views outlined in the following.
% via one of the following views: the \lstinline|Compression Configuration View|, the \lstinline|Distance Configuration View| and the \lstinline|Clustering Configuration View|.
%In the following, we briefly outline each view.
%how we envision these views to support the mapping between domain knowledge and the configuration.
%\viola{I think the above is the question of interest here, which we should try to answer briefly in the following subsections.}
%In the following, we briefly outline each subcomponent of the user interface.
%Finally, the \lstinline|Cluster Visualisation View| presents the calculated clustering to the user via the methods discussed in Sec.~\ref{par:visualisation}.
\question{Empirical evaluation: Usability of the proposed GUI}
%\subsubsection{Basic Configuration View}
%\viola{I would leave this section out and very briefly mention the view at the beginning of the GUI section.}
%\viola{Added brief mentioning of this view at the beginning of the GUI section.}
%This view is the starting point for the clustering.
%It enables the user to choose the basic settings and future options via several drop-down menus.
%For most future options this gives default options possibly the choice between alternative views.
%For example it gives the option of choosing previously used data value sets.
%Further it allows to predefine the clustering algorithm or choose to get help with the choice later.
%
%\subsubsection{Compression Configuration View}\label{sec:compression_gui}
%For mapping domain knowledge concerning the chosen data field to {\em abstractions}, i.e., valid combinations of abstraction rules, we developed a binary response questionnaire.
%For gathering domain knowledge about a chosen data field, 
For the configuration of the \emph{abstraction}, we developed a binary response questionnaire.
%For mapping domain knowledge concerning the chosen data field to compressions,i.e. valid combinations of the 20 compression rules mentioned in Sec.~\ref{sec:compression_impl}, we developed a binary response questionnaire.
%On the left hand side this view shows a questionnaire that maps domain knowledge concerning the chosen data field to compressions, i.e. valid combinations of the compression rules.
%It is realised in the \lstinline|Compression Configuration View| via checkboxes.
%It is realised via checkboxes.
%In our draft, all the questions are of the form ``Should all ... be treated equally?'' such as ``Should all lower case letters be treated equally?''.
The questions aim at the expert's assessment of the importance of certain syntactical features.
%, such as different letters or the word length.
The answers are translated to valid combinations of abstraction rules. 
%Examples and an explanation of when the answer should be yes are shown when hovering over a question.
%Examples and further explanations are shown when hovering over a question.
%On the right hand side the view shows 
%The result of the abstraction of a limited number of the extracted data values is shown dynamically.
Abstracted data values resulting from the application of these rules are shown exemplarily and updated dynamically. 
%when answers to the questionnaire are modified.
%\gabi{This sentence is not well understandable. Is it important?}
%\viola{Modified sentence.}
%Screenshots of the view are shown in 
%\versionchoice
%{Screenshots of the view are shown in \cite{supplementary_material_anonymous}.}
%{Screenshots of the view are shown in \cite{supplementary_material}.}
%{Screenshots of the view are shown in Section~\ref{appendix:tool} of the appendix.}
%{Screenshots of the view are shown in Section~\ref{appendix:tool} of the appendix.}
%Whether this approach is suitable must be empirically evaluated in the future.
%It updates dynamically when checkboxes are selected or deselected.
%We implemented a questionary which maps domain knowledge concerning the chosen data field to compressions, i.e. valid combinations of the compression rules.
%questionaire: lower-case, letters, digits, sequences, special characters, duplication removal

%\subsubsection{Distance Configuration View}\label{sec:distance_gui}

% String distance functions
%\viola{Modified sentence below such that only the Levenshtein distance is mentioned here. The basic edit distance is now mentioned further below.}
%Our tool currently offers two ways to set the weights that are used in the {\em calculation of Levenshtein distance}.
Currently, the weights for the \emph{Levenshtein distance} can be specified directly as numbers or
by placing graphical objects, which represent groups of characters, on a 2D canvas, whereby their distances define the weights for corresponding substitutions.
%
%Alternatively, they may be specified by placing and scaling graphical objects, which we call blobs, on a 2D canvas.
%%, as shown in Fig.~\ref{fig:blobview}.
%Each blob represents a character or a group of characters.
%%Here, the characters and character groups are defined automatically based on the chosen compression.
%The distance between two blobs represents the weight of substituting corresponding characters.
%\viola{Added sentences below. OK?}
We plan to investigate how far the distance configuration process can be simplified, 
for example by deriving substitution weights automatically to decrease the number of parameters that must be specified.
\question{Is the loss of expressiveness acceptable?}

%\subsubsection{Clustering Configuration View}\label{sec:clustering_gui}

For {\em selecting a clustering algorithm and configuring its parameters}, our tool provides a graphical interface based on standard widgets.
%, such as sliders and checkboxes.
How experts could be supported in appropriately selecting and configuring a clustering algorithm based on domain knowledge and experience is still  subject to future research.

Finally, {\em calculated clusterings are visualised} using tables and scatter plots (as presented in Sec.~\ref{par:visualisation}).
%\viola{Remove future work ideas?}
%\gabi{yes}
%\arno{moved to \ref{sec:conclusion}}
%Additionally, we envision a questionnaire on how satisfied the domain experts are with certain aspects of a produced clustering.
%The answers could be used to suggest modifications of the configuration in the next iteration.
%
%Furthermore, domain experts could be supported in interpreting the clustering.
%%, thus identifying quality problems and deciding on improvement strategies.
%For example, a list of generic quality problems in data models (similar to UML class model smells~\cite{Arendt14}), could be provided.
%The experts would be asked to assign these problems to the clusters.
%Subsequently, appropriate quality improvements of the data model (similar to UML class model refactorings~\cite{Arendt14}) could be suggested based on a survey of domain experts or supervised learning.
%%The underlying relation between problems and improvements could be determined via a survey of domain experts or supervised learning.

\section{Initial Evaluation}
\label{sec:evaluation}

% research questions
% set-up: data, configuration(s), experts, ...
% results: strengths & limitations, performance

%The approach to cluster data values in order to find problems in data models is experimental and exploratory.
%\viola{Do we need the following brief paragraph?}
%The presented approach is experimental and exploratory.
%It offers assistance to detect problems with a new point of view.
%The workflow is very interactive and configurable; the problems are detected manually by inspecting computed clusterings.

% goal
%\viola{Added additional research question below.}
%The goal of our approach is to support the detection of quality problems in data models.
%In this paper we focus on quality problems in data models, which may show through quality problems in data.
%Therefore, the evaluation should answer the following research questions.
%For the evaluation, 
We investigate the following research questions:
\begin{enumerate}[\IEEEsetlabelwidth{RQ1:}]
	\item[\textbf{RQ1:}] How far does the approach support the detection of quality problems in data models?
	\begin{enumerate}[\IEEEsetlabelwidth{RQ1.2:}]
		\item[\textbf{RQ1.1:}] Does the approach support the detection of quality problems in data models?
		\item[\textbf{RQ1.2:}] What kinds of quality problems in data models can be revealed with the approach?
	\end{enumerate}
%	\item[\textbf{RQ2:}] What kinds of quality problems in data models can be revealed with our approach?
\end{enumerate}
In the following, the setup of our evaluation is presented.
Findings and threats to validity are discussed thereafter.

\subsection{Setup} \label{sec:evaluation_setup}
%For the evaluation we selected four
%%representative
%data fields from two cultural heritage databases
%%in the LIDO and MIDAS formats
%and discussed clusterings of the corresponding data values with domain experts.
%Our approach is described in Sec.~\ref{sec:procedure}.
%The selected data fields are briefly presented in Sec.~\ref{sec:fields}.

\annonymversion
	{We selected four diverse data fields from two XML databases with cultural heritage data using the data models LIDO~\cite{coburn_lido_nodate} and MIDAS~\cite{bove_marburger_2001_anonym}.}
	{We selected four diverse data fields from two XML databases  with cultural heritage data using the data models LIDO~\cite{coburn_lido_nodate} and MIDAS~\cite{bove_marburger_2001}.}
%For the evaluation of our clustering approach we selected four diverse data fields from two cultural heritage XML databases using the data models LIDO~\cite{coburn_lido_nodate} and MIDAS~\cite{bove_marburger_2001}.
%\gabi{slightly adapted.}
%\viola{Added reference below.}
The data fields are briefly presented in Table~\ref{tab:fields} and Sec.~\ref{sec:example}.
We refer to them by the descriptive names given in the first column.
%which slightly differ from the original titles, i.e. XML element names.
The fourth column shows the number of distinct values in the corresponding database.
Altogether, the selected fields cover both numerical information (dating) and textual information (artist name, measurement unit and attribution qualifier).
%They further cover both information that could be limited to a few values and information that may occur in very diverse manifestations.
Furthermore, we selected fields that expect just a few different values (measurement unit and attribution qualifier) and fields where data may occur in very diverse manifestations (artist name and dating).
%\viola{Added motivation for field selection above. OK?}
%\gabi{Which field(s)?}
%\viola{Added.}
%Furthermore, two of the fields represent information that could be limited to a few values, while the other two may contain diverse values.
%The rest of the evaluation set up is outlined in the following.
%\versionchoice{Details can be found in \cite{supplementary_material_anonymous}.}
%	{Details can be found in \cite{supplementary_material}.}
%	{Details can be found in Section~\ref{appendix:evaluation} of the appendix.}
%	{Details can be found in Section~\ref{appendix:evaluation} of the appendix.}
%

\begin{table*}
	\renewcommand{\arraystretch}{1.3}
	\caption{Overview of data fields used for the evaluation}
	\label{tab:fields}
	\centering
	\begin{tabular}{
			p{\dimexpr 0.15\textwidth-2\tabcolsep} 
			p{\dimexpr 0.06\textwidth-2\tabcolsep} 
			p{\dimexpr 0.39\textwidth-2\tabcolsep} 
			r %p{\dimexpr 0.07\textwidth-2\tabcolsep} 
			p{\dimexpr 0.32\textwidth-2\tabcolsep} }
		\hline
		Field & Model & Explanation & \#Values & Example values (adapted to English) \\
		\hline
		Artist name & MIDAS & Name of an artist & 118,032 & ``Martin, A. (junior)'', ``Raffael?'', ``Walt ..., R.'' \\ 
		Dating & MIDAS & Dating of an object, such as completion or destruction & 52,523 & ``ca. 1840-1850'', ``1895/1902'', ``x'' \\
		Measurement unit & LIDO & Unit of a measurement of an object & 179 & ``-10.5 cm'', ``x 55 cm'', ``? cm'' \\
		Attribution qualifier & LIDO & Characterisation of the attribution of an actor to an event & 56 & ``attributed?'', ``studio / successor'' \\
		\hline
	\end{tabular}
\end{table*}

For the evaluation of our approach, we considered clusterings of the values of the selected data fields.
To answer the research questions, the quality of the clusterings must be evaluated in the intended usage scenario.
Consequently, we applied \emph{external} clustering validation (see Sec.~\ref{par:eval}): 
For the ultimate \emph{evaluation and interpretation of the clusterings} we recruited four domain experts on cultural heritage data.
Their expertise covers the following task areas:
data model development, usage and development of acquisition software, data acquisition, editing of data, and definition of data transformations.
All experts know both data models.

%\viola{Added details on how we got the clusterings below.}
%\subsubsection{Procedure}\label{sec:procedure}
%To investigate the research questions, the whole workflow presented in Section~\ref{sec:workflow} must be evaluated.
%\gabi{I would erase this first sentence as it is future work. Here just the resulting clusterings are evaluated.}
Since the GUI of the tool is still under development, it was not feasible yet to let domain experts go through the whole workflow.
%\gabi{Instead, the domain experts evaluate clustering results.}
%\viola{Moved paragraph concerning experts up.}
%To get the clusterings, we, therefore, used our own domain knowledge to configure the algorithm for clustering the values of the selected data fields.
To get the clusterings, we configured the algorithm based on our own domain knowledge, therefore.
%\viola{Slightly reworded sentence above.}
%We performed the configuration analogously to that of the running example, i.e. based on similar considerations as discussed in Sec.~\ref{sec:algorithm}.
%\viola{Added sentence above.}
%We did this based on criteria similar to those explained for the configuration of the running example in Sec.~\ref{sec:algorithm}.
Our domain knowledge is based on a greater experience in the field of cultural heritage data 
of more than a year, in particular,  good insight into the data models MIDAS and LIDO. 
%Moreover, we conducted several interviews and workshops with cultural heritage data experts on the topic of data quality.
Hence, we assumed that domain experts could have arrived at similar configurations.
% and thus, clusterings.
%\gabi{Either we immediately start with the clustering results and do not mention the workflow here (but in the threats to validity section) or we better argue why this set up is okay.}
%\viola{Moved argumentation from threats to validity section up here.}
%For all four clusterings, we used hierarchical clustering.
%Our domain knowledge is based on experience in the field of cultural heritage data of more than a year, insights into MIDAS and LIDO and several interviews and workshops with cultural heritage data experts on the topic of data quality.
%We used few iterations of reconfiguration after.
We assessed the quality of the initial clustering manually and in some cases modified the configuration slightly a few times until we yielded a clustering that made sense from our point of view.
\shortversion
{\annonymversion
	{The configurations are explained in \cite{supplementary_material_anonymous}.}
	{The configurations are explained in \cite{supplementary_material}.}}
{The configurations are explained in Section~\ref{appendix:evaluation} of the appendix.}

We presented the four clusterings to each of the experts independently using Excel files, each including three sheets.
%\viola{Modified sentences below.}
The first sheet shows a manually compiled excerpt of the corresponding data model documentation.
The other sheets were created by our tool and present the clustering of the abstracted values by means of original values as representatives (cf. Fig.~\ref{fig:clustering}), and the clustering of the original values.
The Excel files are provided in~\cite{eval_sheets}.
For each of the four clusterings,
%, we proceeded as follows.
%\viola{Reworded questions below to clearly differentiate quality problems in data values and their causes in data models.}
%Firstly, 
we firstly asked for an \emph{evaluation} of the clustering with the following questions:
(1) Which kinds of values are included in each cluster?
(2) Does the clustering make sense?
(3) Does the clustering help in detecting quality problems in the set of data values?
(4) Does the clustering bring new insights about the data considered?
%\gabi{I think we should switch questions (1) and (2). Further the experts try to get an  overview of the clustering, then the can judge whether it is makes sense.}
%\viola{Done.}
%\gabi{...new insights about the data considered?}
%\viola{Yes, added.}

%	\begin{enumerate}
%		\item Does the clustering make sense? % Do the single clusters make sense? % Does the grouping of these values in one cluster make sense?
%		\item Which types of values are included in each cluster? % Which values are included in the clusters? % What kinds of values are represented by each of the clusters?
%		\item Does the clustering help in detecting quality problems in the data values? % Is the grouping of these values in one cluster useful for detecting quality problems?
%		%\gabi{in data models?}
%		%\arno{Is it not better if we can draw the conclusion, that many problems can traced back to models?}
%		\item Does the clustering bring new insights?
%	\end{enumerate}
%Secondly, we asked for an \emph{interpretation} of the clustering in the context of data quality using the following questions:
Secondly, we asked for an \emph{interpretation} of the clustering concerning data model quality using the following questions:
(5) Which quality problems of the data values does the clustering reveal?
(6) Does it reveal quality problems of the data values you would hardly detect otherwise?
%\gabi{Are (5) and (6) concerning quality problems in data or data models?}
%\viola{Quality problems in data. Reworded.}
(7) What are causes for these problems in the data model?
(8) Which quality improvements of the data model would you suggest?
%\viola{Remove ``especially''?}
%\gabi{yes}

%	\begin{enumerate}
%		\setcounter{enumi}{4}
%%		\item Are the values in this cluster unexpected? %and/or problematic? Why?
%		\item Which quality problems of the data values does the clustering reveal? % In what way do the values have low quality?
%		%\gabi{Questions 3 and 5 are very similar.}
%		\item Does it reveal quality problems, you would hardly detect otherwise?
%		% \item Why are these values problematic? % What is bad about the values in this cluster?
%		\item What are causes for these problems, especially in the data model? % this? % \item How could these values have been introduced?
%		\item Which quality improvements, especially of the data model, would you suggest?
%	\end{enumerate}

At the end of each session, we asked the expert for a final \emph{conclusion} using the following questions:
(9) Do the clusterings support the detection of quality problems in data models?
(10) What kinds of quality problems in data models can be revealed by clusterings of data values?

\subsection{Findings} \label{sec:evaluation_findings}
Per clustering, we summarise the answers to Questions (1) and (2) in the \emph{evaluation} paragraph and the answers to (5), (7) and (8) in the \emph{interpretation} paragraph. The answers to (3), (4) and (6) were similar for all clusterings and thus, are discussed together with those to (9) and (10) in Sec.~\ref{sec:general_evaluation}. 

\subsubsection{Artist Name}

\paragraph{Evaluation of the Clustering}
% mostly
% yes, but the number of clusters is high
% yes

% clusters: comma, abbreviated first name, additional info in brackets, ?, ...

% yes, it reveals potentials for improvements, particularly missing differentiations, in-depth analysis of large, heterogeneous clusters is necessary, provides awareness of what should be paid attention to when performing deeper analysis
% expert 3: could additionally be used to identify data model requirements coming from domain experts, which they often are not able to formulate directy but are indicated by them giving information in the data where not expected

% yes, shows heterogeneity in syntax and semantics

The domain experts agreed that the clustering makes sense.
They intuitively assigned a specific meaning to most of the clusters. 
%They were able to intuitively assign a specific meaning to most of the clusters. 
%Most clusters were intuitively assigned a specific meaning by the domain experts.
%Among others they discussed the following clusters and their meanings.
%For example, one cluster contains values where the first and middle names are abbreviated due to a lack of knowledge.
For example, one cluster contains values with additional information appended in brackets.
The experts explained that this is often done to ensure the uniqueness of entries as artist names are used as identifiers in MIDAS.
%the artist names are used as identifiers in MIDAS and thus often additional information is appended to ensure uniqueness.
Another cluster comprises values that contain question marks at different positions. 
%%According to the experts, they encode different kinds of uncertainty such as uncertainty concerning the relation between an object and the artist or uncertainty concerning the first name of the artist.
%According to the experts, they encode different kinds of uncertainty e.g. concerning the relation between object and artist or the first name.
According to the experts, they encode different kinds of uncertainty, for example, concerning the first name or the relation between an object and the artist.
Further clusters represent further uncertainties encoded implicitly using other special characters. %such as a dot for abbreviating names that are unknown otherwise. %or a slash for separating the names of potential alternative creators of an object.
%
%\gabi{Given an example here.}
%\viola{Added example below.}
%\viola{Added reference to Sec.~\ref{sec:example} below.}
The values found in the database do not necessarily conform to the syntax rules described in the MIDAS manual (see Sec.~\ref{sec:example}).
%and mentioned in Sec.~\ref{sec:example}.
%An example is 
For example, the value \lstinline|Heitz, Heinrich ?| is not covered by any of the rules.
%Thus, its semantics is unclear.
%\viola{Listed all rules concerning question marks below. Is this necessary?}
%According to the manual, question marks shall be placed after the comma if the first name is unknown.
%If the identity of the artist is uncertain, a question mark in round brackets should be placed after the first name.
%Uncertain attributions are encoded through a question mark between the last name and the comma.
%Thus, the data contains even further heterogeneous implicit encodings.
%Hence, there are additional implicit encodings such that the set of data values is even more heterogeneous than the manual implies.
%Through such additional implicit encodings, 
Thus, the set of data values is even more heterogeneous than the manual implies.
%Further implicit encodings, such as references to other records or indications of multiple artists, are represented by further clusters.

\paragraph{Interpretation of the Clustering}
% data quality problems: implicit encoding of uncertainty concerning first name or association between object and artist, values with low expressiveness (e.g. signatures), concatinations, implicit reference to other records

% expert 3: yes, allows me to identify structural problems very fast through clear overview
% expert 3: alphabetically sorted list would not reveal uncertainty related problems
% expert 3: when looking only at samples I would assume the problem is in the data acquisition, but clustering allows me to understand that there are data model problems

% causes in MIDAS: arbitrary strings allowed, name and homonym supplement in same field, name as an identifier, implicit encoding of AND, OR and uncertainty

% improvements of MIDAS: make uncertainties (including cases in which only initials are known) explicit, use unique numerical identifieres instead of names, syntax constraints, split first and last name, support AND and OR explicitly 

The experts identified an encoding of additional information, which is given implicitly and shows some heterogeneity.
They considered it as problematic
%The experts identified an encoding of additional information as problematic, which is given implicitly and shows some heterogeneity. 
%The encoding of additional information 
%It was 
and largely attributed it to the fact that the artist names are used as identifiers and therefore need to be unique.
The experts suggested using numbers that are automatically generated as identifiers instead.
Different kinds of uncertainty in the data field were considered as a further problem.
%The experts advised supporting the different kinds of 
The experts identified the requirement to support the different kinds of
%to express 
uncertainty explicitly by using additional structure in the data model to increase understandability.
Furthermore, they questioned whether it makes sense to accept arbitrary string values as artist names in MIDAS.
%Further, they questioned whether artist names are acceptable as arbitrary string values in MIDAS.
Some experts mentioned a splitting of the name into several fields.
%
%They discussed the following quality problems and improvement ideas for MIDAS.
%To improve this situation, 
%Additionally, the domain experts  %first of all 
%suggested to use numbers as identifiers instead that are automatically generated.
%\gabi{Paragraph reformulated. Please check.}
%\viola{Checked. I think the use of the artist names as identifiers was considered the greatest problem and cause for heterogeneity. So I moved this aspect before the discussion of uncertainty.}

%Furthermore, separators, such as ampersand and slash encoding AND and XOR relations, should be replaced by structure.
%The implicit and heterogeneous encoding of several types of uncertainty was considered problematic.

\subsubsection{Dating}

\paragraph{Evaluation of the Clustering}
% expert 1: yes, in general, but some clusters are heterogeneous or contain outliers
% expert 2: yes, very much
% expert 3: yes

% clusters: um, nach, vor, ante, ...

% yes
% yes, it gives a very good compressed overview of the values, good starting point for quality assurance, allows deriving of tasks, analyse cluster by cluster in detail and perform improvements

% yes, new insights concerning variants of uncertain dates and combinations
% expert 2: confirmation of expectation but still very helpful and necessary

%\viola{Work in progress}

Overall, the experts considered the clustering meaningful.
The discussions of the following clusters were most interesting:
%For most clusters the domain experts intuitively assigned a specific meaning.
%Among others they discussed the following clusters and their meanings.
%Most clusters were intuitively assigned a specific meaning by the domain experts.
There are several value clusters that contain specific textual indications of uncertainty or imprecision as a prefix to a numerical date, such as \lstinline|around|.
%and \lstinline|before|. % (often only a year).
Often, several (uncertain) dates are given in one value, separated by a minus or slash, such as \lstinline|after 1831/before 1852|.
The experts stated that entries which include a minus or slash probably represent 
%different meanings, namely 
a period of time and an unknown point in time within the interval, respectively.
%\gabi{This explanation does not suit to the example above.}
%\viola{Corrected order of minus and slash in sentence above.}
%one might encode an interval directly and the other an unknown point in time within the given interval.
%Another cluster comprises imprecise dates of the form \lstinline|mid-19th century|.
%These key words are followed by a numerical date (most often only a year) 
%Often this is followed by a special character (mostly slash and minus) and a further date specified in this manner.
%numerical date sometimes with a textual indication of uncertainty, e.g. \lstinline|after 1831/before 1852|.
Other clusters comprise entirely textual values indicating some lack of knowledge, such as \lstinline|without year|.
%\gabi{such as...?}
%\viola{Added.}
%Some clusters contain year indications followed by \lstinline|ante| in a specific syntactical context, which according to the experts encodes dates before Christ.
%The clustering further reveals that additional numerical or textual information is sometimes appended to the date indication in brackets.
%
%In summary, the experts agreed that the clustering gives a very good overview of the values and thus can serve as a starting point for quality assurance.
%In particularly, it provides insights into the implicit encoding of uncertainty.
%
%One expert noted that some clusters %are heterogeneous or 
%contain outliers.
%\viola{Added reference to ec.~\ref{sec:example} below.}
Again, the set of data values considered contains more implicit encodings than mentioned in the MIDAS manual (see Sec.~\ref{sec:example}).
%do not conform to the syntax rules described in the MIDAS manual and mentioned in Sec.~\ref{sec:example}.
%Thus, the data contains even further heterogeneous implicit encodings.
%Thus, through additional implicit encodings, the data values are even more heterogeneous than the manual implies.

\paragraph{Interpretation of the Clustering}
% data quality problems: non-numerical values in contrast to expectation, implicit, heterogeneous textual and symbolic encoding of uncertainty, intervals and ante decrease interpretability (comparability, searchability etc.) 

% analysing (sorted) list would not be feasable and does not provide systematic overview of structures
% compression and clustering make it manageable

% causes in MIDAS: arbitrary strings allowed, implicit heterogeneous encoding of uncertainty, use of non-clear terms (e.g. "um"), encoding of additional information

% improvements of MIDAS: check against ISO 8601 (isolate numerical values), encode meaning of special characters and textual modifiers explicitly in other field(s), support intervals explicitly (also with uncertain boundaries), additional comment field

%\viola{Work in progress}

The implicit and heterogeneous encoding of uncertainty, imprecision and missing knowledge was identified as the main quality problem of the data values.
%
%The experts agreed that compression and clustering make the large set of heterogeneous values manageable.
%Analysing a (sorted) list of all values instead would not be feasible and would not provide a systematic overview of syntactical differences, the experts said.
%
%The experts identified the following quality problems and required improvements of MIDAS.
%Again, 
The experts attributed the values' heterogeneity to the fact that they are not checked against syntax rules.
%As one cause for this, the experts identified the fact that the values are not checked against syntax rules.
%The heterogeneity of the data values was possible because they are not checked against syntax rules.
Furthermore, they expressed the need to separate numerical dates (conforming to ISO 8601~\cite{ISO8601}) from special characters and textual modifiers that encode uncertainty.
%One expert proposed to check the numerical dates against ISO 8601~\cite{ISO8601}.
The requirement to support the expression of uncertainty in a homogeneous and clear way was also mentioned.
%The requirement to avoid unclear indications of uncertainty, 
%erase heterogeneity in the expression of the same type of uncertainty was also mentioned.
Also, the meaning of the separators 
%i.e. intervals and an unknown point in time in an interval, 
should be 
%supported and 
differentiated by additional structure in the data model.
Regarding this, two experts stated the need to support both certain and uncertain boundaries of intervals.
%Additional fields were proposed to encode 

\subsubsection{Measurement Unit}
The evaluation and interpretation of the clustering are discussed in Sec.~\ref{par:eval} and Sec.~\ref{par:interpretation}.

\subsubsection{Attribution Qualifier}
%\gabi{This example is most unintuitive. I would not start with this one. And it deserves concrete examples. Why not starting the other way around with the artist name?}

\paragraph{Evaluation of the Clustering}
% yes, makes sense

% clusters: question mark, multiple statements, multiple statements with question mark, slash, ...

% yes, problems of the data model are revealed
% (at first problems in the data transformation were identified but then also problems in both data models)

% yes, insights into how multiple statements are combined and how uncertainty is expressed implicitly
% yes, insights into specific syntax used to express special meaning
% yes, plus could be used to develop terminology

%\viola{Work in progress}
The experts considered the clustering as useful and %meaningful.
%They 
assigned a potential meaning to each cluster.
%
%For example, one cluster contains values consisting of one or two words without any special characters.
For example, one cluster contains single words followed by a question mark, which %according to the experts 
indicates uncertainty.
Furthermore, there are two clusters of values containing multiple statements separated by an ampersand or a slash.
The experts read them as AND and XOR relations, respectively.
%Furthermore, there is a cluster that contains two words separated by an ampersand, which may point to an AND-relation.
%Yet another cluster contains slash as a separator, which according to the experts probably represents an XOR-relation between the statements.
%The other clusters represent combinations of these structures.
%
%According to the experts, the clustering helps in detecting quality problems in data models as it gives an overview of the syntactically and semantically different types of values.
%One expert noted that the clustering could also support the development of a corresponding terminology.
%The experts stated that the clustering gives insights into how multiple statements are combined in one value and how uncertainty is expressed implicitly.

\paragraph{Interpretation of the Clustering}
% data quality problems: implicit encoding of multiple statements (with AND or OR relations) and uncertainty
% plus missing differentiation between type of attribution and relation between described actor and person actually attributed to the event

% expert 1: not really, analysing the unclustered values would have been feasible and profitable too since the number of values is relatively low
% expert 2: confirmation of expectation
% expert 3: clearer than simple list

% causes in the transformation: & not mapped to field repitition

% causes in LIDO: arbitrary strings allowed, LIDO documentation does not clearly state when and how to repeat the field, meaning of & not explicitly supported?, explaining uncertainty explicitly not supported, OR-relation (probably encoded by slash) not supported, distinction whether it was the described or another person not supported and differentiation to type of attribution not supported
% expert 2: LIDO should be lightweight, i.e. maybe low expressiveness is on purpose
% when LIDO does not support expressing certain information some institutions adapt definitions of LIDO fields locally

% improvements of LIDO: controlled vocabulary (URIs), advice on how/when to repeat field or support other meaning of & explicitly, support uncertainty explicitly and with explanation, support OR-relation explicitly and distinguish from AND relation, support distinction whether it was the described or another person and distinction to attribution type, additional display element
% imrovements are not trivial and must be well thought

%\viola{Work in progress}
The experts identified an implicit encoding of multiple statements and uncertainty in the data values as problematic.
%They considered it as problematic.
%One expert argued that since the number of values is relatively low, manually analysing a list of all data values would have been feasible and profitable too.
%
%Based on the clustering, the experts 
%They identified the following potential quality problems and improvements of LIDO.
%First of all, the clustering showed that the values are heterogeneous in syntax.
%First of all, they concluded that the values are heterogeneous in syntax.
%They attributed the values' heterogeneity in syntax to the fact that arbitrary strings are allowed in the field.
%First of all, they noted that the clustering reveals heterogeneous in syntax.
%The experts identified the fact that arbitrary strings are allowed in the field as a main cause.
%As the main cause for this
%To decrease the values' heterogeneity, they suggested that the values allowed could be limited to those of a controlled vocabulary.
To reduce the heterogeneity, they suggested to limit the set of allowed values to a controlled vocabulary.
%To reduce the heterogeneity of the values, they suggested that the set of allowed values should be limited to a controlled vocabulary.
%(referenced via URIs).
%The clustering further revealed that often data values include multiple statements separated by an ampersand.
%The experts further observed that often data values include multiple statements separated by an ampersand, which encodes an AND-relation.
%The experts further considered data values including multiple statements separated by an ampersand or a slash as problematic.
%They further stated that 
Furthermore, the meaning of the separators ampersand and slash should be distinguished explicitly by the data model via structure.
%Whatever the meaning is should be supported and differentiated explicitly by the data model.
%In case this represents an encoding of AND-relations, the field should be repeated.
%The experts further considered the implicit encoding of AND-relations by an ampersand as problematic since the field is repeatable.
%The values containing an ampersand 
%This indicates that 
The concatenation of multiple statements also raised the question whether the LIDO documentation could be clarified regarding the repetition of this field.
%\viola{Added sentence below.}
Potential problems in the transformation by which the data was created, were also mentioned.
%does not clearly state when to repeat the field.
Besides, the experts mentioned the requirement that LIDO should support the documentation of uncertainty in another field.
%explaining any uncertainty concerning the attribution in another field.
%, thus should be extended by corresponding structure.
%According to the experts, values followed by a question mark indicate that LIDO does not support explaining uncertainty concerning the attribution explicitly, thus could be extended by corresponding structure.
%One of the clusters represents values that contain a slash.
%One of the clusters represents values including multiple statements separated by a slash.
%The experts noted that this indicates
%Moreover, the need to support XOR-relations explicitly (i.e. through structure) and distinguish them from AND-relations was identified.
%Some experts additionally noticed that LIDO does not support the clear distinction whether it is the person described by the data (e.g. ``attributed'') or another related person (e.g. ``school'') that is attributed to the event.
%Additionally, the type of attribution (e.g. ``attributed'' vs. ``former attribution'') is encoded in the field.
%Hence, they suggested further differentiating these types of information through additional data fields.
Some experts suggested the introduction of an additional field to distinguish between the following information:
the type of the attribution (e.g. \lstinline|attributed| vs. \lstinline|alternative attribution|) 
and the relation between the person described with the data and the person attributed to the event (e.g. \lstinline|school| vs. \lstinline|successor|).
%, which may not be the same.
%, as the example values demonstrate. %value \lstinline|school| demonstrates.
%\gabi{This sentence is difficult to understand as there is no concrete example given. }
%\viola{Added examples and reworded sentence.}
%For example, the data value ``school'' indicates that a pupil of the person described by the data is attributed to the event.
%as is the case when the value is ``school''.
%The experts suggested to introduce an additional field for this differentiation.
%differentiating between the type of attribution and the relation between the person described in the data and the referenced person.
%One of the experts stated that such quality improvements of LIDO that add structure are not trivial and must be well thought.

\subsubsection{Conclusions by Domain Experts}\label{sec:general_evaluation}
Concerning Question (3), the experts stated that
%According to the experts, 
the clusterings give good overviews of the heterogeneous syntax of the data values.
%all kinds of values that occur and how they are expressed syntactically.
%the syntactically 
%%and often semantically 
%different types of data values.
%Thereby, unexpected syntax indicating problematic data values is revealed.
%Thereby, unexpected syntax indicating quality problems in the data values is revealed.
This allowed the experts to derive quality problems in the set of data values.
When considering Question (4), they agreed that all four clusterings bring new insights.
% concerning the values' heterogeneity in syntax and semantics.
%which typically indicates low quality of the data and the data model.
%The bottom-up approach allows understanding 
The clusterings reveal how the data model is used in practice as opposed to how it is expected to be used.
Deviations indicate quality problems of the data model, often related to previously unknown requirements.
%\viola{Added sentence below.}
Further, quality problems of the data transformation can be revealed.
%\gabi{some redundancy here}
%Differences often indicate 
Even %in cases in which the experts were 
when being aware of the problems previously, the experts found the clustering useful for understanding the problem in depth and for considering quality improvements.

To Question (6),
they reported that, for data quality assurance, they typically use a list of all data values of a field of interest sorted alphabetically or by the number of occurrences.
They agreed that clusterings provide a much clearer and more systematic overview than their previous practice.
%They stated that the clusterings have the following advantages: 
They stated the following advantages of clusterings:
they allow the detection of a wider range of quality problems in data, the detection is faster, and the backtracking to quality problems in the data model is easier.
The experts argued that the more data values there are in a field, the greater the benefit of clustering.
%The reduction of the number of values by showing representatives of compressed values was considered profitable.
%The clustering of the compressed values shown through representatives was considered especially profitable.
%through representatives of a compressed value was considered especially profitable.
%understand their causes in the data model.
%Thereby, they reveal more quality problems and their relation to the data model, according to the experts.
%Thereby, they reveal more quality problems and require less manual effort, according to the experts.
%with increased efficiency.
%While some quality problems can be reveal by sorting the values alphabetically, others are on
% and require less manual effort to be analysed.

%In summary, concerning RQ1.1, the clusterings supported the domain experts in detecting of quality problems in data models.
%In summary, 
Concerning Question (9) and thus RQ1.1, the experts agreed that the clusterings provide an overview of the data values' heterogeneity in syntax, which supports the detection of quality problems in the data model, in part by revealing previously unknown requirements.
%They observed that this heterogeneity often indicates quality problems in the data model.
%heterogeneity in syntax often implies heterogeneity in semantics, which often indicates quality problems in the data model.
%which allows understanding the data values' heterogeneity in syntax.

%the clusterings allow the detection of quality problems in the underlying data models.
%Thus, the experts considered a corresponding tool that produces clusterings as desirable (question 10).

% answers to questions 9, 10, 11 because we asked them to get a final conclusion

% 9 Can clusterings of data values give new insights concerning data quality?
% answered above

% 10 Could a tool that offers such (configurable) clusterings support the quality assurance of data models?
% ??
%\viola{Answer question 10}

% 11 What kinds of quality problems in data models can be revealed by clusterings of data values?
%
% + types of problems detected:
%   - encoding of multiple information in a single field through specific syntax -> missing structure in data model
%   - implicit encoding of uncertainty
%   - missing syntax constraints
%   - possible reasons for multiple information encoded in a single value (identified by experts 2 and 3):
%     1. data model lacks possibilities to express the information otherwise
%     2. data model supports to express the information otherwise but is misued, potentially due to bad documentation
Concerning Question (10) and thus RQ1.2, the experts observed that the clusterings primarily reveal encodings of multiple information in a single value using specific syntax.
%which indicates missing structure in the data model.
Often, uncertainty about the actual information is encoded implicitly. 
%due to one of the following causes:
%in the same value.
%The experts identified two possible reasons for this: 
According to the experts, this occurs either due to misuse of the data model, potentially caused by documentation issues, or 
%because the data model lacks the ability to express the information explicitly.
because the data model does not support expressing the information explicitly.
%or the data model actually supports that but is misused, potentially due to documentation issues.
%This occurs because either the data model lacks the ability to express the information explicitly or the data model actually supports that but is misused, potentially due to documentation issues.
Often, the values' heterogeneity also indicates the use of wrong data types or a lack of syntax constraints.
Note that some of the mentioned problems of LIDO v1.0 were addressed in LIDO v1.1 Public Beta~\cite{coburn_lido_nodate_v11}.

\subsection{Conclusion} \label{sec:evaluation_conclusion}
\emph{Concerning RQ1.1, we found that, in the chosen setting, the approach supports the detection of quality problems in data models.}
\emph{With regard to RQ.1.2, the findings imply that especially missing structure but
also inappropriate data types, lack of syntax constraints and problems in the documentation can be exposed.}
%\viola{Added sentence above.}
%After identifying specific kinds of information whose expression in the data model needs
%Based on the observation of quality problems in the data models, 
%The clusterings further supported experts in coming up with vague ideas for quality improvements.
The clusterings further helped the experts to come up with ideas for improvements.
%The experts further developed ideas for quality improvements based on the clusterings.
%\gabi{some ideas?} % instead of vague ideas
%\viola{Added observation concerning answers to question 8 above.}
%The quality improvements suggested by experts were not concrete.
%Concerning question (8),  the experts often did not formulate concrete ideas for quality improvements.
%They identified specific kinds of information instead; the data model needs to be reconsidered to clarify how it can cover that information.

%Our findings confirmed that the amount of semi-structured data is quite high in the domain of cultural heritage.
%%This may stem from the data being input manually.
%%Another characteristic of our set up is that 
%%The cultural heritage 
%The input of this data is usually manually and fields are hardly constrained, which may lead to a high amount of heterogeneity of data values.
%%\gabi{Is this correct?}
%%\viola{yes}
%The evaluation further showed that the documentation of uncertainty plays a major role in this domain but is often not supported explicitly by data models yet.
%\viola{Should we remove the above paragraph? It seems redundant to the one below.}

%
%The presented evaluation remains an experiment in the domain of cultural heritage.
%The presented evaluation is limited to the domain of cultural heritage.
%\viola{Condensed statements below.}
The presented evaluation was performed in the domain of cultural heritage, where data is often semi-structured and largely created manually.
This may lead to a high amount of heterogeneity of data values, especially to implicit encodings of uncertainty.
%The presented evaluation was performed in the domain of cultural heritage, which has the properties discussed above.
%\gabi{Dies sollten wir etwas positiver formulieren: The presented evaluation was performed in the domain of cultural heritage., domain where data is often semi-structured and data acquistion is largely performed manually.}
How the approach performs on significantly different kinds of data is subject to future research.
%\gabi{different contexts?}
%From the evaluation performed so far, we conclude that it supports the quality assurance of data models in contexts where the data is not fully structured and its input is typically performed manually.
%\gabi{NachUmformulierung waere der letzte Satz nicht mehr noetig.}
%\arno{Reworded:}
%Thus, we chose cultural heritage data as subject to our evaluation.
%How the approach performs in other contexts is subject to future research.

\subsection{Threats to Validity} \label{sec:evaluation_threads}
%\gabi{This is construct validity.}
%The main threat to construct validity is that the configuration step of the workflow was not performed by external domain experts but ourselves.
The main threat to \emph{construct validity} is that the configurations were not performed by external domain experts but by ourselves.
%The main threat to construct validity is that not all of the user activities of the workflow, especially the configuration, were performed by external domain experts. %, as the approach may suggest.
But as argued in Sec.~\ref{sec:evaluation_setup}, we have experience in the domain and thus assume that external experts could have arrived at similar configurations. %and thus, clusterings.
%\viola{Explain here that we plan to perform another evaluation once the GUI is ready? Currently, this is mentioned very briefly in future work.}
%\viola{Added sentence below.}
Once the GUI is completed, we will conduct a thorough empirical evaluation where the configuration will be performed by domain experts.
%Instead, we performed the configuration step ourselves.
%Our domain knowledge is based on a greater experience in the field of cultural heritage %data 
%of more than a year,  in particular,  good insight into the data models MIDAS and LIDO. Moreover, we conducted several interviews and workshops with data experts in cultural heritage on the topic of data quality.
%Hence, we assume that domain experts could have arrived at similar configurations and thus, clusterings.
%%The evaluation and interpretation of the clustering were performed by domain experts.

Threats to \emph{external validity} could be the selection of the databases, fields and domain experts.
%The limitation to the domain of cultural heritage definitely is a threat.
%That both databases describe cultural heritage objects definitely is a limitation.
%However, the two data models underlying the selected databases differ greatly.
The two data models we selected are diverse:
%represent the main two kinds of models: 
LIDO is event-oriented, intended for data exchange and developed by a national working group, whereas MIDAS is object-oriented, intended for data acquisition and developed locally in a cultural heritage institution.
%LIDO is developed by a national working group.
%MIDAS is developed locally in a cultural heritage institution.
As explained in Sec.~\ref{sec:evaluation_setup}, the selected data fields are also quite diverse.
%They cover both numerical and textual information.
%Furthermore, two of the fields represent information that could be limited to a few values, while the other two may contain diverse values.
To counter the relatively low number of experts, we selected them carefully to cover a broad range of tasks related to data acquisition, data models and data transformations.
All experts provided valuable input to all questions.
% threat: set of domain experts: how representative are they?
% threat: small sample sizes concerning clusterings presented

%Threats to internal validity include a potential relation between the presentation of the questions and the answers as well as potential learning effects during the experiment.

% conclusion validity threat: reliable clusterings?
% conclusion validity threat: heterogeneous knowledge of experts

%\arno{Suggestion:
%	In the upcoming GUI (view Sec.~\ref{sec:gui}) we will implement features to support the configuration step based on our experiences.
%	Further we will implement the settings for the example clusterings as pre-selectable default settings.}
%\gabi{Better fits to future work.}
%\gabi{Further threats: external: selection of databases and fields, and set of domain experts: how representative are they?,  small sample sizes concerning clusterings presented and number of experts, internal:  is there a relation between the presentation of the questions and the answers? Are there learning effects during the experiment?, conclusion validity: reliable clusterings? heterogeneous knowledge of experts}
%\viola{Should we list only the threats or should we additionally explain how we countered them?}
%\gabi{The more we can counter them the better. The adjectives (external,...) concern the validity, not the threats.}
\question{Open task: evaluation of the whole workflow with domain experts}

\section{Related Work}
\label{sec:related}

Our approach brings together several different research activities: 
{\em clustering of data to detect data quality problems}, {\em approaches to homogenise data} and {\em quality assurance of models}. 
Hence, we consider related work in these directions. 

% Data quality analysis via clustering
% - especially "A Pattern-Based Framework for Addressing Data Representational Inconsistency" by Yi et al. 2016 \cite{yi2016}
% Example-driven meta-model development

%\arno{Do we need a defensive section? Suggestion:}
%\subsection{New approach}
%	This approach combines several well-known concepts and techniques.
%	Consequently we have to ensure that our approach is novel for identifying problems in data models. %within the subject of problem detection.
%	To ensure this, we conducted an excessive literature study ... 
%	
%	As a basis, we have created a comprehensive list of search terms: 
%	...
%	With each combination of search terms we visited the first 50 results from google scholar.
%	For each search result, that could possibly be related to our approach, we took a closer look at the abstract.
%	If we were still not convinced that the paper was not related to our approach, we read the paper carefully. 
	
\question{How can the approach be combined with existing approaches?}

\subsection{Detecting Problems in Data Quality with Clustering}
%\subsection{Data clustering}
%\viola{I think this title is too broad. Clustering is always applied to some kind of data. To narrow it down maybe we should state what kind of data we consider and that we focus on approaches in the context of data quality.}
%\viola{Suggestion: Clustering of Qualitative Data for Quality Analysis}
%Clustering is a useful technique for grouping data that shows some similarities.
%Data can be structured, semi-structured or unstructured.
	%\viola{Slightly reworked and supplemented below.}
A variety of approaches apply clustering based on edit distances to detect \emph{minor inconsistencies in textual data values}.
%	In the context of data quality analysis, clustering is typically applied to detect \emph{inconsistencies in textual data values}.
	For example, such clustering is applied to detect {\em misspellings, typos and abbreviations} in textual geographical data~\cite{PPS21}, correct misspelled data values without external reference data~\cite{Cis08}, support the creation of authority files~\cite{FrenchPS00} and detect {\em duplicates} in medical records~\cite{SauleauPB05}.
%	
%	Pellegrino et al.~\cite{PPS21} use the Levenshtein distance and hierarchical clustering to detect {\em misspellings, typos and abbreviations} in textual geographical data. 
%	Ciszak~\cite{Cis08} applies a modified Levenshtein distance and clustering to 
%	support
%%	data cleaning.
%%	This allows 
%	the correction of misspelled data values without external reference data.	
%%	Similarly, French et al.~\cite{FrenchPS00} apply clustering to bibliographic entries to detect spelling variants and ultimately create authority files.
%%	Similarly, French et al.~\cite{FrenchPS00} apply clustering to bibliographic entries to support the creation of authority files.
%	Similarly, French et al.~\cite{FrenchPS00} support the creation of authority files.
%	Sauleau et al.~\cite{SauleauPB05} enable the detection of {\em duplicates} in medical records using an edit distance and clustering.
%	In contrast to these approaches, our approach does not aim at revealing minor inconsistencies in data values.
%	Instead of revealing minor inconsistencies in data values, 
Our approach provides an overview of more significant differences in the syntax of data values to reveal problems in the underlying data model.
%	In contrast to these approaches, our approach does not aim at revealing minor inconsistencies in data values, but at providing an overview of significant differences in the syntax of the data values to reveal problems in the underlying data model.
%	Our goal is to provide an overview of significant differences in the syntax of data values in a selected data field to reveal problems in the underlying data model. %on a more abstract level.
	%	Consequently, the approach to clustering differs:
	%%	as explained in Sec.~\ref{sec:compression}. %as well.
	%	the dissimilarity between data values is not calculated based on individual characters but on interesting syntactical features determined by domain experts.
	Thus, we do not calculate the dissimilarity between data values based on individual characters but on interesting syntactical features determined by domain experts (see  Sec.~\ref{par:clustering}).
	
	%\viola{Added non-clustering approach below. I am not sure if and where we should mention it. In summary, from my understanding this approach may produce \emph{similar results} as ours via a \emph{very different method} for achieving a \emph{very different goal}.}
	
%	While Dai et al.~\cite{} provide a quantitative measure for a specific type of data field heterogeneity, we suggest a qualitative approach, which allows finding quality issue of the underlying data model.
	
%	\viola{Added paragraph below. Please check.}
	Dai et al.~\cite{DaiKOSV06} present a quantitative measure for data field heterogeneity based on cluster entropy and soft clustering.
	They focus on semantically different types of information given in the same column.
	They do not consider further forms of heterogeneity such as implicit encodings of additional information.
	Our qualitative approach allows finding quality issues of the underlying data model.
	
\subsection{Approaches to Homogenising Data}
%Often, data shall be homogenised to enable more effective and efficient data processing. 
%We start with considering an approach that supports the homogenisation of {\em pure data values} using a pattern-based approach and go over to clustering approaches for homogenising {\em semi-structured data}.
%We start with a pattern-based approach that supports the homogenisation of {\em pure data values} and go over to clustering approaches for homogenising {\em semi-structured data}.
	
	%A different approach to detecting inconsistencies in textual data values is presented by Yi et al.~\cite{yi2016}.
%	Yi et al.~\cite{yi2016} propose 
	A \emph{pattern-based approach to homogenising data values} was proposed in~\cite{yi2016}.
%	, which 
	It provides an overview of the data values' syntax similar to ours.
%	The approach by Yi et al.~\cite{yi2016} 
%	It provides an overview of the data values' difference in syntax similar to ours.
%	It provides an overview of their differences in syntax similar to ours.
%	The method to achieve that, however, differs significantly from ours.
%	Instead of using clustering to analyse the data in a bottom-up manner, 
%	This approach, however, 
	It requires users to iteratively design patterns (consisting of regular expressions) manually.
	Instead, we use clustering to analyse the data in a bottom-up manner.
%	Those patterns are sequences of ``fields and separators expressed as regular expressions''~\cite{yi2016}.
%	Further, the motivation differs: 
	This approach aims to unify inconsistent data values that represent the \emph{same} type of information, whereas ours aims at detecting quality problems in data models, in particular \emph{different} types of information being encoded in the same field.

	Approaches to {\em clustering semi-structured data}, such as XML data~\cite{AMN+11}, %and a fuzzy-based approach \cite{CDL+05} 
	head towards the homogenisation of data structures.
%	head towards the detection of similar structures to homogenize data structures.
%	 for more effective and efficient data processing.
%	Those approaches do not focus on data values but on structures and thus define the similarity of data values on a more abstract level than we do.
	The underlying similarity measures focus on structural aspects.
	For measuring the similarity between data values, values are interpreted as multisets of words and token-based measures are applied.
%	When the similarity between data values is considered, data values are interpreted as multisets of words and token-based measures are used.
%	Approaches that consider the similarity between data values interpret them as multisets of words and use token-based measures.
%	For measuring the similarity between data values, they interpret them as multisets of words and use token-based measures.
%	Thus, they consider the similarity of data values on a more abstract level than we do
	The authors of \cite{AMN+11} identified the need to integrate %semantic information in form of 
	domain knowledge into such clustering techniques.
%	Those approaches do not focus on domain-specific similarities of data values as we do.
	In summary, those approaches do not focus on domain-specific similarities of data values as we do and they do not aim to analyse the quality of data models.

\subsection{Identifying Quality Problems in Data Models	and Meta-models} %Conceptual Models 
%\viola{Remove ``Conceptual Models'' from title?}
%\arno{Done: They do not occur in the text anymore. }
	%\viola{Merged sections on data models and meta-models}
%	A conceptual basis for \emph{data model} quality management was laid in \cite{MS94}. 
	A framework for \emph{data model} quality management was proposed in~\cite{MS94}.
	It comprises quality factors of varying importance, quality metrics, and improvement strategies.
%	The authors developed a framework with quality factors of varying importance, quality metrics, and improvement strategies.
%	They evaluated that framework in~\cite{MS03}.
%	It is evaluated in in~\cite{MS03}. 
%	The quality factors they identified are completeness, simplicity, flexibility, integration, understandability, and implementability.
	
%	In a similar vein, 
%	Similarly, a collection of quality attributes for \emph{meta-models} was presented by Bertoa and Vallecillo~\cite{BV10}.
	In a similar vein, a collection of quality attributes for \emph{meta-models} was presented in~\cite{BV10}.
%	 and the idea of defining measures for those quality attributes was mentioned.
%	As a next step, the authors suggest the definition of measures for those quality attributes.
%	Based on the quality attributes defined in \cite{BV10},
	Based on those quality attributes,
%	and
	an empirical study on the perception of meta-model quality was performed~\cite{HKB+16}.
	It showed that ``the perceived quality was mainly driven by the meta-models' completeness, correctness and modularity''~\cite{HKB+16}.
%	It showed that ``the perceived quality was mainly driven by the meta-models' completeness, correctness and modularity while other quality attributes could be neglected''~\cite{HKB+16}.
	The authors noted that, in general, completeness and correctness are very hard to measure.		
%	Our approach allows analysing the completeness of models by providing an overview of additional information encoded implicitly in data values due to a lack of explicit support.
	
%	Having quality attributes at hand, %quality-driven 
	The detection and resolution of meta-model smells is presented in~\cite{BRI+19}, an approach that is based on quality assurance for models in general~\cite{AT13}.
	Furthermore, there is an approach to meta-model testing via unit test suites and domain-specific expected properties with metaBest as well as an example-based construction of meta-models with metaBup~\cite{LGL14,LGL14b,Lopez-Fernandez15}. 
%	These approaches allow the analysis (i.e. validation and verification) of meta-model quality with respect to known requirements, either in form of domain-specific expected properties of the meta-model or generic smells.
	
%	\gabi{I would erase this as it is mostly redundant to the introduction: ``One of the findings of Moody and Shanks~\cite{MS94} was that metrics are of limited use for analysing data model quality as research participants rated qualitative descriptions of quality problems more useful than quantitative ones. 
%	We propose a \emph{qualitative analysis} of data model quality instead, which may be of larger help in improvement processes for data models. 
%	Our approach allows analysing models with respect to some of the quality attributes identified in \cite{MS94} and \cite{BV10}, such as completeness, detailedness and comprehensibility, not via generic measures but via an interactive approach driven by \emph{domain knowledge}.''}	
		
	While analysis approaches based on metrics~\cite{MS94}, smells~\cite{BRI+19} and expected domain-specific properties of models~\cite{LGL14b} are concerned with the quality of existing structures, our bottom-up approach can find out \emph{missing structure} and allows investigating previously \emph{unknown requirements}.
	Due to this complementary nature of our approach, we are convinced that
	it is worthwhile to investigate the relevance of data value clustering also for analysing meta-model quality, especially if unstructured attribute values are allowed.
%	data value clustering is also relevant for analysing the quality of meta-models, that include less unstructured attributes.
%	It could also be relevant for analysing logs of models-at-runtime (cf. \cite{SZ16}).
	
%	We conjecture that meta-models which allow some forms of unstructured data in attribute values would profit the most.
	Models-at-runtime (cf. \cite{SZ16}), for example, may be used to store structural information about system logs. 
	Applying our approach to models-at-runtime may reveal problems in their meta-models due to some heterogeneity in system logs. 

\section{Conclusion}
\label{sec:conclusion}
%\viola{Replace ``domain expert'' by ``user''?}
% advantages of bottom-up approach
%   exploaration, new undetected problems, new insights
%   generic since only data values are considered

%We presented an approach to support the detection of quality problems in data models.
%We presented a generic approach to support the detection of quality problems in data models via clusterings of data values by syntactic similarity.
%We presented a generic approach to support the detection of data quality problems via clusterings of data values by syntactic similarity.
We present a bottom-up approach to detecting quality problems in data models via clusterings of data values along syntactic similarity.
%In this paper, we focused on quality problems in data models which manifest in heterogeneous data. % quality problems in data.
%\viola{Directly state quality analysis of data models as goal of approach above, i.e. remove detection of \emph{data} quality problems?}
%\gabi{yes}
%To support the detection of data quality problems, particularly those caused by quality problems in data models, we presented a bottom-up approach based on clustering data values by syntactic similarity.
%\viola{Adedd sentence below.}
The approach is generic in that it is independent of the database technology used and can be adapted to different domains via the configuration.
%It can be configured according to domain knowledge and allows an explorative analysis of existing data.
%\viola{Remove the following sentence?}
The investigation of clusterings can provide new insights concerning the actual usage of a data model, from which domain experts can derive unknown quality problems and requirements of the data model.
%and improvements of the data model.
%Domain experts may use their insights for reasoning about quality problems and improvements of a data model.
%Domain experts may derive quality problems and possible improvements of the data model.
We provide a proof-of-concept implementation that allows experimenting with different configurations.
% for data value clustering. 
%\gabi{Really? We have written in the paper that we configured the clustering process, the domain experts just evaluated the results.}
%\viola{Removed ``domain experts''. OK? Or should we mention again that the GUI is still under development?}
%of the approach.
%We used this implementation to evaluate how far our clustering approach supports the detection of quality problems in data models. 
Our evaluation in the domain of cultural heritage data showed that the approach supports the detection of quality problems in data models, especially missing structures. 
In future work, we will further investigate how domain knowledge can be mapped to configurations of data value clustering straightforwardly without technical knowledge.
%Based on the findings, we will further develop and empirically evaluate the GUI of the tool.
% (1)
%Specifically we will explore how clustering parameters can be specified easier with less technical knowledge required.
% (2)
We will also investigate facilities to support experts in deciding whether further iterations are needed and in modifying a  configuration suitably.
%Moreover, we will develop a questionnaire for evaluating a clustering and deciding whether a further iteration is needed. 
%follow-up view for the process to help the domain experts to improve the parameters on further iterations.
%Currently we envision a questionnaire on how satisfied the domain experts are with certain aspects of a produced clustering.
%The answers given by an expert shall either directly modify clustering parameters or cause hints for parameter inputs.
%The answers given by an expert could be used to suggest modifications of the configuration in the next iteration.
%Based on the answers given by experts, modifications of the configuration could be suggested.
% (3)
%Furthermore, experts shall be supported in interpreting a clustering, for example, by means of generic quality problems and improvements of data models (similar to class model smells and refactorings~\cite{AT13}).
%\viola{Modified sentence below such that data adaption is also mentioned. OK?}
Furthermore, experts shall be supported in categorising detected problems and improving the data model (e.g. similar to class model smells and refactorings~\cite{AT13}) as well as adapting the data to the changes. 
%for example, by means of generic quality problems and improvements of data models (similar to class model smells and refactorings~\cite{AT13}).  
%Furthermore, experts shall be supported in interpreting the clustering.
%For example, they could use a list of generic quality problems in data models (similar to UML class model smells~\cite{AT13}) to 
%assign problems to clusters.
%%identify data model problems in clusters
%%to clusters.
%Based on those identifications, appropriate quality improvements of the data model could be suggested (similar to UML class model refactorings~\cite{AT13}).
Ultimately, the whole workflow and the GUI will be evaluated empirically.
%	%The underlying relation between problems and improvements could be determined via a survey of domain experts
%
%It is also up to future research to support domain experts in identifying quality problems and deciding on improvement strategies via the clustering.

Our approach 
%to the detection of quality problems in data models 
opens up new lines of research: 
%Application to meta-models
So far, we have applied it to detect quality problems in data models.
It may also be useful to  find out missing structure and to investigate previously unknown requirements in conceptual models and meta-models as long as heterogeneous data values are concerned.
%The clustering approach could be used to find, for example, noise in log information as well as incompleteness.
%\viola{I do not clearly understand the example above. Incompleteness of what? For which reason/motivation would we want to find noise and incompleteness? For model quality analysis again? Or do you suggest using our approach for completely different tasks?}
%\viola{Is the above still appropriate after we removed approaches that cluster system logs from related work?} 
%\gabi{Sentence rephrased.}
% should b e more vague. Metamodels for models at runtime maz not alwazs fit hundred percent to the information of system logs. 
%
%More than one data field
%Up to now, our clustering approach takes just one data field as input at a time and clusters its values.
%In the future, we want to experiment with several related fields or even a whole pattern of the data model. 
In future, we also want to experiment with the clustering of values of several related data fields (identified, for example,  with  association rule mining~\cite{AgrawalIS93}).
%Data values may even be interrelated by some data model pattern to detect further quality issues of data models.
% or even a whole pattern of the data model to detect further quality issues of data models.
%\gabi{Sentence adapted.}
%\viola{Added reference to association rule mining above.}
%To identify related data fields, association rule mining~\cite{AgrawalIS93} could be used.
%We are interested in finding out further quality issues of data models, such as some lack of agreement on concepts and terminology, by clustering suitable amounts of data. 
%
%In addition, 
%The experts we interviewed for the evaluation derived some quality problems of related data transformations from the clusterings.
And last but not least, observations by the interviewed domain experts imply that
%The interviewed domain experts observed that the clusterings 
%observed that the heterogeneity of data values can also be caused by data transformations.
%Thus, 
our approach may also be useful to explore the quality of data transformations from the bottom-up since heterogeneity of data values may also indicate quality issues in data transformations.
We expect that quality assurance of model transformations would profit from those findings. 

	\section*{Acknowledgment}
This work is partly	funded by the German Federal Ministry of Education and Research (Grant No.: 16QK06A).
We would like to thank Markus Matoni, Oguzhan Balandi, Martha Rosenkötter and Regine Stein for their valuable comments.
	
	\clearpage
	
	\bibliographystyle{IEEEtran}
	\bibliography{IEEEabrv,literature}	
	
	\shortversion
		{}
		{\clearpage
\appendices

\section{Evaluation}\label{appendix:evaluation}
In the following, we present configuration details of the evaluation of our approach presented in Sec.~\ref{sec:evaluation} as well as the Excel sheets we used to present clusterings to domain experts.

\subsection{Configurations}

% duplicate removal in all four cases
% hierarchical clustering with complete linkage in all cases

The configuration settings that were used to calculate the four clusterings underlying our evaluation are specified at~\cite{eval_configs} and outlined in the following.
For each clustering, we outline the data value abstraction, the weights used for calculating the edit distance and the configured clustering algorithm applied.
%Our thought process that went into the configurations is also outlined in the following.
%Note that concerning the distance weights, only their ratios are relevant.
%%Further, note that we specified the weights based on our domain knowledge and expectations but there is still a degree of arbitrariness.
%Further, note that based on our domain knowledge we were able to identify vague relations between different weights, such as ``A should be weighted much higher than B''.
%However, there remains a degree of latitude in determining concrete values that satisfy these relations.
%Therefore, we support multiple iterations to experiment with different configurations.
%\viola{Added brief explanation above. Should we move this to Sec.~\ref{sec:distance}?}
%But there is still a degree of arbitrariness in the specification of concrete values, which conform to these relations.
%We present the weights that were used to calculate the four clusterings evaluated and interpreted by the domain experts.
%Further note that 
%In all four cases, duplicate removal was part of the abstraction.
%Furthermore, in all four cases hierarchical clustering with complete linkage was applied.

\subsubsection{Artist Name}
The documentation of this field is outlined in Sec.~\ref{sec:example}.

\paragraph{Data Value Abstraction}

% letters: sequence sequences lower or upper case
% digits: digits
% special: none

Sequences of upper or lower case letters separated by space were mapped to the same letter as they probably all represent sequences of names, such as first name followed by middle names.
All digits were mapped to the same digit as the concrete digits do not alter the values' meaning significantly but the length of digit sequences may hint at the kind of information that is encoded.
Special characters were preserved as they often encode special meaning, which we are interested in.
Redundant abstracted values were removed.

\paragraph{Distance Calculation}
% basic edit distance

%[    0,    1,   10,   15,     1,   200,   100,    1],  #
%[    1,    0,   11,   16,     2,   201,   101,    2],  # a-zA-Z
%[   10,   11,    0,   25,    11,   210,   110,   11],  # 0-9
%[   15,   16,   25,    0,    16,   215,   115,   16],  # space
%[    1,    2,   11,   16,     0,   201,   101,    2],  # -'
%[  200,  201,  210,  215,   201,     0,   300,  201],  # ,
%[  100,  101,  110,  115,   101,   300,   200,  101],  # special
%[    1,    2,   11,   16,     2,   201,   101,    2],  # rest

The basic edit distance was used with the insertion and deletion weights given in Table~\ref{tab:weights_artist_short}.
Letters (representing sequences of upper or lower case letters separated by space), hyphens and apostrophes are expected in the abstracted artist name values, thus weighted low.
As digits and additional blank spaces cause some variation in meaning, they are weighted higher.
%Other special characters, which we do not expect in artist names, may impact the meaning even more, they are weighted the highest.
We actually do not expect other special characters in artist names, but if they occur, they often encode special meaning, which we are interested in revealing.
Therefore, they are weighted much higher.
We expect that, analogous to typical spelling variants, for some artist names first, middle and last names are given in this order while for others the last name may be given first, followed by a comma followed by first and middle name.
Commas are highest since we wanted to separate these variants in the clustering.
The column ``other'' includes mostly individual letters not included in the 26 letters of the basic Latin alphabet, such as ``Ł'', thus the weight is low.
\gabi{Example?}
\viola{Reworded and added example.}

\begin{table}
	\renewcommand{\arraystretch}{1.3}
	\caption{Basic Edit Distance Weight Matrix of Artist Names}
	\label{tab:weights_artist_short}
	\centering
	\begin{tabular}{l|ccccccc}
		  & Letters & - and ' & Digits & Space & Special & Comma  & Other \\ \hline
		- & 1       & 1       & 10     & 15    & 100     & 200    & 1     \\		
	\end{tabular}
\end{table}

%\begin{table}
%	\renewcommand{\arraystretch}{1.3}
%	\caption{Distance Weight Matrix of Artist Names}
%	\label{tab:weights_artist}
%	\centering
%	\begin{tabular}{l|cccccccc}
%		& - & letters & digits & space & -' & , & special & rest \\ \hline
%		- & 0 & 1 & 10 & 15 & 1 & 200 & 100 & 1 \\
%		letters & 1 & 0 & 11 & 16 & 2 & 201 & 101 & 2 \\
%		digits & 10 & 11 & 0 & 25 & 11 & 210 & 110 & 11  \\  
%		space & 15 & 16 & 25 & 0 & 16 & 215 & 115 & 16  \\  
%		-' & 1 & 2 & 11 & 16 & 0 & 201 & 101 & 2  \\
%		, & 200 & 201 & 210 & 215 & 201 & 0 & 300 & 201  \\
%		special & 100 & 101 & 110 & 115 & 101 & 300 & 200 & 101  \\
%		rest & 1 & 2 & 11 & 16 & 2 & 201 & 101 & 2  \\    
%	\end{tabular}
%\end{table}

\paragraph{Clustering}
We used hierarchical clustering with complete linkage and applied a distance threshold of 700.

% distance_threshold = 700

\subsubsection{Dating}
The documentation of this field is outlined in Sec.~\ref{sec:example}.

\paragraph{Data Value Abstraction}

% letters: to lower
% digits: integer
% special: none

All upper case letters were mapped to their lower case equivalent as capitalisation does not seem to significantly impact the meaning of the dating values.
As we do not expect a wide variety in the textual components of the values, we do not abstract from concrete letters or the length of letter sequences.
All sequences of digits were mapped to the same digit as they represent similar meanings.
Special characters were preserved as they often encode special meaning, which we are interested in.
Redundant abstracted values were removed.

\paragraph{Distance Calculation}

% Levenshtein distance

%[ 0,  1,  4],  #
%[ 1,  1,  4],  # 0-9
%[ 4,  4,  4],  # rest

The Levenshtein distance was used with the weights given in Table~\ref{tab:weights_dating}.
Mainly, digits (representing sequences of digits) are expected and considered quite similar to each other.
Other characters, i.e. letters and special characters, are weighted higher as they are expected to cause much greater dissimilarity in the meaning of the dating values.
\gabi{Why considerably higher weights?}
\viola{Reworded.}
Because we do not see a difference between the insertion or deletion and substitution of characters, we unified the weights all to 4.
In contrast to the basic edit distance,
this refinement reduces the dissimilarity between values of the same length.
For example, this reduces the dissimilarity between values like \lstinline|x| and \lstinline|y|, 
%which both are used to indicate unknown dates.
which both were used in this data field as placeholder for unknown dates.
%or 
It further reduces the dissimilarity between \lstinline|1895/1902| and \lstinline|1908-1909|, where \lstinline|/| represents a point in time between two dates and \lstinline|-| is used to represent periods of time.
Due to other more decisive anomalies expected in this data field, we decided that such differences in values should be weighted lower.

\begin{table}
	\renewcommand{\arraystretch}{1.3}
	\caption{Levenshtein Distance Weight Matrix of Datings}
	\label{tab:weights_dating}
	\centering
	\begin{tabular}{l|ccc}
		       & - & Digits & Other \\ \hline
		-      & 0 & 1      & 4     \\
		Digits & 1 & 2      & 4     \\
		Other  & 4 & 4      & 4     \\        
	\end{tabular}
\end{table}

\paragraph{Clustering}
We used hierarchical clustering with complete linkage and applied a maximum cluster number of 25.
% n_clusters = 25

\subsubsection{Measurement Unit}
The documentation of this field is outlined in Sec.~\ref{sec:example}.

\paragraph{Data Value Abstraction}

% letters: none
% digits: float
% special: none

Letters were preserved.
All sequences of digits were mapped to the same digit.
Those separated by a comma were all mapped to another digit.
Special characters were preserved as they often encode special meaning, which we are interested in.
Redundant abstracted values were removed.
Details are explained in Sec.~\ref{sec:compression}.

\paragraph{Distance Calculation}

% Levenshtein distance

The Levenshtein distance was used with the weights given in Table~\ref{tab:weights_units}.
Details are explained in Sec.~\ref{sec:distance}.

\begin{table}
	\renewcommand{\arraystretch}{1.3}
	\caption{Levenshtein Distance Weight Matrix of measurement units}
	\label{tab:weights_units}
	\centering
	\begin{tabular}{l|cccc}
		        & \multicolumn{1}{l}{-}    & \multicolumn{1}{l}{Digits} & \multicolumn{1}{l}{Letters} & \multicolumn{1}{l}{Special} \\ \hline
		-       & -                        & {\color[HTML]{656565} 2}   & {\color[HTML]{656565} 1}    & {\color[HTML]{656565} 2} \\
		Digits  & {\color[HTML]{656565} 2} & 0                          & {\color[HTML]{656565} 3}    & {\color[HTML]{656565} 4} \\
		Letters & {\color[HTML]{656565} 1} & {\color[HTML]{656565} 3}   & 1                           & {\color[HTML]{656565} 3} \\
		Special & {\color[HTML]{656565} 2} & {\color[HTML]{656565} 4}   & {\color[HTML]{656565} 3}    & {2}              
	\end{tabular}
\end{table}

\paragraph{Clustering}
We used hierarchical clustering with complete linkage and applied a distance threshold of 3.5.
% distance_threshold = 3.5

\subsubsection{Attribution Qualifier}

The LIDO documentation describes the field as follows.
\emph{``Definition: A qualifier used when the attribution is uncertain, is in dispute, when there is more than one actor, when there is a former attribution, or when the attribution otherwise requires explanation.
How to record: Example values: attributed to, studio of, workshop of, atelier of, office of, assistant of, associate of, pupil of, follower of, school of, circle of, style of, after copyist of, manner of...''}

\paragraph{Data Value Abstraction}

% letters: lower or upper case letters
% digits: digits
% special: none

All lower and upper case letters were mapped to the same letter as we expect a variety of letters in these values, but based on the documentation we expect entries of similar length.
All digits were mapped to the same digit as the concrete digit does not matter in this context.
Special characters were preserved as they often encode special meaning, which we are interested in.
Redundant abstracted values were removed.

\paragraph{Distance Calculation}

% Levenshtein distance

%[    0,    1,    2,   30,   20,   100,   90],  #
%[    1,    0,    2,   30,   20,   100,   90],  # a-z
%[    2,    2,    0,   30,   20,   100,   90],  # A-Z
%[   30,   30,   30,    0,   30,   100,   90],  # 0-9
%[   20,   20,   20,   30,    0,   100,   90],  # space
%[  100,  100,  100,  100,  100,   100,  100],  # special
%[   90,   90,   90,   90,   90,   100,   90],  # rest

The basic edit distance was used with the insertion and deletion weights given in Table~\ref{tab:weights_attribution}.
Mainly, letters are expected, thus weighted low.
Since blank spaces may imply interesting extensive explanations of the attribution, they are weighted higher.
As digits are not expected at all, they are weighted even higher.
Since special characters are often used to encode special meaning, such as uncertainty, they are weighted the highest.
\gabi{Why do you use such high weights here in contrast to the ones in previous examples?}
\viola{Only the relation between the weights is relevant, the height itself is not.}

\begin{table}
	\renewcommand{\arraystretch}{1.3}
	\caption{Basic Edit Distance Weight Matrix of Attribution Qualifiers}
	\label{tab:weights_attribution}
	\centering
	\begin{tabular}{l|cccc}
		& Letters & Space & Digit & Special \\ \hline
		- & 1     & 20    & 30    & 100 \\		
	\end{tabular}
\end{table}
%\begin{table}
%	\renewcommand{\arraystretch}{1.3}
%	\caption{Levenshtein Distance Weight Matrix of Attribution Qualifiers}
%	\label{tab:weights_attribution}
%	\centering
%	\begin{tabular}{l|ccccc}
%		        & -   & letters & space & digits  & special \\ \hline
%		-       & 0   & 1       & 20    & 30      & 100     \\
%		letters & 1   & 0       & 20    & 30      & 100     \\		
%		space   & 20  & 20      & 0     & 30      & 100     \\  
%		digits  & 30  & 30      & 30    & 0       & 100     \\
%		special & 100 & 100     & 100   & 100     & 100     \\
%	\end{tabular}
%\end{table}

\paragraph{Clustering}
We used hierarchical clustering with complete linkage and applied a distance threshold of 100.
% distance_threshold = 100

\subsection{Calculated Clusterings}
Fig.~\ref{fig:excel} shows an excerpt of an Excel sheet used to present result clusterings to domain experts.
More precisely, it is an excerpt of the Excel sheet showing the clustering of abstracted measurement unit values via representatives.
In the second row the number of original values in each cluster is indicated.
The number of abstracted values per cluster is indicated in the third row.
Next to each representative value, the number of original values it represents is shown.
%The Excel files showing the clusterings as they were presented to the domain experts

All four Excel files can be found at~\cite{eval_sheets}.
%are provided with this document.

\begin{figure*}	
	\centering
	\includegraphics[width=\linewidth]{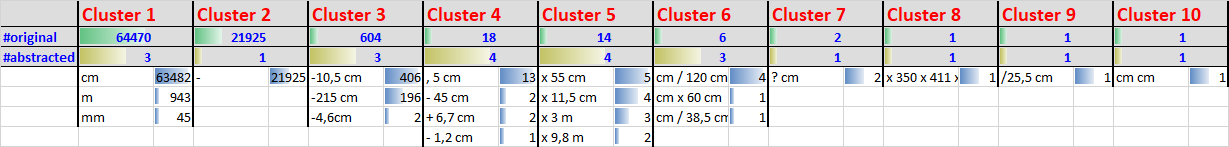}
	\caption{Excerpt of an Excel sheet showing a clustering of abstracted measurement unit values via representatives.}
	\label{fig:excel}
\end{figure*}

		\section{Tool Support} 
\label{appendix:tool}

%In this section we present some of the developed views from the graphical user interface.
In this section, we present some of the views of our graphical user interface concerned with configuring the data value clustering process.
Our goal is to provide an interface with which the numerous parameters can be configured intuitively and easily by domain experts based on their domain knowledge. 
The interface shall require as little technical understanding of the clustering process as possible.
The views are briefly described in Sec.~\ref{sec:gui}, details are provided in the following.
%These are designed to simplify the configuration of the steps of the approach.

%\viola{I think Fig.~\ref{fig:config_abstraction} and Fig.~\ref{fig:config_blobview} are most interesting. 
%	Maybe show Fig.~\ref{fig:config_hierarchical2} as an example for clustering algorithm configuration views. 
%	Maybe also show Fig.~\ref{fig:choose_clustering}. 
%	I would rather not show the others.}
%\arno{Scatter plot and dendrogram could be added.
%	I would also present matrix view as direct comparison to the blob view
%	Marked Figures with * in caption that might get removed.}

%screenshot names: 
%
%config_centre
%config_abstraction
%config_choose_costmap
%config_costmap
%config_blobview
%config_clustering_choice
%config_clustering_hierarchical1
%config_clustering_hierarchical2
%config_clustering_kmedoids
%config_clustering_dbscan
%config_clustering_affinity
%config_clustering_spectral

%\begin{figure}
%	\centering
%	\includegraphics[width=\linewidth]{appendix/config_centre}
%	\caption{* Configuration Centre}
%	\label{fig:config_centre}
%\end{figure}

%\subsection{Abstraction}
\subsection{Configuring the Abstraction of Data Values}
\begin{figure*}
	\centering
	\includegraphics[width=\linewidth]{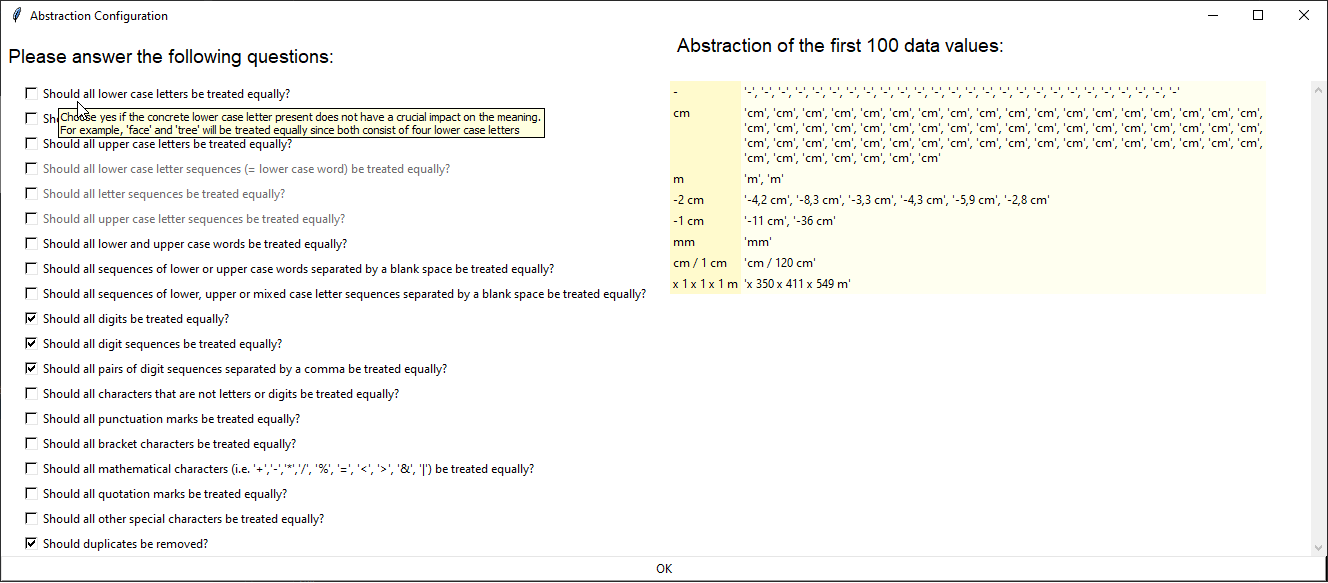}
	\caption{Configuration of the data value abstraction.}
	\label{fig:config_abstraction}
\end{figure*}
%\viola{This explanation does not really provide more details than the paragraph in Sec.~\ref{sec:gui}. Should we leave it out then?}
Fig.~\ref{fig:config_abstraction} shows the interactive view for configuring the abstraction step.
On the left-hand side, it shows a binary response questionnaire on the importance of several syntactical features of the data values.
These features are related to the abstraction rules discussed in Sec.~\ref{sec:compression}.
%Each of the options has an associated tool tip, that contains an explanation and an example.
Per option, a tool tip showing an explanation and an example is provided.
%For a representation of the configured abstraction step the first $100$ values are considered.
For demonstrating the configuration, the abstraction of the first $100$ original values is dynamically visualised on the right-hand side.
Per abstracted value, the corresponding original values are listed.
The screenshot shows the configuration of the abstraction for the measurement unit values, discussed in Sec.~\ref{sec:compression}.
%For demonstrating the result of the configured abstraction step, the first $100$ values are considered.
%The mapping of the configured abstraction is dynamically presented on the right hand side. 
%The abstracted values are dynamically updated on the right hand side.

%\begin{figure}
%	\centering
%	\includegraphics[width=\linewidth]{appendix/config_choose_costmap}
%	\caption{* Choice between Weight inputs}
%	\label{fig:choose_costmap}
%\end{figure}

\subsection{Configuring the Calculation of Distances}
\begin{figure}
	\centering
	\includegraphics[width=\linewidth]{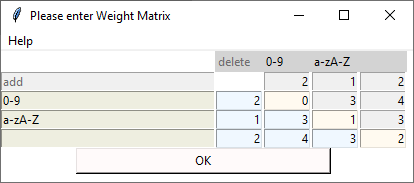}
	\caption{Configuration of the distance weights via the matrix view.}
	\label{fig:config_matrixview}
\end{figure}
For specifying the weights of the weighted Levenshtein distance we provide two input methods.
The first one requires the domain expert to input the weights directly as numbers.
The corresponding view is presented in Fig.~\ref{fig:config_matrixview}.
It shows the configuration for the running example on measurement units explained in Sec.~\ref{sec:distance}.
The view is structured analogously to the distance weight matrices presented, for example in Table~\ref{tab:weights}.
%On the right side the rows can be specified by enumerating the charcters.
On the left-hand side, the characters belonging to the same group are enumerated in a row.
%Hereby, 
The first column and row contain the weights of deleting and adding characters.
The last column and row represent all other characters not listed before. 
The matrix must be symmetric to fulfil the symmetry axiom for metrics.
Thus, the entry in a field is automatically copied to the symmetrically corresponding field. 

\begin{figure}
	\centering
	\includegraphics[width=\linewidth]{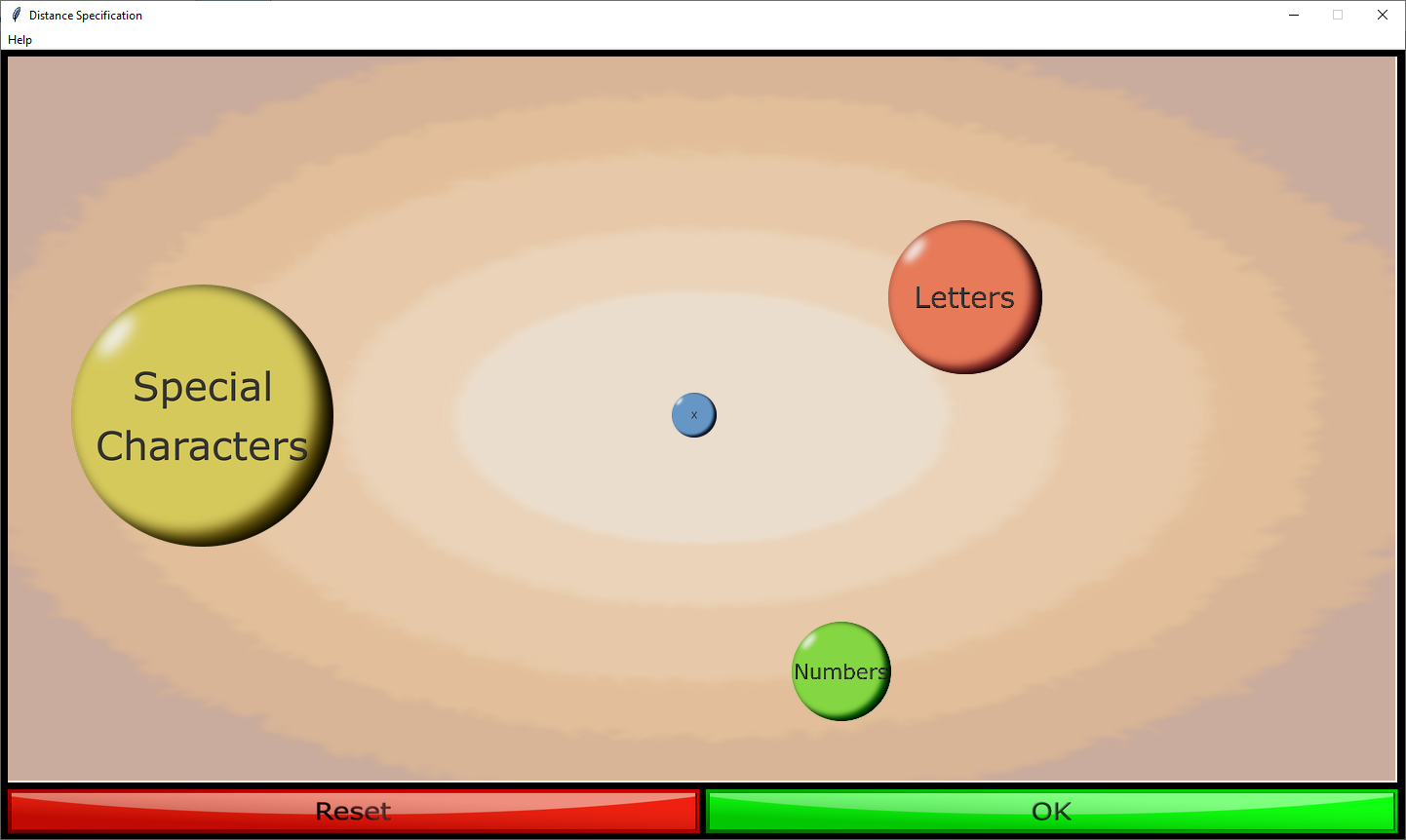}
	\caption{Configuration of the distance weights via the blob view.}
	\label{fig:config_blobview}
\end{figure}
%\viola{This explanation does not really provide more details than the paragraph in Sec.~\ref{sec:gui}. Should we leave it out then?}
An alternative is to input distance weights via a graphical view, which we call \emph{blob view}.
An example blob view is presented in Fig.~\ref{fig:config_blobview}. 
It configures distance weights of the running example on measurement units.
Groups of characters are presented as graphical objects, called \emph{blobs}.
They correspond to the columns and rows of the matrix in Fig.~\ref{fig:config_matrixview}.
The values in the matrix are formed as follows:
The graphical distance between two blobs is interpreted as the weight for substituting  %between
characters of corresponding groups.
To configure additions and deletions we use an additional blob, the small blue blob labelled with an X.
For configuring the weights, the user can move the blobs on the 2D canvas using drag and drop.
The weight for substitutions within a group is represented by the size of the corresponding blob.
The user can modify the size using the mouse wheel while hovering over the blob.

We are aware that this input does not give the full expressiveness of the matrix input as it would require $n-1$ dimensions where $n$ is the number of blobs.
In contrast, this view is limited to a 2D canvas.
Our intention is to provide an intuitive way to configuring distance weights.
It is a potentially more relatable visualisation compared to entering numerical values into a $n*n$ matrix.
A comprehensive user study is pending.

%\subsection{Selection of the Cluster Algorithm}
%\begin{figure}
%	\centering
%	\includegraphics[width=\linewidth]{appendix/config_clustering_choice}
%	\caption{Assistance in the selection of the clustering algorithm}
%	\label{fig:choose_clustering}
%\end{figure}
%In total we currently support six distinct clustering algorithms.
%To help the user with the choice we provide a short binary response questionnaire.
%The questionnaire is presented in Fig.~\ref{fig:choose_clustering}.
%It asks for expected characteristics of the data and the clustering that limit the set of appropriate clustering algorithms.
%Based on the answers a suggestion is made dynamically on the right side.

%\subsection{Cluster Configuration}
\subsection{Configuring the Clustering Algorithm}
%\begin{figure}
%	\centering
%	\includegraphics[width=\linewidth]{appendix/config_clustering_hierarchical1}
%	\caption{* Configuration of hierarchical clustering (1/2)}
%	\label{fig:config_hierarchical1}
%	%rename fig:config_hierarchical2 if removed
%\end{figure}
\begin{figure}
	\centering
	\includegraphics[width=\linewidth]{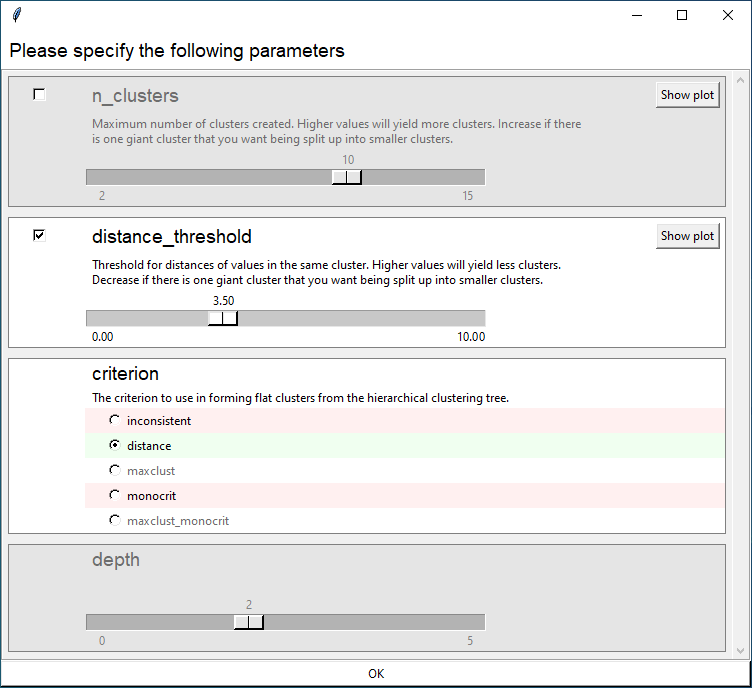}
	\caption{Configuration of hierarchical clustering.}
	\label{fig:config_hierarchical2}
\end{figure}
All the clustering algorithms come with many parameters.
We implemented a modularized view to configure these parameters as shown in Fig.~\ref{fig:config_hierarchical2} for the hierarchical clustering of abstracted measurement unit values.
Each module contains the name of a parameter and an explanation.
We distinguish between three kinds of parameters: boolean ones, numerical ones and enumerations.
Each module contains a checkbox, a slider or an enumeration with radio buttons, respectively.
Appropriate minimum and maximum values are automatically assigned to the sliders for numerical parameters.
For enumerations, %fitting 
tooltips are provided for each option.

Because we present the parameters of the clustering algorithms to the interface, the modules also mimic their dependencies.
%A module can have dependencies. 
For example, between \lstinline|n_clusters| and \lstinline|distance_threshold| in Fig.~\ref{fig:config_hierarchical2}, there is an alternating dependency: 
only one of the parameters is required.
\gabi{Why?}
\arno{Reformulated. These specific parameters describe different conditions on the final result for hierarchical clustering: Either all distances between clusters are higher than the distance\_threshold (float) or alternatively the number of clusters does not exceed n\_clusters (int)}
The parameter \lstinline|depth| is enabled only if the option ``inconsistent'' is chosen for the parameter \lstinline|criterion|.

%
%\begin{figure}
%	\centering
%	\includegraphics[width=\linewidth]{appendix/config_clustering_kmedoids}
%	\caption{* Configuration of kmedoids clustering}
%	\label{fig:config_kmedoids}
%\end{figure}
%
%
%\begin{figure}
%	\centering
%	\includegraphics[width=\linewidth]{appendix/config_clustering_dbscan}
%	\caption{* Configuration of DBSCAN clustering}
%	\label{fig:config_dbscan}
%\end{figure}
%
%
%\begin{figure}
%	\centering
%	\includegraphics[width=\linewidth]{appendix/config_clustering_affinity}
%	\caption{* Configuration of affinity propagation clustering}
%	\label{fig:config_affinity}
%\end{figure}
%
%
%\begin{figure}
%	\centering
%	\includegraphics[width=\linewidth]{appendix/config_clustering_spectral}
%	\caption{* Configuration of spectral clustering}
%	\label{fig:config_spectral}
%\end{figure}

\section{Data Value Clustering} 
\label{appendix:clustering}
%\subsection{Data Value Clustering}\label{sec:algorithm}
\begin{figure*}	
	\centering
	\includegraphics[width=\linewidth]{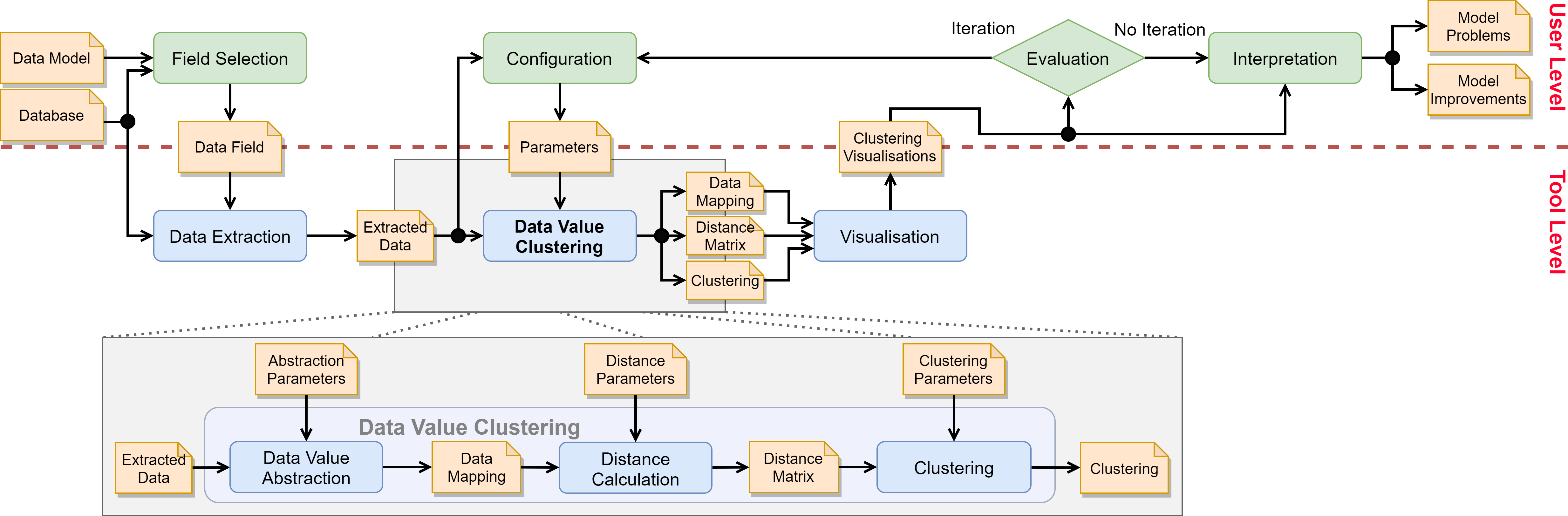}
	\caption{Detailed workflow of the approach to detecting quality problems in data models by clustering heterogeneous data values.}
	\label{fig:workflowdetailed}
\end{figure*}
In this section, we present the core %three-step 
algorithm of our approach in detail. % that underlies the clustering of data values.
It is visualised at the bottom of Fig.~\ref{fig:workflowdetailed}.
%``Execution'' step of the presented workflow.
%The ``Execution'' step of the presented workflow corresponds to the execution of the algorithm explained in this section.
%In the following we will discuss the algorithm for clustering data values that underlies the ``Execution'' step of the workflow in detail.
The inputs of this algorithm are a set of data values extracted from a database and a configuration of each of the algorithm steps.
%the compression configuration, the distance calculation, and the selected clustering algorithm.
%The main output is a clustering of the data values.
%Additionally, the distance matrix is output, which contains pairwise numerical distances between the data values.
%The algorithm consists of the following three steps:
The algorithm for data value clustering consists of the following steps: 
(1) an abstraction from the original data values, 
%(2) a calculation of pairwise distances between the compressed values and 
(2) a calculation of a distance matrix containing pairwise numerical distances between the abstracted data values and
(3) a clustering of the abstracted values based on the calculated distances.
%(2) a calculation of the distance matrix and 
%(3) a clustering of the values based on the distance matrix.
The main output of the algorithm is a clustering of the abstracted data values.
%Additionally, the distance matrix is output, which contains pairwise numerical distances between the data values.
Additionally, the mapping between original and abstracted values and the distance matrix are outputs.
%\viola{Added mapping between original and compressed values as output.}
In the following, these three steps and their configuration are discussed in detail.
For each step, we also explain the configuration for the running example, which we specified 
based on our domain knowledge as outlined in Sec.~\ref{sec:evaluation_setup}.
%Details on the development process of these configurations are given in Sec.~\ref{sec:evaluation_setup}.
%\viola{Added notes on how the configuration came about above. Or should we mention this even earlier? Maybe in Sec.~\ref{par:clustering}}

%\gabi{Give a short overview of the three steps here.}
%\viola{done}

%\gabi{Include a reference to this work in this section.}
%For incorporating domain knowledge in a suitable way, clustering with domain-specific usefulness scores is presented in \cite{CCC+17}. 
%	\gabi{How to compare with that work?}
%	\viola{I think it is interesting that they argue that ``incorporating domain expert input often improves clustering
%		performance'' since this is what we do at the configuration phase. So we could refer to their argumentation for motivating our approach. However, I do not consider this as related work since they propose to modify existing clustering algorithms to respect domain relevant scores. We just use existing clustering algorithms but do not optimize them.}
%\viola{Added reference in Sec.~\ref{par:config} as a motivation for configuration phase. OK?}
%\gabi{ok}

\subsubsection{Data Value Abstraction}\label{sec:compression}
% reasons for compression: efficiency and abstraction from detail 
% lower case, replacements, reduction, duplication removal, 
% compression depends on type of field (see questionary)
% discuss adequate compression for running example
%\gabi{Why do you compress the data?}
%\viola{Restructured section.}

% intro
The first step is an abstraction from the original data values.
%, i.e. from irrelevant syntactical details.
Only syntactical features that, according to domain experts, are of interest for clustering are maintained.
%The intention of this step is a first grouping of data values along syntactical properties.
%\gabi{This sentence added.}
%The original data values are mapped to a smaller set of shorter values, which only include syntactical features that are of interest for clustering.
%Thus, it is an abstraction from the concrete values.
%
%For our running example on measurement units, those interesting features include the occurrence of special characters such as minus and slash as they might indicate the encoding of special information.
%The lengths of digit sequences, however, represent irrelevant details that should be discarded by the compression since they do not impact the values' meaning significantly.
%\gabi{Start with some examples that clarify what kinds of compression are useful. We should think of several concrete examples for clustering that are rather diverse.}
%\viola{Added example above.}
%
% input & output
The inputs of this step are a set of data values and a configuration of the abstraction function.
%The compression function is a surjective; it maps the set of original values to a smaller set of compressed values.
%The output is a list of compressed values and a mapping between the original and the compressed values.
The output is a mapping between the original and the abstracted values.
%\viola{Moved information about input and output up here. OK?}

% motivation
%There motivation for the compression is that 
%The compression step basically consists of abstraction and reduction.
%The compression step is basically an abstraction from concrete values and a reduction of the length of values and the number of values.
% - abstraction
%In contrast to existing approaches to string clustering that detect spelling variants~\cite{PPS21,Cis08,FrenchPS00,SauleauPB05}, we do not intend to detect minor variations in the syntax of data values that have exactly the same meaning.
In contrast to existing approaches to string clustering that detect, for example, misspellings, typos, spelling variants and abbreviations~\cite{PPS21,Cis08,FrenchPS00}, we do not intend to detect minor variations in the syntax of data values that have exactly the same meaning.
%\gabi{What do you mean with ``with the same semantics''?}
%\viola{Added examples above.}
%The abstraction from concrete values is useful since, in contrast to existing approaches that detect spelling variants~\cite{PPS21,Cis08,FrenchPS00,SauleauPB05}, we do not intend to detect minor variations in syntax with same semantics.
%\arno{It is not necessary to do this, because it can be compensated by a good distance configuration.
%	The main difference is, that we cluster much coarser to detect more basic problems.
%	However the compression step (1) makes more distance functions useful, (2) makes it easier to find fitting parameters and (3) improves performance.}
%\viola{In my understanding an \emph{abstraction} from concrete values is indeed necessary. It is, however, not necessary to do it in the way we do it. Alternatively, it can be included in the distance calculation, as explained further below. But this would still be an abstraction from the concrete values.}
%compare the values directly when clustering.
%intend to cluster the original values by direct similarity.
Instead, the goal is to provide an \emph{overview of significant differences in the syntax} of all original  data values as syntactical heterogeneity may indicate quality problems of the data model.
Thus, based on domain knowledge, we abstract from the original values to remove irrelevant syntactical details, such as the length of digit sequences.
In our realisation of the abstraction step, the original data values are mapped to a smaller set of shorter values.
This has a positive impact on the performance of the distance calculation and the clustering performed subsequently.
An abstraction is defined by a \emph{set of rules dependent on the configuration}.
%The abstraction is defined by a list of %abstraction 
%with
These rules are based on two observations concerning data values.
%of data values from the cultural heritage domain.
First, due to their significant difference in meaning, three groups of characters can be distinguished on the top level, namely letters, digits and special characters.
Second, there are three interesting levels of abstraction from concrete data values:
%More precisely, we distinguish letters, digits and other non-letter characters.
%The abstraction levels are the following:
(1) abstracting from a concrete character of a specific group (considering, e.g., ``a'' equivalent to ``b'' and ``1'' to ``2''),
(2) abstracting from the length of a sequence of characters of a specific group (considering, e.g., ``a'' equivalent to ``painting'' and ``1'' to ``245''), and
(3) abstracting from the length of a sequence of characters of a specific group containing some separators (considering, e.g., ``a'' equivalent to ``the last supper'' and ``1'' to ``23.7'').
%\gabi{Down to here, the text explains design and should be merged into Sec. 3.}
%These rules are interesting as they roughly correspond to expected data formats in data models such as (1) single characters, (2) strings, integers and floats, while allowing finer distinctions as in (3).
%These rules are interesting as they roughly correspond to expected data formats in data models such as (1) single characters, (2) integers and (3) floats, while allowing finer distinctions concerning strings.
%These rules allow finer distinctions of strings, where some rules roughly correspond zu expected data formats in data models such as (1) single characters, (2) integers and (3) floats.
%These rules allow finer distinctions of strings.
%These rules roughly correspond 
%For numbers these rules correspond to expected data formats in data models such as (1) single characters, (2) integers and (3) floats while allowing finer distinctions.
These rules roughly correspond to expected data formats in data models while allowing finer distinctions.
For example, integers and floats can be mapped to (2) and (3) directly.
In other cases, the rules correspond to data formats with additional constraints.
%\viola{Modified above.}
%\viola{Added relation to data model above.}
%\gabi{Is the correspondence between rules and data formats correct?}
%%\viola{Yes it is. Actually, (1) and (2) also allow finer distinctions.}
%\arno{No. (1) = char, (2) = word, int, (3) = float (seperated by whitespaces), sentence (multiple words seperated by whitespaces). Modified Sentence.}
%\viola{Modified again.}
%\gabi{The examples do not fit to the data formats mentioned.  Do you mean: These rules roughly correspond to expected data formats in data models such as (1) single characters, (2) strings or integers and (3) floats or strings potentially restricted with  additional constraints.}
%\viola{Added examples concerning digits. Should we remove the examples concerning letters?}
%\viola{Unrestricted strings cannot be assigned to one of our categories. We consider letters, special characters and digits separately. But strings can contain all of these characters. Thus, in the sentence on data formats we do not mention strings.}
%\arno{Modified again: The correspondence applies only for numerical formats. We might should remove the sentence completely.}
Selected abstraction rules are applied to each of the original data values.
They replace each match of a specific regular expression by some character.
%Most rules consists of a regular expression and a character.
%At a rule's application, each match of the regular expression is replaced by the specified character.
%Additionally, there is one rule for replacing an upper case letter by its lower case equivalent and another one for removing duplicate values after all other rules have been applied.
There are additional rules for replacing an upper case letter by the equivalent lower case letter and for removing duplicate values after all other rules have been applied.
%Additionally, subsequent duplicate removal of compressed data values may be part of the compression.
%Whether further abstractions are suitable is subject to future research.
%Further abstractions may be added in the future.
Further abstraction rules corresponding to typical data formats such as dates may be added in the future.
\question{How can the performance be further increased?}

% configuration
%Having presented how the abstraction is performed, we still have to clarify which syntactical features are of interest.
%The answer highly depends on the data field analysed.
%Which syntactical features are of interest highly depends on the data field analysed.
It depends on the data field analysed which syntactical features are of interest.
That's why the abstraction must be \emph{configured based on domain knowledge}.
% based on their knowledge about the data model.
The configuration typically depends on expectations about the syntax of the data values.
% of interest.
It further depends on the kinds of syntax variations of the values that cause significant variations in their meaning.
Ultimately, the configuration determines which abstraction rules are applied.
%This can also be assessed based on the type of information represented.
%Both of these aspects are evaluated by domain experts based on the type of information represented by the data field, i.e. the data model and its documentation.

\question{How can domain knowledge systematically and user-friendly be mapped to a compression configuration?}
\question{Which other compressions are suitable?}

% input & output
%The inputs of the compression step are the original data values and the configuration of the compression function.
%Its output is a list of compressed values and a mapping between the original and the compressed values.
%\gabi{The information about input and output should come earlier.}

%\viola{Discuss typical compressions and their use cases?}
% e.g. compressing character sequences when analysing titles

%\viola{Mention impact on performance}

%\viola{Discuss compression of running example}
% running example
For the \emph{running example}, we configured the abstraction as follows:
We expect that a measurement unit value consists of a few letters representing an abbreviation of a measurement unit, such as \lstinline|cm|.
%We do not expect a wide variety in these abbreviations as there is only a small number of units relevant to measurements in the context of cultural heritage research.
%\gabi{Actually we expect a fixed enumeration of possible units, don't we?}
%Based on the LIDO documentation, we also do not expect digits or special characters.
% such as a minus sign.
%\gabi{Can we present some part of the documentation?}
%\viola{Done in Sec.~\ref{sec:example}.}
To identify quality problems in the data model LIDO, we need to find out what kinds of values are included in the data that do not suit our expectation and alter the values' meaning significantly.
Hence, interesting features for clustering are especially non-letter characters (i.e. special characters and digits), longer sequences of letters and unexpected combinations of letters.
%as there is only a small number of units relevant to measurements in the context of cultural heritage research.
%deviations in the concrete letters included.
The kinds of digits and the length of digit sequences (possibly separated by a decimal separator), however, are not considered decisive for the meaning of the values. 
%\gabi{digits altogether are not interesting, are they?}
%\viola{They are interesting because they are unexpected and change the meaning of the values significantly. They can indicate quality problems in the data models and transformations.}
%\gabi{Sentence adapted. Okay?}
%\viola{We see digits as non-letter characters, besides special characters. Adapted sentence.}
%\gabi{What does that mean?}
%\viola{Reworded sentence.}
%The length of digit sequence (possibly separated by a comma) is also not of interest.
% since variations in these lengths typically do not point to major variations in meaning.
%Concerning special characters, the meaning may depend on both the concrete characters used and the length of sequences of such characters.
%Special characters may be used as separators with special meanings which should not be abstracted away.
%\gabi{This sentence modified. OkaY?}
%\viola{OK}
%\gabi{Such kinds of compressions correspond to expected data formats in data models. Usually we distinguish one character, a number, a string, an enumeration, etc.}
%\viola{Yes, but the compressions allow finer distinctions, e.g one digit vs. a sequence of digits.}
%
%In summary, 
Thus, sequences of digits %of arbitrary length
are transformed into a single digit, sequences of digits containing a decimal separator are transformed into another single digit, while letters and special characters are preserved. %during the compression.
When applied to measurement unit values, 179 values given originally are abstracted to 22 values.
%this abstraction mapped the 179 distinct values given originally to 22 abstracted values.
The columns of Fig.~\ref{fig:clustering} represent several groups of original values that were mapped to the same abstracted value.
%For example, \lstinline|-10.5 cm| and \lstinline|-0.8 cm| were mapped to \lstinline|-2 cm|, while \lstinline|x 55 cm| and \lstinline|x 103 cm| were mapped to \lstinline|x 1 cm|.

% motivation
%In some cases, the abstraction of data values may be sufficient to provide an overview of the values' heterogeneous syntax.
%If the abstraction reduces the amount of values enough, 
If the abstraction produces a manageable amount of values, 
this may be sufficient to provide an overview of the values' diverse syntax.
%This is especially the case when the abstraction produces a manageable amount of values.
Otherwise, our approach suggests clustering the abstracted values to produce a manageable number of clusters containing similar abstracted values.
%\gabi{This should come earlier. Actually it is a nice motivation for the clustering (which needs distance definition before).}
As a prerequisite, distances between abstracted data values have to be computed. 
%\gabi{This motivation moved here.}
%\viola{ok}

\subsubsection{Distance Calculation}\label{sec:distance}
% string comparison
% different functions for calculating distances between strings
% show distances of running example

% intro
The next step is the calculation of a distance matrix containing pairwise distances (i.e., dissimilarities) between all the abstracted values, again depending on domain knowledge.
%
% input & output
The inputs of the distance calculation are a set of abstracted data values and a configuration of the distance function.
%distance function that is selected and suitably configured by the user.
Its output is a distance matrix. %a matrix that contains numerical distances between all pairs of compressed values.
Note that the distances between all original values that were mapped to the same abstracted value are considered to be zero.
%\viola{Moved information about input and output up here. OK?}

% motivation
%Cluster analysis is based on distances between the objects to be clustered.
Since our ultimate goal is to group all the abstracted values by syntactic similarity, the amount of similarity
%dissimilarity %or the corresponding dissimilarity 
must be quantified first.
For example, special characters are often used to encode special meaning.
% in the data values.
Hence, they serve as operators, and should cause high dissimilarity.
%\viola{Added observation above.}
%in the meaning of data values.
%
%The dissimilarity may also be considered as the distance between values.
%Thus, we calculate the pairwise distances between the compressed string values. 
%Thus, the next step of our algorithm is the calculation of pairwise distances between all the compressed string values.
%The next step of our algorithm is the calculation of pairwise distances, i.e. dissimilarities, between the compressed string values.
%Note that the distance between all original values that were mapped to the same compressed value is zero.
%We propose using an edit distance for this task as this 

% definition
String dissimilarity is typically measured via {\em edit distances} allowing different kinds of string operations~\cite{Navarro01}.
%\gabi{Why is it not useful here?}
%\viola{It is. We use it.}
%Since the compared strings may be of different lengths the Hamming distance\todo{cite}, which allows only substitutions, cannot be applied.
%The Jaro–Winkler distance\todo{cite} is also not suitable since ...?
%Since they are intuitive, allow comparing strings of different lengths and yielded satisfying results when we applied the approach to cultural heritage data, we consider the following edit distances to be useful in our context:
%We consider the following edit distances to be useful in our context:
Having applied the approach to cultural heritage data, we achieved reasonable results with the following edit distances:
%\gabi{Why? Motivate this selection first.}
%\viola{Added note that others may be suitable as well.}
%\gabi{What is the reason for this selection?}
%\viola{Difficult to answer. Modified sentence above.}
%\viola{Added modularity argument below.}
the basic edit distance that allows insertions and deletions of string characters only, 
and the Levenshtein distance~\cite{levenshtein1966}, which additionally allows substitutions. % in addition.
%and the Damerau–Levenshtein distance~\cite{damerau64}, which additionally allows transpositions of adjacent characters.
%We do not exclude that there may be other edit distances that are also suitable in our context.
%Through its modularity the approach supports other string distances to be integrated.
%We were able to achieve reasonable results with these two kinds of distances.
They may be accomplished with further distances when needed. 
\question{Which other edit distances are suitable?}

% weights:
%In both cases, we use configurable weights (i.e. costs) depending on the edit operations (i.e. insertion, deletion, or substitution) and the character(s) involved.
%In both cases, 
The dissimilarity between two values %that differ by insertion, deletion, substitution or transposition of certain characters 
depends on the data field analysed.
In a field representing the name of a person, for example, the insertion of an additional letter does not really cause dissimilarity whereas for a field representing the height of an object it does.
Therefore, we use \emph{configurable weights} %(i.e. costs) 
for each edit operation (namely insertion, deletion and substitution) applied to each possible character.
%\viola{Added two remarks below.}
Bear in mind that a character of an abstracted value, depending on the abstraction rules applied, may represent a character of a specific group or a sequence of certain characters.
Note that only the ratios of weights are relevant.
Domain experts must configure the weights based on their \emph{domain knowledge}.
%Note that only the ratios of weights are relevant.
%Further, note that we specified the weights based on our domain knowledge and expectations but there is still a degree of arbitrariness.
%Further, note that based on domain knowledge vague relations between weights can be identified.
%Typically, domain knowledge can be translated into vague relationships between weights.
Typically, vague relations between weights can be derived from domain knowledge.
%identified based on domain knowledge.
%Again, the configuration depends on their expectation. % concerning the syntax of the data values.
In general, edit operations of unexpected characters should be weighted higher than those of expected ones as they may indicate quality problems in the data model.
%\gabi{Why?}
%\viola{Added reason.}
Additionally, the more influence edit operations of certain characters have on the meaning of data values, the higher those weights should be.
%the weight of those characters should be.
%, such as ``A should be weighted much higher than B''.
However, there remains some leeway in determining concrete values that satisfy these relations.
Therefore, we support multiple iterations to experiment with different configurations.
%\viola{Extended paragraph above. OK?}

% when to use which edit distance:
%\viola{Moved the following paragraph down.}
%\viola{Modified paragraph below to emphasize advantage of basic edit distance.}
%\gabi{very nice.}
The basic edit distance corresponds to the Levenshtein distance if the weights of each substitution equal the sum of the weights of the corresponding deletion and insertion.
In this case, only the weights for deletions and insertions must be specified.
%As this seems sufficient in most cases,
Hence, it can serve as a starting point.
%The basic edit distance requires significantly less weights to be specified than the Levenshtein distance.
%In general, the basic edit distance should be sufficient in most cases and thus is a good starting point.
%The Levenshtein distance should be preferred over the basic edit distance 
%The Levenshtein distance should be used
%In later iterations
For further refinement, lower weights for character substitutions may be specified
if
%the substitution operation is needed, thus
%the substitution of a certain character by another is supposed to be less expensive than first deleting and then inserting a character.
%In the cultural heritage data, there are cases in which
different characters are considered quite similar but not equivalent (e.g. different kinds of quotation marks)
and the length of values is considered decisive for their meaning.
%Thus, they are not compressed into the same character.
%In such cases, the substitution operation is useful as a substitution may cause less dissimilarity than the insertion and deletion.
%Then the deletion and insertion of those characters may cause less dissimilarity in the context of a corresponding substitution than in other cases.
Then substitutions of those characters have less impact on the meaning than their deletions and insertions in other contexts.
%Thus, the weight for this substitution should be lower than the sum of the weights for the deletion and insertion.
%Lower weights for substitutions cause greater similarity between values of the same length.
%As the substitution operation is therefore needed, the Levenshtein distance should be used.

%the influence of inserting, deleting or substituting characters on the meaning of the value is decisive and should be mapped to lower or higher weights respectively.
%% input & output
%The inputs of the distance calculation are a set of compressed data values and a distance function selected and suitably configured by the user.
%Its output is a matrix that contains numerical distances between all pairs of compressed values.

%\viola{Discuss typical relations between weights of specific characters and their use cases}

%\viola{Discuss distance configuration of running example}

\question{How can domain knowledge systematically and user-friendly be mapped to a distance configuration?}

%Considering our \emph{running example} again, 
The chosen weights for the \emph{running example}
%The weights for our \emph{running example}
%of measurement units 
are presented in Table~\ref{tab:weights}.
The first column and row contain the weights for character deletions and insertions, respectively.
The other cells show the weights for substitutions of corresponding characters.

We chose the weights for deleting and inserting characters according to our expectation concerning the syntax of the values.
%For the \emph{running example} on measurement units, we chose the weights for deleting and inserting characters according to our expectation concerning the syntax of the values explained above.
Digits and special characters are unexpected, while letters are expected. 
Therefore, the weight for letters is lower than those for digits and special characters.
% and theirs is equal as both kinds are unexpected. 
%Thus, we specified the same weights for digits and special characters (which are unexpected characters), while we set that of letters  lower (as they are expected characters). % half as large.
%\viola{Show weight matrix?}

\begin{table}
	\renewcommand{\arraystretch}{1.3}
	\caption{Distance Weight Matrix for measurement units}
	\label{tab:weights}
	\centering
\begin{tabular}{l|cccc}
	        & \multicolumn{1}{l}{-}    & \multicolumn{1}{l}{Digits} & \multicolumn{1}{l}{Letters} & \multicolumn{1}{l}{Special} \\ \hline
	-       & -                        & {\color[HTML]{656565} 2}   & {\color[HTML]{656565} 1}    & {\color[HTML]{656565} 2} \\
	Digits  & {\color[HTML]{656565} 2} & 0                          & {\color[HTML]{656565} 3}    & {\color[HTML]{656565} 4} \\
	Letters & {\color[HTML]{656565} 1} & {\color[HTML]{656565} 3}   & 1                           & {\color[HTML]{656565} 3} \\
	Special    & {\color[HTML]{656565} 2} & {\color[HTML]{656565} 4}   & {\color[HTML]{656565} 3}    & {2}              
\end{tabular}
\end{table}

%We consider the three kinds of characters (namely letters, digits and special characters) as equally dissimilar to each other since they typically represent equally dissimilar meanings when occurring in data values.
%\viola{Added brief explanation above.}
%\gabi{I do not understand this sentence. The weights in the matrix are all different.A digit is more similar to a letter than to a special char.}
%\viola{Revised explanation below.}
%\gabi{Reformulated. Please check.}
%\viola{OK}
%\gabi{why?}
%\arno{We observed, that characters in data fields the values often are specified either to be a text or a number. 
%	Special characters are often used to express additional knowledge. 
%	Therefore we decided to distinguish between those three groups by default.}
%\gabi{This is an important information when using this clustering for detecting data model problems.}
%Thus, 
The numbers that are not on the diagonal represent substitutions of characters of different types.
We have no reason to assume that such substitutions have less impact on the meaning than deletions and insertions of corresponding characters in other contexts.
Thus, we specified the weights of such substitutions as the sum of deleting and inserting the characters.
%We specified the weights of substituting one such character by another as the sum of deleting and inserting the characters respectively.
%As argued in Section~\ref{sec:compression}, we consider different letters similar but not equivalent.
%As argued in Section~\ref{sec:compression}, 
%We observed that letters and non-letter characters typically represent very different types of information in data values.
%Based on typical semantic (dis)similarities, 

%\arno{remove following sentences?}
%\gabi{If space is needed.}
%For the running example considering measurement units, we chose the same weights for the deletion and insertion of  digits and other non-letter characters based on our expectations explained above.
%We chose the weights for deleting and inserting letters half as large.
%since we expect to find characters in measurement units. 
%Since all digit sequences (possibly separated by a comma) were replaced by the same digit during the compression (see Section~\ref{sec:compression}), the case of substituting one digit by another does not occur, and thus its weight is irrelevant.
%Note that, after the abstraction, only two different digits are left, one representing pure digit sequences (that are integers) and the other  one representing digit sequences that are each separated by a comma (that are decimal numbers).
After the abstraction, only two different digits are left, which represent integers and decimal numbers.
%one representing integers and the other representing decimal numbers.
%\gabi{Do you mean two different digits or numbers, i.e., integers and floats?}
%\viola{I mean two different digits, 1 and 2, are used to represent two different types of numbers, namely integers and floats.}
%In the context of measurement units, integers and decimal numbers can be considered equivalent.
%Therefore, we set the weight for substituting one digit by  the other to zero.
%Since in the context of measurement units, both can be considered equivalent, we set the weight for substituting one digit by the other to zero.
Since any numbers can be considered equivalent here, we set the weight for substituting one digit by the other to zero.
%Note that this is the reason why some points in the scatter plot in Fig.~\ref{fig:mds} overlap completely.
%
%Since we expect all measurement units to be expressed by abbreviations, we expect the data values to be of very similar (short) length.
%Thus, altering a specific letter has probably a lower impact on the general meaning of the value than inserting or deleting a letter.
%\viola{Modified explanation below.}
We consider different letters as non-equivalent but as quite similar to each.
Further, we expect abbreviations of measurement units, thus values of similar (short) length.
% other opposed to non-letter characters.
Hence, substitutions of letters have less impact on the meaning than deletions and insertions of letters in other contexts.
%\viola{Added explanation above.}
%Therefore, we consider different letters not as equivalent but as quite similar to each other opposed to non-letter characters.
% as the represented meaning typically d.
%\gabi{What does that mean?}
%\viola{Reworded sentence.}
So we chose the weight of substituting one letter by another 
%lower than %by another as half of the sum of
%the sum of 
equal to
the weights of deleting and inserting a letter, respectively.
Thus, for example, the values \lstinline|cm|, \lstinline|mm| and \lstinline|m| are equally dissimilar to each other.
%\gabi{Deleting and inserting a letter is equal to substituting it?}
%\viola{As explained above, we think that substituting letters causes less variations in meaning here than inserting or deleting a letter.}
%\viola{Modified explanation above.}
%\gabi{why? This applies to characters inside of one kind?}
%\viola{It applies to the substitution of a letter by another letter and to the substitution of a special character by another special character. As explained above, it does not apply to digits.}
The same applies to special characters.
%
%The distances between the abstracted measurement unit values calculated with this configuration are visualised in Fig.~\ref{fig:mds}.
The distances calculated with this configuration are visualised in Fig.~\ref{fig:mds}.
%\gabi{Again, all these settings are not argued. Why do you do it like this?}

\subsubsection{Clustering}\label{sec:clustering}
% order values into homogenous groups
% give overview of supported algorithms
% show clustering of running example

% intro
The final step of the algorithm consists in clustering a set of abstracted data values.
%
% input & output
The inputs of this step are a set of abstract data values, a distance matrix %a matrix of pairwise distances between the compressed values 
and a configured clustering algorithm.
%with a specific configuration.
%\gabi{The set of abstract data values is also an input.}
%\viola{Yes, added.}
Its output is a clustering of the abstracted data values.
%\viola{Moved information about input and output up here. OK?}

%% motivation
%%In some cases, the abstraction of data values may be sufficient to provide an overview of the values' heterogeneous syntax.
%%If the abstraction reduces the amount of values enough, 
%When the abstraction produces a manageable amount of values, 
%this may be sufficient to provide an overview of the values' heterogeneous syntax.
%%This is especially the case when the abstraction produces a manageable amount of values.
%Otherwise, our approach suggests clustering the abstracted values to produce a manageable number of clusters containing similar abstracted values.
%%\gabi{This should come earlier. Actually it is a nice motivation for the clustering (which needs distance definition before).}

% definition
%different clustering algorithms:
% affinity propagation, dbscan, hierarchical, kmedoids, optics, spectral
%Hence, the next step consists in applying a suitable clustering algorithm.
%using the distance matrix.
%to the pairwise distances between the compressed values that were calculated via a string distance function.
%\gabi{via a distance function? ..since the edit distance is already a concrete distance.}
%\viola{Modified. The important point is that the not every cluster algorithm can work with any distance function. We can only apply algorithms that support string distance functions.}
%Thus, 
We can only use clustering algorithms that operate on string distances.
%More precisely, they must allow to use a string distance such as the edit distance.
% TODO: any other constraints? which types of algorithms can (not) be used?
%The clustering algorithm k-means~\cite{macqueen1967}, for example,  is not an option since it is limited to the (squared) Euclidean distance which does not apply to string values.
%Whereas 
The set of suitable algorithms includes hierarchical clustering~\cite{hierarchical}, k-medoids~\cite{kmedoids}, DBSCAN~\cite{dbscan}, OPTICS~\cite{optics}, affinity propagation~\cite{affinity}, and spectral clustering~\cite{spectral}.
\question{Which other algorithms are suitable?}
%not be limited to one specific distance measure, as for example k-means is limited to squared Euclidean distance.
%Thus, clustering algorithms that only allow to use the (squared) Euclidean distance (e.g. k-means) cannot be used.

% configuration
The domain expert has to select one of these algorithms and \emph{configure} its parameters dependent on the data field analysed.
% based on domain knowledge.
%Hence, this configuration must be performed based on domain knowledge.
%Density-based algorithms such as DBSCAN~\cite{EsterKSX96} or OPTICS~\cite{AnkerstBKS99} should be preferred, for example,  when the data is expected to be noisy since they are robust to outliers.  %\gabi{Why?}
%Further 
Aspects that may influence the selection of an appropriate clustering algorithm and the parameter configuration are:
	determinism of the computed clustering, %(i.e. whether multiple executions should yield the same clustering),
	expected heterogeneity of cluster density, %(i.e. whether clusters of varying density are expected),
	expected cluster shapes (such as chains and spheres),
	expected heterogeneity of cluster sizes, %(i.e. whether clusters of varying size are expected), 
	robustness to outliers
	and	desired speed of execution.
\question{What do these aspects mean in the context of data values?}
%\gabi{These aspects should be explained in more detail.}
%\viola{Added brief explanation in brackets.}
%\viola{I think most of the given explanations are just redundant formulations and do not increase understandability. So I would remove them again.}
%\gabi{What do these aspects mean for data value clustering? For example,  a clustering should always be deterministic. So what do we expect from these clusterings?}
%\viola{I do not think that a clustering should always be deterministic. I do not understand your point.}
% supporting domain experts
%However, most of these aspects are very difficult to translate into domain languages.
The mapping between domain knowledge and these aspects %the configuration of the clustering step 
%is less obvious and more complicated here 
%is more complex than mappings of domain knowledge were in the previous steps.
is complex.
%Therefore, we must rely even more on (guided) experimental and iterative configuration. % here than in the previous steps.
%\viola{Added need for future research below.}
We provide a setting that allows experimenting with a variety of clustering algorithms.
It is up to future research to investigate which clustering algorithms and parameter settings are most suitable %in the context of our approach 
for detecting quality problems in data models
and how this depends on the kind of data considered.
%and for specific kinds of data.
%It is also open how the kind of data considered influences the algorithm choice. 
\question{Which clustering algorithms are most suitable for which kinds of data?}
\question{How can domain knowledge systematically and user-friendly be mapped to parameters of clustering algorithms?}

For our \emph{running example}, we chose hierarchical clustering with complete linkage to cluster abstracted measurement units and set the distance threshold to 3.5.
%\viola{reason}
This yielded 10 clusters, which are visualised in Fig.~\ref{fig:mds} and partly presented in Fig.~\ref{fig:clustering}.
%since we did not expect any noise, avoid the chaining phenomenon, ...?
% TODO: why
%\gabi{Any reason for this choice?}

%\subsection{Cluster Interpretation}
%% TODO: maybe remove this subsection and discuss cluster interpretation in evaluation section
%% multiple possible reasons for problems that manifest in data
%% present possible reasons: acquisition (human, software), model, transformation
%% discuss affected quality dimensions or even problems
%% discuss running example
%\viola{Do we want to discuss the kinds of problems that can be detected here or in Section~\ref{sec:evaluation} (Evaluation)?}
}

\end{document}